\documentclass{article}

\usepackage[preprint]{neurips_2025}

\usepackage[utf8]{inputenc} % allow utf-8 input
\usepackage[T1]{fontenc}    % use 8-bit T1 fonts
\usepackage{hyperref}       % hyperlinks
\usepackage{url}            % simple URL typesetting
\usepackage{booktabs}       % professional-quality tables
\usepackage{amsfonts}       % blackboard math symbols
\usepackage{nicefrac}       % compact symbols for 1/2, etc.
\usepackage{microtype}      % microtypography
\usepackage{xcolor}         % colors
\usepackage{amsmath}
\usepackage{graphicx}
\usepackage{multirow}
\usepackage{makecell}
\usepackage{siunitx}
\usepackage{wrapfig} 
\usepackage{tabularx}
\usepackage{adjustbox}
\usepackage{subcaption}
\usepackage{enumitem}

\setlist[enumerate]{leftmargin=2em}
\setlist[itemize]{leftmargin=2em}

\title{AlphaVAE: Unified End-to-End RGBA Image Reconstruction and Generation with Alpha-Aware Representation Learning}

\author{%
  \textbf{Zile Wang}$^{1}$\thanks{Equal contribution} \quad
  \textbf{Hao Yu}$^{1*}$ \quad
  \textbf{Jiabo Zhan}$^{1,2}$\thanks{This work was done when Jiabo Zhan was an intern at Tsinghua University.} \quad
  \textbf{Chun Yuan}$^{1}$\thanks{Corresponding author}  \\[.5em]
  $^{1}$Tsinghua University $^{2}$Beihang University \\[.2em]
  \texttt{\{wangzile23,yuh24\}@mails.tsinghua.edu.cn, yuanc@sz.tsinghua.edu.cn}
}

\newcommand{\uline}{\underline}
\newcommand{\benchmark}[0]{\textsc{Alpha}}
\newcommand{\method}[0]{\textsc{AlphaVAE}}

\begin{document}

\maketitle

\begin{abstract}
Recent advances in latent diffusion models have achieved remarkable results in high-fidelity RGB image synthesis by leveraging pretrained VAEs to compress and reconstruct pixel data at low computational cost.  
However, the generation of transparent or layered content (RGBA image) remains largely unexplored, due to the lack of large-scale benchmarks.  
In this work, we propose \benchmark{}, the first comprehensive RGBA benchmark that adapts standard RGB metrics to four-channel images via alpha blending over canonical backgrounds.  
We further introduce \method{}, a unified end-to-end RGBA VAE that extends a pretrained RGB VAE by incorporating a dedicated alpha channel. The model is trained with a composite objective that combines alpha-blended pixel reconstruction, patch-level fidelity, perceptual consistency, and dual KL divergence constraints to ensure latent fidelity across both RGB and alpha representations.  
Our RGBA VAE, trained on only 8K images in contrast to 1M used by prior methods, achieves a +4.9 dB improvement in PSNR and a +3.2\% increase in SSIM over LayerDiffuse in reconstruction. It also enables superior transparent image generation when fine-tuned within a latent diffusion framework.
Our code, data, and models are released on \url{https://github.com/o0o0o00o0/AlphaVAE} for reproducibility.
\end{abstract}

\section{Introduction}
\label{sec:intro}

Recent progress in conditional image synthesis has been driven by the synergistic combination of diffusion models~\citep{ddpm, improved-ddpm, 9878449} and variational autoencoders (VAEs)~\citep{kingma2022autoencodingvariationalbayes, Higgins2016betaVAELB, vqvae}, which together form a powerful framework for high-fidelity image generation.
Specifically, latent diffusion models~\citep{9878449} leverage a pretrained VAE decoder to decompress high-resolution RGB images from a significantly lower-dimensional latent representation, drastically reducing computational cost and memory footprint.
This compressed-domain synthesis has enabled state-of-the-art results in generating large-scale, semantically aligned images across diverse domains.
Consequently, the reconstruction fidelity of the VAE becomes a crucial determinant of overall generation quality.

Despite the maturity of RGB-centric pipelines, generation of transparent or layered content (RGBA) remains severely underexplored, even though most professional editing tools and graphic design workflows rely on layer-based composition with transparency. 
The scarcity of large-scale RGBA datasets and the sensitivity of pretrained latent diffusion models to shifts in latent space statistics pose significant obstacles. Moreover, existing approaches such as Text2Layer~\citep{zhang2023text2layerlayeredimagegeneration} and LayerDiffuse~\citep{layerdiffuse} typically model RGB and alpha channels separately, incurring extra encoder and decoder branches that increase parameter count and inference cost, and often reduce to matting-like post-processing rather than true reconstruction as shown in Figure~\ref{fig:blend}.

To address these challenges, we propose a unified, end-to-end RGBA VAE called \method{} that extends a pretrained three-channel VAE with an additional alpha channel via zero-initialization and channel-specific weight splits. 
We carefully design a composite training objective, combining pixel-level alpha-blended reconstruction loss, patch-level fidelity, perceptual consistency, and dual KL constraints, to ensure the reconstruction quality and preserve the original latent distribution while learning high-fidelity transparency representations. 

Traditional image quality metrics such as PSNR, SSIM~\cite{1284395}, and LPIPS~\cite{8578166} are defined for three-channel RGB images and cannot be directly applied to four-channel RGBA data.
Moreover, existing reconstruction and generation benchmarks exclusively target RGB content, resulting in the absence of a unified evaluation protocol for transparency-aware synthesis.
To bridge this gap, we introduce \benchmark{}, the first comprehensive benchmark for RGBA image reconstruction and generation, which leverages alpha blending over a fixed set of canonical backgrounds to seamlessly extend standard RGB metrics to four-channel images.

Our experiments demonstrate that, with only 8K training images, our method outperforms LayerDiffuse trained on 1M samples across multiple quantitative metrics (e.g. +4.9 dB for PSNR, +3.2\% for SSIM).
Furthermore, when fine-tuning the corresponding latent diffusion model with \method{}, we achieve superior transparent image generation compared to baseline pipelines.

\begin{figure}[t]
    \centering
    \includegraphics[width=\linewidth]{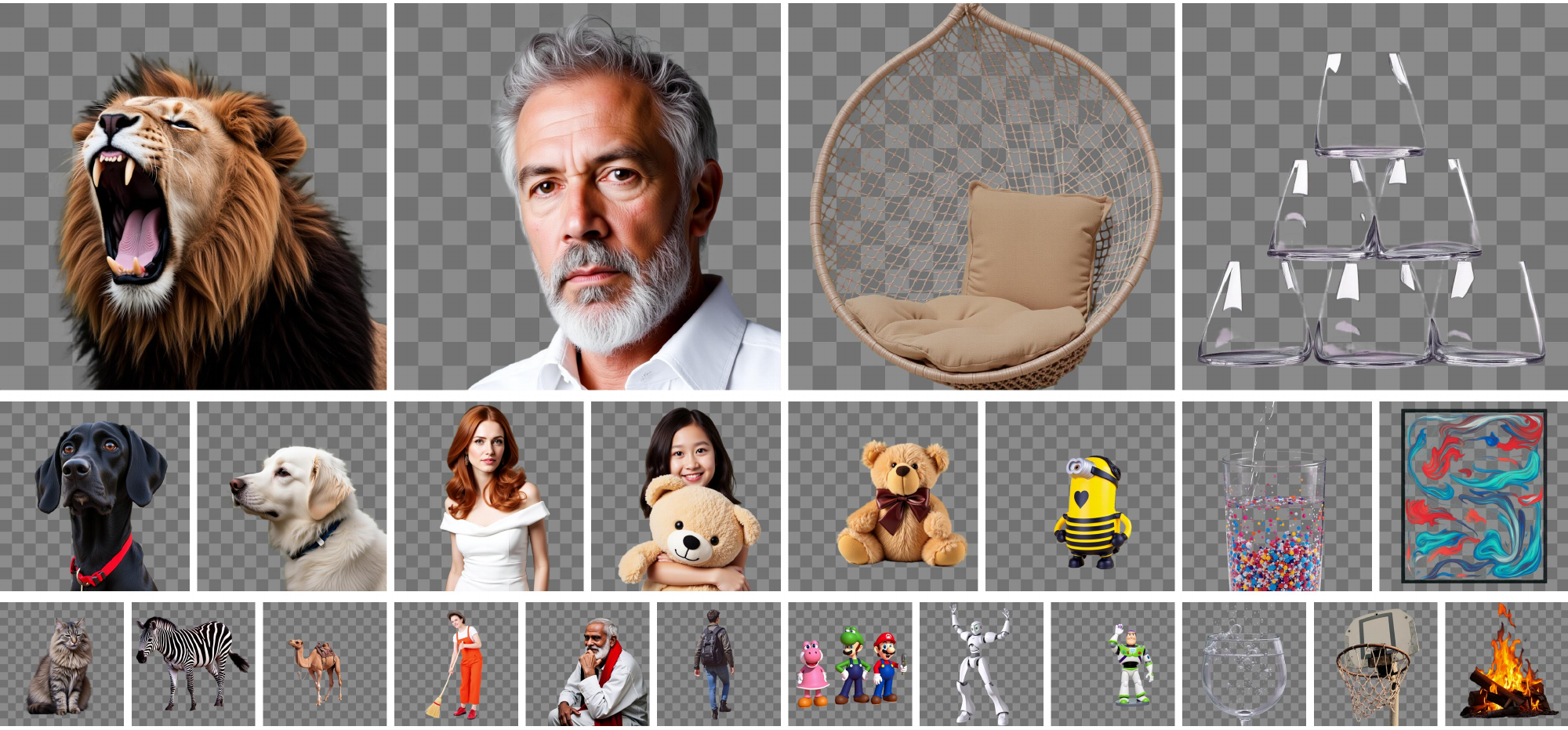}
    \vspace{1mm}
    \begin{tabularx}{\linewidth}{XXXX}
        \centering\makebox[0.25\linewidth]{animal} &
        \centering\makebox[0.25\linewidth]{portrait} &
        \centering\makebox[0.25\linewidth]{object} &
        \centering\makebox[0.25\linewidth]{transparent}
    \end{tabularx}
    \vspace{-1.2em}
    \caption{Qualitative results of image generation. Texts shown in each column correspond to category labels from the AIM-500 dataset and \benchmark{} test split. We sample images from different categories in the AIM-500 dataset and feed them into a captioning model~\cite{Qwen2.5-VL} to generate prompts. These prompts are then used to guide the generation of transparent images.}
    \label{fig:qualitative_diffusion}
    \vspace{-1.4em}
\end{figure}

In summary, our contributions are threefold:
\begin{itemize}
  \item We introduce \benchmark{}, a comprehensive RGBA benchmark that includes a curated dataset spanning diverse transparency scenarios and a suite of evaluation metrics that seamlessly extend standard RGB protocols through alpha blending.
  \item We propose \method{}, a unified end-to-end RGBA VAE architecture that augments a pretrained three-channel VAE with a dedicated alpha channel, accompanied by specialized reconstruction and regularization losses tailored for RGBA data.
  \item We validate our framework with extensive experiments on multiple datasets, demonstrating state-of-the-art VAE reconstruction and superior latent-diffusion–based transparent image generation using much less training data.
\end{itemize}

% The rest of this paper is organized as follows. Section~\ref{sec:related} reviews related work on transparency-aware modeling. In Section~\ref{sec:benchmark}, we detail the \benchmark{} benchmark design. Section~\ref{sec:method} describes \method{} architecture and training objectives. Experimental results are presented in Section~\ref{sec:exp}, followed by conclusions in Section~\ref{sec:conclusion}.

\section{Related Work}
\label{sec:related}

\textbf{Alpha channel processing in RGBA images.}
Alpha channel processing is closely related to image decomposition~\citep{Akimoto_2020_CVPR,10.1145/3592128,yang2024generative}, layer extraction~\citep{https://doi.org/10.1111/cgf.13577} and image matting~\citep{Xu_2017_CVPR, SEMat, kim2024zim, rethink_p3m, rim}. Traditional approaches in transparent image processing often build on image decomposition, color segmentation, and geometric reasoning in RGB space~\cite{sss}. Recent diffusion-based methods such as DreamLayer~\citep{huang2025dreamlayersimultaneousmultilayergeneration} extend layered generation by modeling inter-layer relationships through attention mechanisms, enabling more coherent multi-layer compositions. In parallel, image matting focuses on estimating alpha matte from natural images, representing fractional foreground occupancy at the pixel level. Semantic Image Matting~\cite{sun2021sim} introduces class-aware matting constraints via semantic trimaps and a multi-class discriminator. Glance and Focus Matting~\cite{li2022bridging} decomposes the task into high-level segmentation and low-level detail refinement, enabling end-to-end matting with improved generalization to real-world images. Referring Image Matting~\cite{rim} extends the problem to natural language-guided alpha extraction and introduces the RefMatte dataset for instruction-based transparency modeling.

\textbf{Transparency-aware representation learning in RGBA images.}
Modeling the alpha channel is essential for both reconstructing and generating RGBA images~\citep{fontanella2024generating, quattrini2024alfie, layerdiffuse, huang2025dreamlayersimultaneousmultilayergeneration}. Recent diffusion-based generation methods have also attempted to produce RGBA images, such as Text2Layer~\citep{zhang2023text2layerlayeredimagegeneration}, which introduces layered architectures for foreground-background separation, but treats transparency implicitly through compositing attention, without explicit alpha modeling objectives. LayerDiffuse~\citep{layerdiffuse} attempts to address this via a bypass VAE that encodes alpha as a latent offset. However, the learned representation resembles coarse matting masks rather than capturing continuous and structured alpha distributions. More recently, the Anonymous Region Transformer (ART)~\citep{pu2025artanonymousregiontransformer} builds upon the Diffusion Transformer (DiT)~\citep{DiT} architecture to enable efficient generation of transparency-aware images with region-level control and variable opacity using an anonymous layout design. However, these methods still fall short in generating high-quality RGBA images with an accurate and coherent alpha channel.

% \begin{figure}[t]
%     \centering
%     \fbox{\parbox{0.9\linewidth}{\centering Placeholder for visualizations of dataset samples and composition charts (e.g., pie charts of domain and type distribution).}}
%     \caption{Overview of datasets used in the evaluation benchmark. We show representative RGBA samples from Alpha and AIM-500, as well as dataset composition summaries (e.g., domain and content-type distributions).}
%     \label{fig:dataset-samples}
% \end{figure}

\section{\benchmark{}: Evaluating RGBA Image Reconstruction and Generation}
\label{sec:benchmark}

We propose \benchmark{}, the first benchmark specifically designed for evaluating four-channel RGBA image reconstruction and generation. By leveraging background blending, \benchmark{} enables the use of well-established three-channel evaluation metrics to assess image quality. To support this benchmark, we provide a set of metrics and a dedicated dataset that jointly evaluate the quality of RGB and alpha channels. Details are presented below.

\subsection{Evaluation Protocol}
\label{subsec:evaluation_protocol}

% \vspace{0.3em}
\textbf{Motivation.}  
Standard image–to–image (i2i) metrics—\emph{e.g.}, PSNR, SSIM~\cite{1284395}, FID~\cite{10.5555/3295222.3295408}, LPIPS~\cite{8578166}, and LAION-AES~\cite{10.5555/3600270.3602103}—are defined for three-channel RGB inputs.  Directly extending them to four-channel RGBA images is problematic: pixels with zero opacity (\(\alpha=0\)) become visually irrelevant, yet naïve \(4\text{-channel}\) formulations still penalize the corresponding RGB values.  We therefore evaluate each image after compositing it onto a set of canonical backgrounds, converting the task back to the well-studied RGB setting while fully respecting transparency.

% \vspace{0.3em}
\textbf{Alpha blending.}  
Given an RGBA image \(x\in\mathbb{R}^{4\times H\times W}\) with RGB part \(x_{\mathrm{rgb}}\) and alpha matte \(x_{\alpha}\in[0,1]\), and an RGB background \(b\in\mathbb{R}^{3\times H\times W}\), alpha blending is defined as
\begin{equation}
\label{eq:alpha_blending}
    \mathcal{A}(x, b) \;=\;
    x_{\mathrm{rgb}}\odot x_{\alpha} \;+\;
    b\odot (1 - x_{\alpha}),
\end{equation}
where \(\odot\) denotes element-wise multiplication\footnote{For simplicity, element-wise multiplication is implied when the operator is omitted, unless stated otherwise.}. 
This operation produces a standard RGB image that faithfully reflects the visual appearance of \(x\) composited over background \(b\).

% \vspace{0.3em}
\textbf{Metric extension.}  
Let \(\mathcal{M}_3\) be any RGB metric,
\(\mathcal{M}_3:\mathbb{R}^{N\times3\times H\times W}\!\times\!\mathbb{R}^{N\times3\times H\times W}\to\mathbb{R}\).
Define its RGBA counterpart \(\mathcal{M}_4\) by averaging \(\mathcal{M}_3\) over a predefined background set \(\mathcal{B}\):
% \begin{equation}
% \label{eq:m4}
%     \mathcal{M}_4(\mathcal{X}, \hat{\mathcal{X}})
%     \;=\;
%     \frac{1}{|\mathcal{B}|}\!
%     \sum_{b\in\mathcal{B}}
%     \mathcal{M}_3\!\bigl(\mathcal{A}(\mathcal{X}, b),\,\mathcal{A}(\hat{\mathcal{X}}, b)\bigr),
% \end{equation}
% \begin{equation}
%     \mathcal{M}_4(\mathcal{X},\mathcal{\hat{X}})
%     = \mathop{\mathbb{E}}\limits_{b\in\mathcal{B}}\biggl[
%         \mathop{\mathbb{E}}\limits_{(x, \hat{x})\in\mathcal{X}\times\hat{\mathcal{X}}}\Bigl[
%             \mathcal{M}_3\Bigl(\mathcal{A}(x, b),\mathcal{A}(\hat{x},b)\Bigr)
%         \Bigr]
%     \biggr]
% \end{equation}
% \begin{equation}
%     \mathcal{M}_4(\mathcal{X},\mathcal{\hat{X}})
%     = \mathop{\mathbb{E}}\limits_{b\in\mathcal{B}}\biggl[
%     \dfrac{1}{|\mathcal{X}\times\hat{\mathcal{X}}|}
%     \sum_{(x, \hat{x})\in\mathcal{X}\times\hat{\mathcal{X}}}
%         % \mathop{\mathbb{E}}\limits_{(x, \hat{x})\in\mathcal{X}\times\hat{\mathcal{X}}}
%         \Bigl[
%             \mathcal{M}_3\Bigl(\mathcal{A}(x, b),\mathcal{A}(\hat{x},b)\Bigr)
%         \Bigr]
%     \biggr]
% \end{equation}
\begin{equation}
\label{eq:m4}
    \mathcal{M}_4(\mathcal{X},\mathcal{\hat{X}})
    = 
    \frac{1}{|\mathcal{B}|}\!
    \sum_{b\in\mathcal{B}}
    % \mathop{\mathbb{E}}\limits_{b\in\mathcal{B}}\biggl[
        \mathop{\mathbb{E}}\limits_{(x, \hat{x})\in\mathcal{X}\times\hat{\mathcal{X}}}\Bigl[
        % \mathbb{E}_{(x, \hat{x})\in\mathcal{X}\times\hat{\mathcal{X}}}\Bigl[
            \mathcal{M}_3\Bigl(\mathcal{A}(x, b),\mathcal{A}(\hat{x},b)\Bigr)
        \Bigr]
    % \biggr]
\end{equation}
where \(\mathcal{X}\) and \(\hat{\mathcal{X}}\) denote the ground-truth and reconstructed sets, respectively.
% and \(\mathcal{A}\) is applied element-wise.

% \vspace{0.3em}
\textbf{Background set.}  
We adopt nine solid colours that span the pixel \(\{0,0.5,1\}^3\):  
\texttt{black}, \texttt{gray}, \texttt{white}, \texttt{red}, \texttt{green}, \texttt{blue}, \texttt{yellow}, \texttt{cyan}, and \texttt{magenta}.  Using low-frequency, texture-free backgrounds eliminates semantic bias and yields numerically stable values for all metrics.

% \vspace{0.3em}
\textbf{Reported scores.}  
Final scores are reported as the average across all background colors. This pipeline allows robust evaluation of both RGB fidelity and transparency-aware perceptual quality, while remaining compatible with existing metrics.

\subsection{Dataset Collection}
\textbf{Datasets preparation.}
High-quality RGBA training data is scarce due to the difficulty of acquiring accurate alpha matte and consistent foreground transparency in the wild. However, we observe that existing high-quality image matting datasets, originally designed for alpha prediction, can be effectively repurposed for RGBA generation. These datasets typically contain paired foreground and alpha-mask images extracted from real-world compositions. To convert such data into RGBA format, we simply combine the RGB foreground $I_\mathrm{fg}$ and the corresponding alpha matte $\alpha$ to form a four-channel image $x = \texttt{concat}(I_\mathrm{fg}, \alpha)$.
This approach ensures that the resulting RGBA images are aligned, photorealistic, and contain diverse transparency patterns. We choose \textbf{8,124} high-quality images, from ten image matting datasets, including Adobe Image Matting dataset~\cite{Xu_2017_CVPR}, AM-2K~\cite{li2022bridging}, Distinctions-646~\cite{Qiao_2020_CVPR}, HHM-2K~\cite{Sun_2023_CVPR}, Human-1K~\cite{Liu_2021_ICCV}, P3M-500-NP~\cite{10.1145/3474085.3475512}, PhotoMatte85~\cite{Lin_2021_CVPR}, realWorldPortrait-636~\cite{yu2020mask}, SIMD~\cite{sun2021sim} and Transparent-460~\cite{cai2022TransMatting}, covering a broad range of object categories and transparency patterns. All images are preprocessed into four-channel RGBA format by compositing foregrounds with their corresponding alpha mattes. By leveraging these matting datasets, we eliminate the need for manual labeling and enable scalable, high-quality training for RGBA reconstruction and generation tasks.

\textbf{Train/Test split.}
For each dataset, we follow a standardized protocol to reserve a small portion for testing during training. Specifically, we select the test subset as 5\% of the dataset size.
% \[
% \texttt{val\_size} = \lfloor 0.05 \times N \rfloor
% \]
% where \( N \) denotes the total number of samples in a dataset. 
This ensures sufficient coverage for testing even in smaller datasets, while preserving training diversity. The final split yields \textbf{7,722} training images and \textbf{402} testing images.

\textbf{Overview.}
We present the dataset statistics for training and evaluation in Table~\ref{tab:dataset-stats}. The training set, \benchmark{} train split, comprises 7,722 RGBA images spanning diverse domains, with an emphasis on challenging transparency effects such as hair, glass, and soft shadows. The evaluation set, \benchmark{} test split, includes 402 RGBA images exhibiting similar complexities. This comprehensive setup allows us to assess the adaptability of AlphaGen across varied transparent image generation scenarios.

% \subsection{Dataset Statistics}Å

% AlphaSet-8K consists of 8,124 high-quality RGBA images sourced from ten matting datasets, 

\begin{table}[t]
\centering
\caption{Statistics of each dataset in \benchmark{}. Resolution indicates the average image dimensions in each dataset, computed as the mean height multiplied by the mean width of all images.}
\label{tab:dataset-stats}
\vspace{.4em}
\begin{tabular}{lccc}
\toprule
\textbf{Dataset} & \textbf{\# Train Images} & \textbf{\# Test Images} & \textbf{Resolution} \\
\midrule
Adobe Image Matting dataset & 449 & 23    & 1292$\times$1082 \\
AM-2K                       & 1,900 & 100   & 1471$\times$1195 \\
Distinctions-646            & 607  & 31  & 1569$\times$1732 \\
HHM-2K                      & 1,900 & 100  & 3570$\times$4041 \\
Human-1K                    & 953 & 50  & 2060$\times$2094 \\
P3M-500-NP                  & 475  & 25  & 1374$\times$1313 \\
PhotoMatte85                & 81  & 4   & 2304$\times$3456 \\
realWorldPortrait-636       & 605  & 31  & 1038$\times$1327  \\
SIMD                        & 318  & 16  & 2346$\times$2079 \\
Transparent-460             & 434  & 22  & 3799$\times$3767 \\
\midrule
\textbf{Total} & 7,722 & 402 & 2176$\times$2240 \\
\bottomrule
\end{tabular}
\end{table}

% \begin{table}[htbp]
% \centering
% \caption{Statistics of training and testing datasets used in AlphaGen.}
% \label{tab:dataset-stats}
% \begin{tabular}{lccc}
% \toprule
% \textbf{Dataset} & \textbf{Split} & \textbf{\# Images} & \textbf{Resolution} \\
% \midrule
% \multirow{2}{*}{AlphaSet-8K} & Train & 7,722  & 2175$\times$2239 \\
%                              & Test  & 402    & 2194$\times$2266 \\
% \cmidrule(lr){1-4}
% AIM-500                      & Test  & 500    & 1397$\times$1260 \\
% \bottomrule
% \end{tabular}
% \end{table}

\section{\method{}: Teaching VAE to Reconstruct RGBA Images}
\label{sec:method}

\begin{figure}[t]
    \centering
    \includegraphics[width=\linewidth]{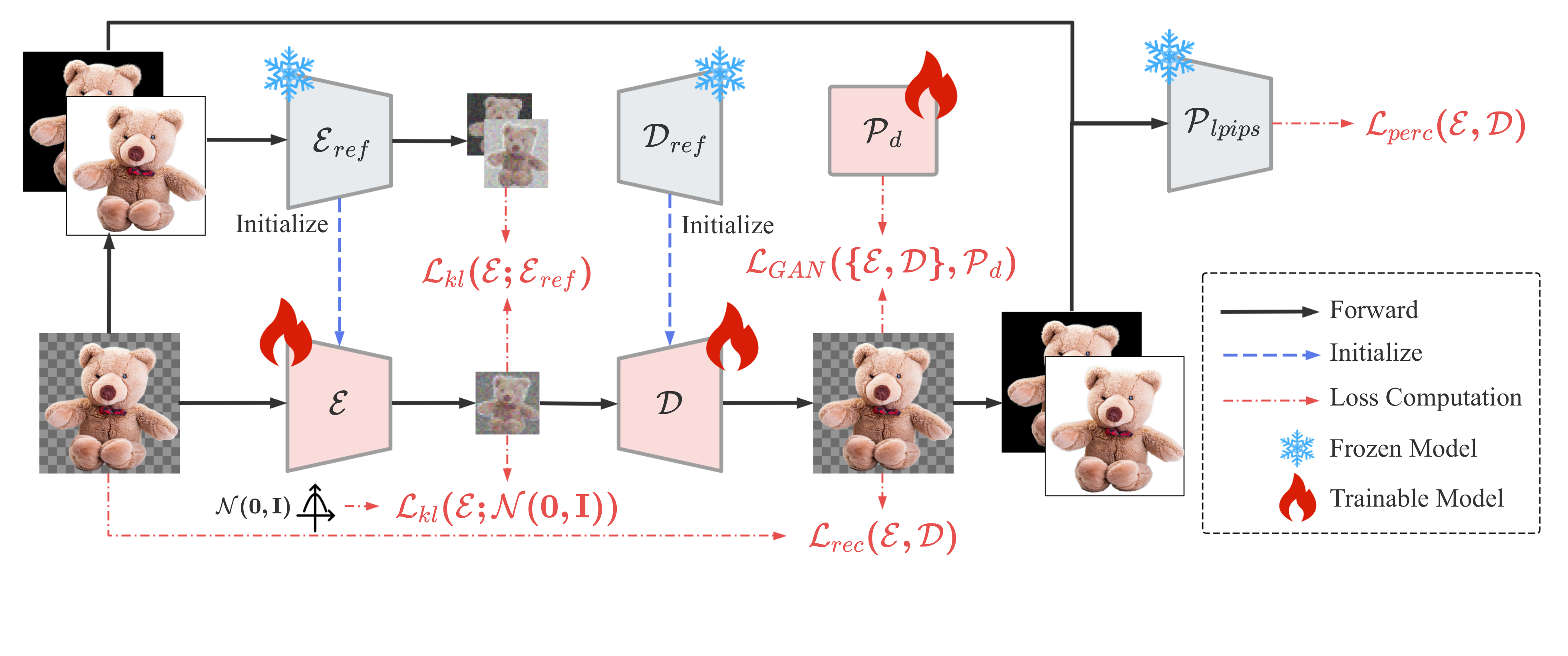}
    \vspace{-3.8em}
    \caption{Training pipeline of \method{}.}
    \label{fig:pipeline}
\end{figure}

\subsection{Model Architecture and Initialization}

Variational Autoencoders (VAEs) encode high‑dimensional inputs into a structured latent space and then decode this representation to reconstruct the original data. The encoder  
\begin{equation}
\mathcal{E}:\;\mathbb{R}^{C \times H \times W}\;\longrightarrow\;\mathbb{R}^{D \times h \times w}
\end{equation}
maps an image \(x\in\mathbb{R}^{C \times H \times W}\) to a distribution in latent space, where the dimensionality reduction
\(
D\,h\,w < C\,H\,W
\)
encourages the model to capture salient, abstract features in compact form.  
The decoder  
\begin{equation}
\mathcal{D}:\;\mathbb{R}^{D \times h \times w}\;\longrightarrow\;\mathbb{R}^{C \times H \times W}
\end{equation}
reconstructs the input from its latent code, effectively up‑sampling and re‑rendering the encoded information.

In the RGBA setting, \(x\in\mathbb{R}^{4\times H\times W}\) comprises an RGB part \(x_{\mathrm{rgb}}\) and an alpha (transparency) channel \(x_{\alpha}\). To reuse a pre‑trained RGB (three‑channel) VAE while supporting four channels, we extend the first convolution of \(\mathcal{E}\) and the final convolution of \(\mathcal{D}\) from three to four channels.

Let \(W^{\mathcal{E}}_{0}\in\mathbb{R}^{k\times k \times 4 \times D_{0}}\) and \(b^{\mathcal{E}}_{0}\in\mathbb{R}^{D_{0}}\) denote the weight and bias of the first convolution in \(\mathcal{E}\), and let \(W^{\mathcal{D}}_{L}\in\mathbb{R}^{k\times k \times D_{L} \times 4}\) and \(b^{\mathcal{D}}_{L}\in\mathbb{R}^{4}\) denote the weight and bias of the last convolution in \(\mathcal{D}\).  
We copy the pre‑trained RGB parameters into the first three channel slices and initialize the newly added alpha‑channel slices as
\[
W^{\mathcal{E}}_{0}[:,:,4,:]=\mathbf{0}, \qquad
W^{\mathcal{D}}_{L}[:,:,:,4]=\mathbf{0}, \qquad
b^{\mathcal{D}}_{L}[4]=\mathbf{1},
\]
while keeping \(b^{\mathcal{E}}_{0}\) and the first three components of \(b^{\mathcal{D}}_{L}\) unchanged.  
This simple yet effective initialization preserves the rich feature extraction learned from RGB images and lets the network gradually learn to exploit the transparency channel during fine‑tuning. The pipeline is illustrated in Figure~\ref{fig:pipeline}.

\subsection{Training Objective}

We train an RGBA VAE with a composite loss function that includes the following terms.

\subsubsection{Reconstruction Loss}

In image reconstruction tasks, a common approach is to penalize the difference between the original image $x$ and its reconstruction $\hat{x} = \mathcal{D}(\mathcal{E}(x))$, where $\mathcal{E}$ is an encoder and $\mathcal{D}$ is a decoder.
A straightforward reconstruction loss would be the squared L2 norm, which calculates the mean squared error (MSE) between the reconstructed and original image tensors:
\begin{equation}
    \mathcal{L}_{rec}(\mathcal{E}, \mathcal{D}) = \left\lVert \hat{x} - x \right\rVert_2^2
\end{equation}
As described in Section~\ref{subsec:evaluation_protocol}, the ideal difference should be the RGB difference after blending a background.
Thus, our proposed reconstruction loss, $\mathcal{L}_{rec}$, is the expected L2 difference between the alpha-blended reconstruction and the alpha-blended original image, which we call Alpha-Blending Mean Square Error (ABMSE), taken over a distribution of backgrounds $b$:
\begin{equation}
    \mathcal{L}_{rec}(\mathcal{E}, \mathcal{D}) = \mathop{\mathbb{E}}\limits_{b}\left\lVert \mathcal{A}(\hat{x}, b) - \mathcal{A}(x, b) \right\rVert_2^2
\end{equation}
% By substituting the definition of $\mathcal{A}$ and expanding the squared L2 norm, this loss can be decomposed.
The expression inside the expectation becomes:
\begin{equation}
    \mathcal{A}(\hat{x}, b) - \mathcal{A}(x, b) = (\hat{x}_{rgb} \hat{x}_{\alpha} - x_{rgb} x_{\alpha}) - b (\hat{x}_{\alpha} - x_{\alpha})
\end{equation}
To simplify the calculations, each pixel in $b$ is considered to be independent and identically distributed (i.i.d.).
Let $P = \hat{x}_{rgb} \hat{x}_{\alpha} - x_{rgb} x_{\alpha}$ (the difference in premultiplied RGB values) and $\Delta_{\alpha} = \hat{x}_{\alpha} - x_{\alpha}$ (the difference in alpha values).
Then, $\left\lVert (\hat{x}_{rgb} \hat{x}_{\alpha} - x_{rgb} x_{\alpha}) - b (\hat{x}_{\alpha} - x_{\alpha}) \right\rVert_2^2 = \left\lVert P - b \Delta_{\alpha} \right\rVert_2^2$.
Expanding\footnote{The detailed deduction is displayed in Appendix~\ref{app:formula}.} this squared norm and taking the expectation with respect to $b$:
\begin{equation}
\label{eq:reconstruction_loss}
\mathcal{L}_{rec} (\mathcal{E}, \mathcal{D})
= \mathop{\mathbb{E}}\limits_{b}\left[ \left\lVert P - b \Delta_{\alpha} \right\rVert_2^2 \right]
% = \left\lVert P \right\rVert_2^2
% - 2 \mathbb{E}[b]\cdot P \cdot \Delta_{\alpha} 
% +  \mathbb{E}[b^2] \cdot\Delta_{\alpha}^2 
= \left\lVert P \right\rVert_2^2
  - 2\Delta_\alpha \left\langle \mathbb{E}[b], P \right\rangle 
  + \Delta_\alpha^2 \left\lVert \mathbb{E}[b^2] \right\rVert_1
\end{equation} 
Thus, $\mathbb{E}[b]$ and $\mathbb{E}[b^2]$ can be pre-calculated. 
In practice, the distribution of $b$ is estimated from the ImageNet training split.  
Detailed statistics of $b$ are provided in Appendix~\ref{app:params}.

% However, the fact is that the color channels where $\alpha=0$ are invisible. In other words, if the alpha channel equals 0, the RGB part should not be taken consideration.
% Thus, we design a novel reconstruction loss considering the characteristic of alpha channel. 
% We define the loss term as the expectation of the L2 loss with alpha blending over different backgrounds $b\in \mathbb{R}^{4\times H\times W}$. Formally,
% \begin{equation}
%     \mathcal{L}_{rec} = \mathop{\mathbb{E}}\limits_{b}\left\lVert
%     \mathcal{A}(\mathcal{D}\circ\mathcal{E}(x),b) - \mathcal{A}(x,b)
%     \right\rVert_2^2
%     \quad \text{, where\;\;} \mathcal{A}(x,b) = x_{rgb}\cdot x_{\alpha} + b\cdot(1-x_{\alpha})
% \end{equation}
% where for $x\in \mathbb{R}^{4\times H\times W}$ and $b\in\mathbb{R}^{3\times H\times W}$, $x_{rgb}$ and $x_{\alpha}$ represents the first 3 dimensions and the last dimension of $x$.
% By decomposing the fomula, we gain:
% \begin{align}
%     \mathcal{L}_{rec} = 
%     \left\lVert \hat{x}_{rgb}\cdot \hat{x}_{\alpha} - x_{rgb}\cdot x_{\alpha} \right\rVert_2^2
%     - 2\;\mathbb{E}[b]\cdot(\hat{x}_{rgb}\cdot \hat{x}_{\alpha}- x_{rgb}\cdot x_{\alpha})(\hat{x}_{\alpha}-x_{\alpha})
%     + \mathbb{E}[b^2]\cdot\left\lVert \hat{x}_{\alpha}-x_{\alpha}\right\rVert_2^2
% \end{align}
% The distribution of $b$ are calculated by ImageNet train split.

% Red Channel - Mean: -0.0357, Second Moment: 0.3163
% Green Channel - Mean: -0.0811, Second Moment: 0.3060
% Blue Channel - Mean: -0.1797, Second Moment: 0.3634

\subsubsection{Perceptual Loss}

In VQVAE and SD, perceptual loss is calculated by a pretrained model $\mathcal{P}_{lpips}$:
\begin{equation}
    \mathcal{L}_{perc}(\mathcal{E}, \mathcal{D}) = \mathcal{P}_{lpips}(\hat{x}, x)
\end{equation}
However, this model receives three-channel images.
Thus, before feeding a image into $\mathcal{P}_{lpips}$, we also perform alpha blending on it, i.e., $\mathcal{P}_{lpips}(\mathcal{A}(\hat{x},b), \mathcal{A}(x,b))$. 
Note that the model $\mathcal{P}_{lpips}$ can not be simply represented by a polynomial which can simplifies the function a composition of $x$, $\hat{x}$, and $\mathbb{E}\left[b^{k}\right]$.
Finding that the distribution of channel value is concentrated at $0$ and $1$, we apply white-colored and black-colored images as the background, i.e.,
\begin{align}
    \mathcal{L}_{perc}(\mathcal{E}, \mathcal{D})
    & = \mathop{\mathbb{E}}\limits_{b\in\{\mathbf{0}, \mathbf{1}\}}\left[\mathcal{P}_{lpips}(\mathcal{A}(\hat{x},b), \mathcal{A}(x,b))\right] \\
    & = \frac{1}{2}\left(
    \mathcal{P}_{lpips}(\mathcal{A}(\hat{x},\mathbf{0}), \mathcal{A}(x,\mathbf{0}))
    +
    \mathcal{P}_{lpips}(\mathcal{A}(\hat{x},\mathbf{1}), \mathcal{A}(x,\mathbf{1}))
    \right)
\end{align}

\subsubsection{Regularization Loss}

% In VAE, a KL term is applied to regulate the distribution of $\mathcal{E}(\;\cdot\;)$. In most cases, the reference distribution is a standard Gaussian distribution $\mathcal{N}(\mathbf{0},\mathbf{I})$. 
% Besides, because we fine-tune the four-channel VAE $\mathcal{E}$ from a pre-trained three-channel VAE $\mathcal{E}_{ref}$, we aim to regulate the latent distribution simultaneously, such that we can apply our fine-tuned VAE into the diffusion model with a few steps of fine-tuning. Thus, we add another KL regulation term where the reference distribution is the raw VAE. 
% Note that the reference VAE takes 3-channel images as inputs, so we also blend the 4-channel image to feed into the reference VAE.
% Formally:

In our Variational Autoencoder (VAE) framework, we incorporate a regularization loss to govern the characteristics of the latent space distribution generated by the encoder, denoted as $\mathcal{E}(\cdot)$. The regularization is achieved through two distinct Kullback-Leibler (KL) divergence terms.

The first KL divergence term, $\mathcal{L}_{kl}(\mathcal{N}(\mathbf{0},\mathbf{I}))$, enforces a standard regularization on the encoder. It encourages the posterior distribution of the latent variables $z$ given an input $x$, $q(z|x)$ (as approximated by the encoder $\mathcal{E}(z|x)$), to conform to a prior, which is a standard multivariate Gaussian distribution $\mathcal{N}(z;\mathbf{0},\mathbf{I})$. This is a common practice in VAEs to ensure a well-structured and continuous latent space. The formal definition is:
\begin{equation}
    \mathcal{L}_{kl}(\mathcal{E};\mathcal{N}(0,\mathbf{I})) 
    = \mathop{\mathbb{E}}\limits_{x}
    \left[
    KL\left(\mathcal{E}(z|x)||\mathcal{N}(z;\mathbf{0},\mathbf{I}) \right)
    \right]
\end{equation}
The second KL divergence term, $\mathcal{L}_{kl}(\mathcal{E}_{ref})$, is specifically introduced due to our fine-tuning strategy. Our current four-channel VAE encoder, $\mathcal{E}$, is adapted from a pre-trained three-channel reference VAE encoder, $\mathcal{E}_{ref}$. To ensure that the latent distribution of our fine-tuned encoder $\mathcal{E}$ remains compatible with that of $\mathcal{E}_{ref}$, thereby facilitating its integration into a diffusion model with minimal subsequent fine-tuning, we introduce this additional constraint. This term minimizes the KL divergence between the latent distribution produced by our encoder $\mathcal{E}$ and that produced by the reference encoder $\mathcal{E}_{ref}$.

A critical consideration is the input channel mismatch: $\mathcal{E}$ processes four-channel images, while $\mathcal{E}_{ref}$ expects three-channel images. To address this, we also employ the blending function $\mathcal{A}(x,b)$. For the input to our fine-tuned encoder $\mathcal{E}$ in this context, the blended three-channel image $\mathcal{A}(x,b)$ is concatenated with a fourth channel consistently set to unity (i.e., $[\mathcal{A}(x,b);1]$). The loss is formulated as:
\begin{equation}
    \mathcal{L}_{kl}(\mathcal{E};\mathcal{E}_{ref}) 
    = \mathop{\mathbb{E}}\limits_{x}
    \Biggl[
    \mathop{\mathbb{E}}\limits_{b\in\{\mathbf{0}, \mathbf{1}\}}
        \biggl[
        KL\biggl(
            \mathcal{E}\Bigl(z\Bigl|[\mathcal{A}(x,b);1]\Bigr)\biggl|\biggl|
            \mathcal{E}_{ref}\Bigl(z\Bigl|\mathcal{A}(x,b)\Bigr)
        \biggr)
        \biggr]
     \Biggr]
\end{equation}

% \begin{equation}
%     \mathcal{L}_{kl}(p) = KL(\mathcal{E}||p)
% \end{equation}

% \begin{equation}
%     \mathcal{L}_{kl}(\mathcal{N}(0,\mathbf{I})) = KL(\mathcal{E}(\;\cdot\;)||\mathcal{N}(\;\cdot\;;\mathbf{0},\mathbf{I}))
% \end{equation}

% \begin{equation}
%     \mathcal{L}_{kl}(\mathcal{E}_{ref}) = KL(\mathcal{E}(\;\cdot\;)||\mathcal{E}_{ref}(\;\cdot\;))
% \end{equation}

\subsubsection{GAN Loss}

In ~\cite{esser2021taming, 9878449}, a patch-based discriminator $\mathcal{P}_{d}$ is usually applied to learn a rich representation.
\begin{equation}
    \mathcal{L}_{GAN}(\{\mathcal{E}, \mathcal{D}\}, \mathcal{P}_{d}) = \log{\mathcal{P}_{d}(x)} + \log(1-\mathcal{P}_{d}(\hat{x}))
\end{equation}
Following VQGAN~\cite{esser2021taming}, we also compute the adaptive weight $\lambda_{adapt}$ according to
\begin{equation}
    \lambda_{adapt} = \frac{\nabla_{\mathcal{D}_L}[\mathcal{L}_{rec}]}{\nabla_{\mathcal{D}_L}[\mathcal{L}_{GAN}] + \epsilon}
\end{equation}
where $\mathcal{D}_L$ represents the last layer of $\mathcal{D}$, and $\epsilon = 10^{-4}$ ensures numerical stability.

Thus, the final objective is:
\begin{equation}
\begin{split}
    \mathcal{E}^*, \mathcal{D}^* 
    = \mathop{\arg\min}\limits_{\mathcal{E}, \mathcal{D}}
    \max\limits_{\mathcal{P}_{d}}
    \Bigl[
        \mathcal{L}_{rec} (\mathcal{E}, \mathcal{D}) 
        + \lambda_{perc}\mathcal{L}_{perc} (\mathcal{E}, \mathcal{D}) 
        + \lambda_{norm}\mathcal{L}_{kl}(\mathcal{E};\mathcal{N}(0,\mathbf{I})) \\
        + \lambda_{ref}\mathcal{L}_{kl}(\mathcal{E};\mathcal{E}_{ref}) 
        + \lambda_{GAN}\lambda_{adapt}\mathcal{L}_{GAN}(\{\mathcal{E}, \mathcal{D}\}, \mathcal{P}_{d})
    \Bigr]
\end{split}
\end{equation}

\definecolor{deepgreen}{HTML}{006400} % Dark green
\definecolor{mypurple}{HTML}{800080}  % Purple
\definecolor{black}{HTML}{000000}  % Purple

% Custom command for value over colored percentage
% #1: Metric value (potentially bolded/underlined)
% #2: Color for the percentage (e.g., deepgreen, mypurple)
% #3: Percentage string (e.g., +10.50\%)
\newcommand{\valueoverpercentage}[3]{\makecell{#1\\[-0.3em] \scriptsize{\textcolor{#2}{(#3)}}}}

\begin{table}[t]
  \centering
  \caption{VAE reconstruction results on AIM-500 and \benchmark{} test split. $\uparrow$ indicates higher values are better, and $\downarrow$ indicates lower values are better. \textbf{Best} and \underline{second-best} values per metric are indicated in bold and by underlining, respectively. Differences relative to the LayerDiffuse baseline are colored \textcolor{deepgreen}{green} for improvement. These formatting conventions are maintained for all subsequent result tables.}
  \vspace{.4em}
  \label{tab:vae-quantitative-results}
  \resizebox{\textwidth}{!}{%
  \renewcommand{\arraystretch}{1.2} % 增加行高
  \begin{tabular}{c c ccccc ccccc}
    \toprule
    \multirow{3}{*}{\textbf{Method}} & \multirow{3}{*}{\textbf{Base Model}} & \multicolumn{5}{c}{\textbf{AIM}} & \multicolumn{5}{c}{\textbf{Alpha}} \\
    \cmidrule(lr){3-7} \cmidrule(lr){8-12}
    & & PSNR $\uparrow$ & SSIM $\uparrow$ & \makecell{LAION\\AES} $\uparrow$ & rFID $\downarrow$ & LPIPS $\downarrow$
      & PSNR $\uparrow$ & SSIM $\uparrow$ & \makecell{LAION\\AES} $\uparrow$ & rFID $\downarrow$ & LPIPS $\downarrow$ \\
    \midrule
    LayerDiffuse & SDXL & 32.0879 & 0.9436 & 4.9208 & 17.7023 & \uline{0.0418} & 32.4531 & 0.9473 & 5.0306 & 6.4832 & \uline{0.0324} \\
    \midrule
    \multirow{2.5}{*}{\makecell[c]{AlphaVAE\\(Ours)}} & SDXL & \valueoverpercentage{\uline{35.7446}}{deepgreen}{+3.6567} & \valueoverpercentage{\uline{0.9576}}{deepgreen}{+0.0140} & \valueoverpercentage{\uline{4.9456}}{deepgreen}{+0.0248} & \valueoverpercentage{\textbf{10.9178}}{deepgreen}{-6.7845} & \valueoverpercentage{0.0495}{mypurple}{+0.0077} & \valueoverpercentage{\uline{35.5597}}{deepgreen}{+3.1066} & \valueoverpercentage{\uline{0.9605}}{deepgreen}{+0.0132} & \valueoverpercentage{\uline{5.0500}}{deepgreen}{+0.0194} & \valueoverpercentage{\uline{3.7305}}{deepgreen}{-2.7527} & \valueoverpercentage{0.0402}{mypurple}{+0.0078} \\ 
     & FLUX & \valueoverpercentage{\textbf{36.9439}}{deepgreen}{+4.8560} & \valueoverpercentage{\textbf{0.9737}}{deepgreen}{+0.0301} & \valueoverpercentage{\textbf{4.9511}}{deepgreen}{+0.0303} & \valueoverpercentage{\uline{11.7884}}{deepgreen}{-5.9139} & \valueoverpercentage{\textbf{0.0283}}{deepgreen}{-0.0135} & \valueoverpercentage{\textbf{38.1987}}{deepgreen}{+5.7456} & \valueoverpercentage{\textbf{0.9792}}{deepgreen}{+0.0319} & \valueoverpercentage{\textbf{5.0616}}{deepgreen}{+0.0310} & \valueoverpercentage{\textbf{1.6600}}{deepgreen}{-4.8232} & \valueoverpercentage{\textbf{0.0145}}{deepgreen}{-0.0179} \\ 
    \bottomrule
\end{tabular}}
\end{table}

\section{Experiments}
\label{sec:exp}

\subsection{Implementation Details}
We train VAE and diffusion model on the same dataset, where image-caption pairs are constructed using Qwen-VL-2.5~\cite{Qwen2.5-VL} as the captioning model. Detailed configurations for VAE and diffusion model are provided in the following subsections.
All experiments are conducted on 4 NVIDIA H800 GPUs.

% \paragraph{VAE} 
\textbf{VAE.}
We fine-tune the VAE components of both SDXL~\cite{podell2023sdxl} and FLUX.1[dev]~\cite{flux2024} models. As a form of data augmentation, each input transparent image is blended with a randomly generated solid-color background with a probability of 0.3 during training. Full-parameter fine-tuning is performed for 30,000 iterations using a global batch size of 8. We employed the AdamW optimizer with a learning rate of $1 \times 10^{-5}$.

\textbf{Diffusion.} 
After training the VAE, we fine-tune the 
% Transformer module in FLUX.1[dev]~\cite{flux2024} 
corresponding latent diffusion module
via Low-Rank Adaptation (LoRA) with a rank of 64. The model is optimized using the Prodigy optimizer at a learning rate of 1.0, and trained for 20,000 iterations with a global batch size of 8.

% \begin{table}[htbp]
%   \centering
%   \caption{Generation results on on AIM-500 and \benchmark{} test split.}
%   \label{tab:generation-quantitative-results}
%   % \setlength{\tabcolsep}{0em} % 增加列间距
%   % \renewcommand{\arraystretch}{1.1} % 增加行高
%   \begin{tabular}{cccc}
%     \toprule
%     \multirow{2.4}{*}{\textbf{Method}} & \multirow{2.4}{*}{\textbf{Setting}} & \multicolumn{2}{c}{FID $\downarrow$} \\
%     \cmidrule(lr){3-4}
%                  &                     & AIM & Alpha \\
%     \midrule
%     LayerDiffuse & - & \uline{160.2664} & 80.2703 \\
%     \midrule
%     \multirow{3.5}{*}{\makecell[c]{AlphaGen\\(Ours)}} & main & \valueoverpercentage{\textbf{155.6606}}{deepgreen}{-4.6058} & \valueoverpercentage{\textbf{74.1224}}{deepgreen}{-6.1479} \\
%     \cmidrule(lr){2-4}
%      & w/o Ref KL & \valueoverpercentage{161.4522}{mypurple}{+1.1858} & \valueoverpercentage{\uline{74.5844}}{deepgreen}{-5.6859} \\
%     \bottomrule
%   \end{tabular}
% \end{table}

\subsection{Quantitative Results}

% \textbf{VAE.} 
% We conduct quantitative evaluations of \method{} using the metrics proposed in \benchmark{}. Specifically, we evaluate the VAEs in both SDXL and FLUX models on the \benchmark{} test split and the AIM-500 dataset. As shown in Table~\ref{tab:vae-quantitative-results}, our method significantly outperforms LayerDiffuse~\cite{layerdiffuse} in terms of PSNR and rFID. In particular, the substantial improvement in rFID on the AIM-500 dataset highlights the strong generalizability of our approach.

\textbf{VAE.} 
We conduct quantitative evaluations of \method{} using the metrics proposed in \benchmark{}. Specifically, we evaluate the VAEs in both SDXL and FLUX models on the \benchmark{} test split and the AIM-500 dataset. As shown in Table~\ref{tab:vae-quantitative-results}, our method significantly outperforms LayerDiffuse~\cite{layerdiffuse} in terms of PSNR and rFID across both datasets. On AIM-500, \method{} improves PSNR from 32.09 to 35.74 (SDXL) and 36.94 (FLUX), yielding absolute gains of +3.66 and +4.86 dB, respectively. The rFID also drops sharply from 17.70 to 10.92 (SDXL) and 11.79 (FLUX), indicating improved perceptual realism. Similar trends are observed on the Alpha benchmark, where PSNR increases by +3.11 and +5.75 dB, and rFID decreases from 6.48 to as low as 1.66. These improvements demonstrate the robustness and generalizability of our proposed VAE framework, with consistently superior reconstruction quality under multiple metrics.

\textbf{Diffusion.}
% We evaluate the FID metric of our diffusion model on the \benchmark{} test split and the AIM-500 dataset to measure the distributional discrepancy between generated and real images. The results indicate that our generation model exhibits strong generalization capability.
We evaluate the FID metric of our diffusion model on the \benchmark{} test split and the AIM-500 dataset to measure the distributional discrepancy between generated and real images.  
As shown in Table~\ref{tab:generation-quantitative-results}, the results indicate that our generation model exhibits strong generalization capability.  
Specifically, on AIM-500, our method achieves an FID of 155.66, improving by 4.61 points compared to LayerDiffuse (160.27).  
On the \benchmark{} split, we attain a FID of 74.12, an improvement (reduction) of 6.15 points from the 80.27 baseline.

\begin{table}[ht]
  \centering
  \caption{Generation results on AIM-500 and \benchmark{} test split.}
  \label{tab:generation-quantitative-results}
\resizebox{.9\textwidth}{!}{

  \begin{tabular}{ccccccc}
    \toprule
    \multirow{2.5}{*}{\textbf{Setting}} & \multicolumn{3}{c}{\textbf{AIM}} & \multicolumn{3}{c}{\textbf{Alpha}} \\
    \cmidrule(lr){2-4} \cmidrule(lr){5-7}
                 & LayerDiffuse         & Ours                                                            & w/o Ref KL                                                     & LayerDiffuse & Ours                                                            & w/o Ref KL                                                      \\
    \midrule
    \textbf{FID} $\downarrow$    & \uline{160.2664}     & \valueoverpercentage{\textbf{155.6606}}{deepgreen}{-4.6058}    & \valueoverpercentage{161.4522}{mypurple}{+1.1858}             & 80.2703      & \valueoverpercentage{\textbf{74.1224}}{deepgreen}{-6.1479}     & \valueoverpercentage{\uline{74.5844}}{deepgreen}{-5.6859}      \\
    \bottomrule
  \end{tabular}

}
\end{table}

\subsection{Qualitative Results}
\begin{figure}[t]
  \centering

  % -------- First subfigure --------
  \begin{subfigure}[t]{0.5\linewidth}
    \centering
    \includegraphics[width=\linewidth]{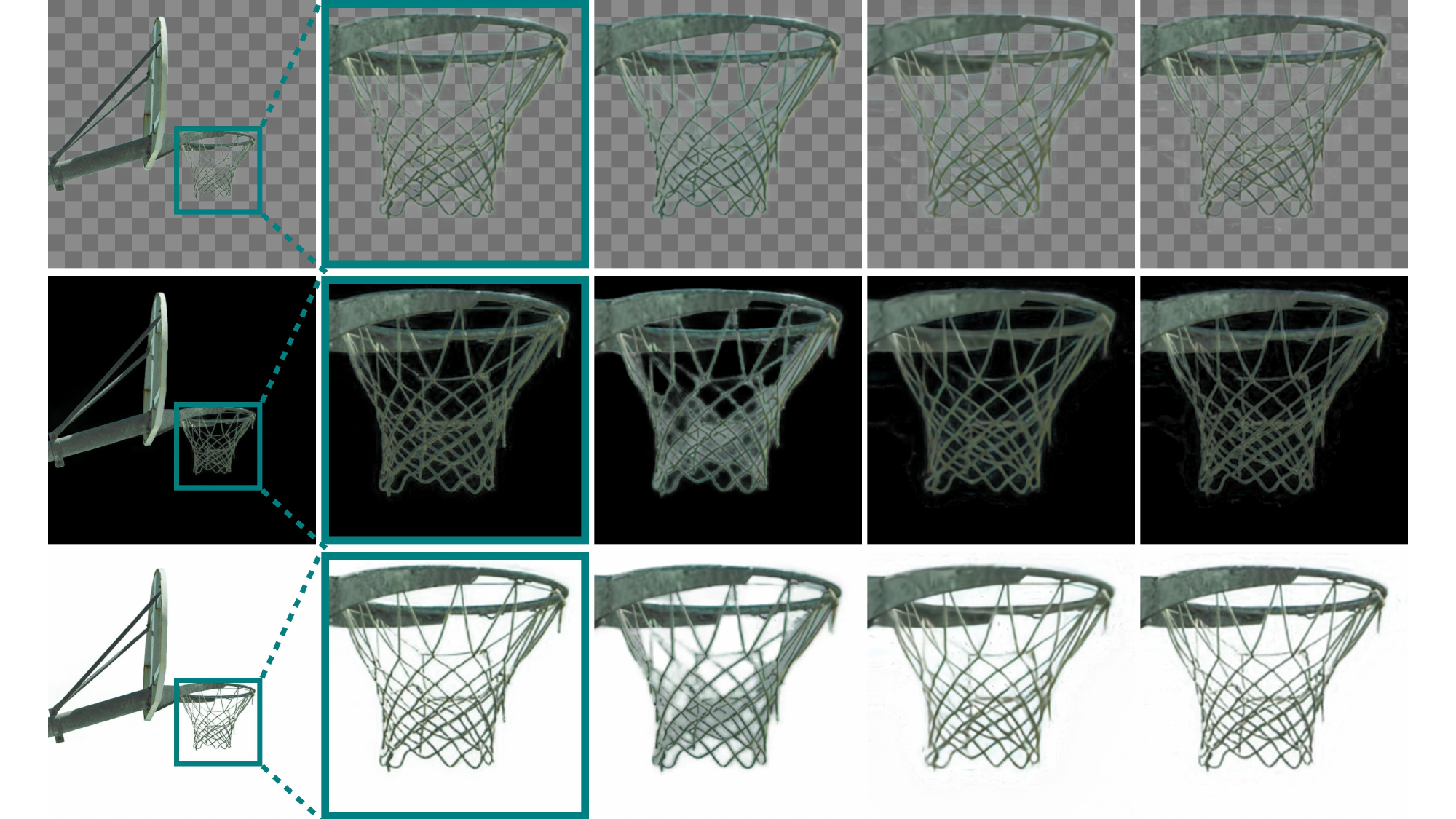}
    \vspace{1mm}
    \adjustbox{max width=\linewidth}{
      \begin{tabularx}{\linewidth}{XXXXX}
        \centering \makebox[0.18\linewidth]{GT} &
        \centering \makebox[0.18\linewidth]{GT*} &
        \centering \makebox[0.18\linewidth]{LD} &
        \centering \makebox[0.18\linewidth]{Ours*} &
        \centering \makebox[0.18\linewidth]{Ours}
      \end{tabularx}
    }
  \end{subfigure} \hspace{-1.5em}
  % \hfill
  % -------- Second subfigure --------
  \begin{subfigure}[t]{0.5\linewidth}
    \centering
    \includegraphics[width=\linewidth]{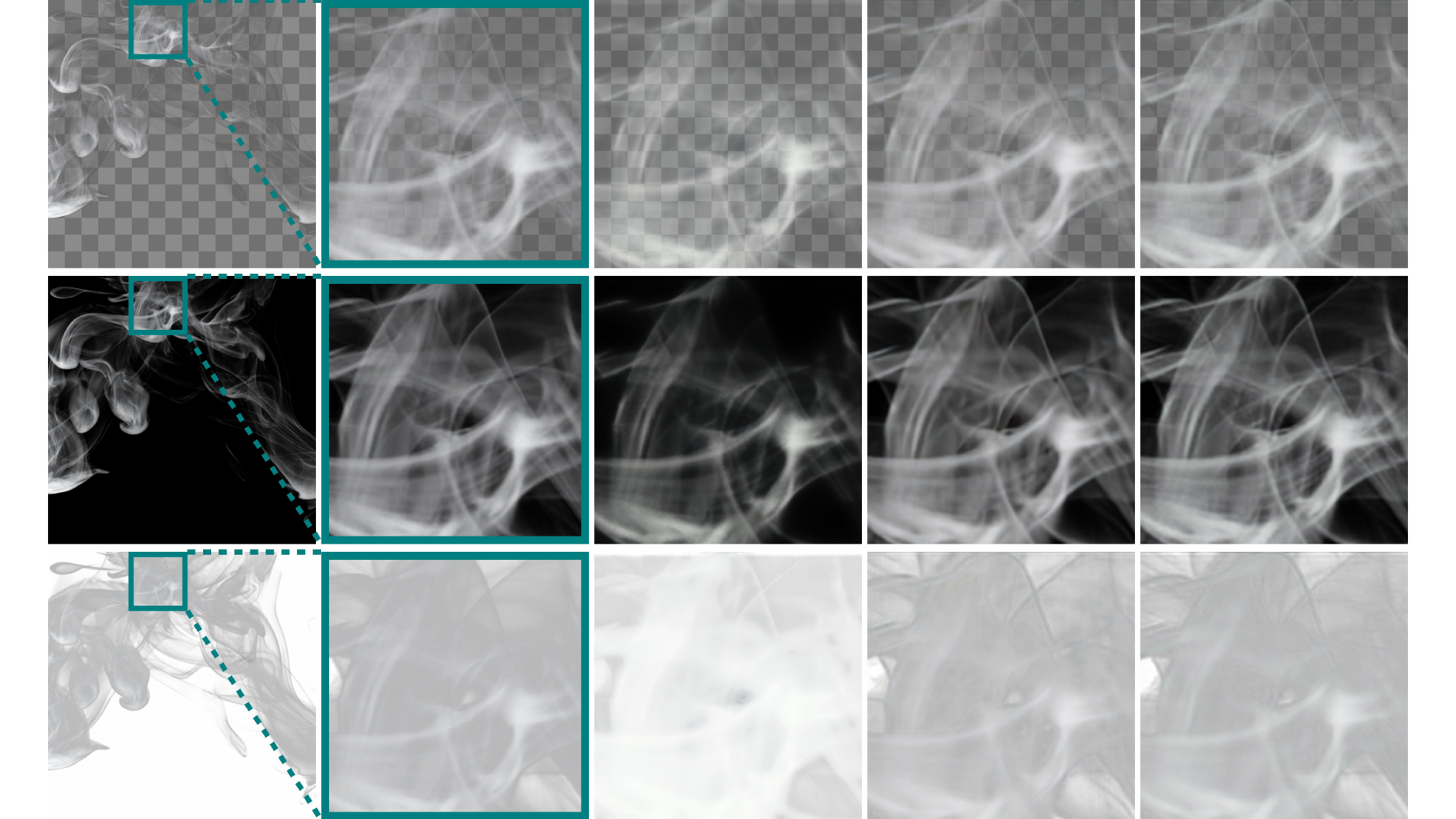}
    \vspace{1mm}
    \adjustbox{max width=\linewidth}{
      \begin{tabularx}{\linewidth}{XXXXX}
        \centering \makebox[0.18\linewidth]{GT} &
        \centering \makebox[0.18\linewidth]{GT*} &
        \centering \makebox[0.18\linewidth]{LD} &
        \centering \makebox[0.18\linewidth]{Ours*} &
        \centering \makebox[0.18\linewidth]{Ours}
      \end{tabularx}
    }
  \end{subfigure}
  \vspace{-0.7em}
  \caption{Qualitative Results of image reconstruction on transparent images. Each row shows reconstruction results (with 4 channels) composited over three backgrounds: checkerboard, black, and white (from top to bottom) for better visualization. LD represents layerdiffuse~\cite{layerdiffuse}, Ours* denotes \method{} based on SDXL, and Ours is \method{} based on FLUX. The above images have rich alpha channel information from the AIM-500 dataset.}
  \label{fig:qualitative_vae}
  \vspace{-1.5em}
\end{figure}

\begin{figure}[t]
    \centering
    \includegraphics[width=\linewidth]{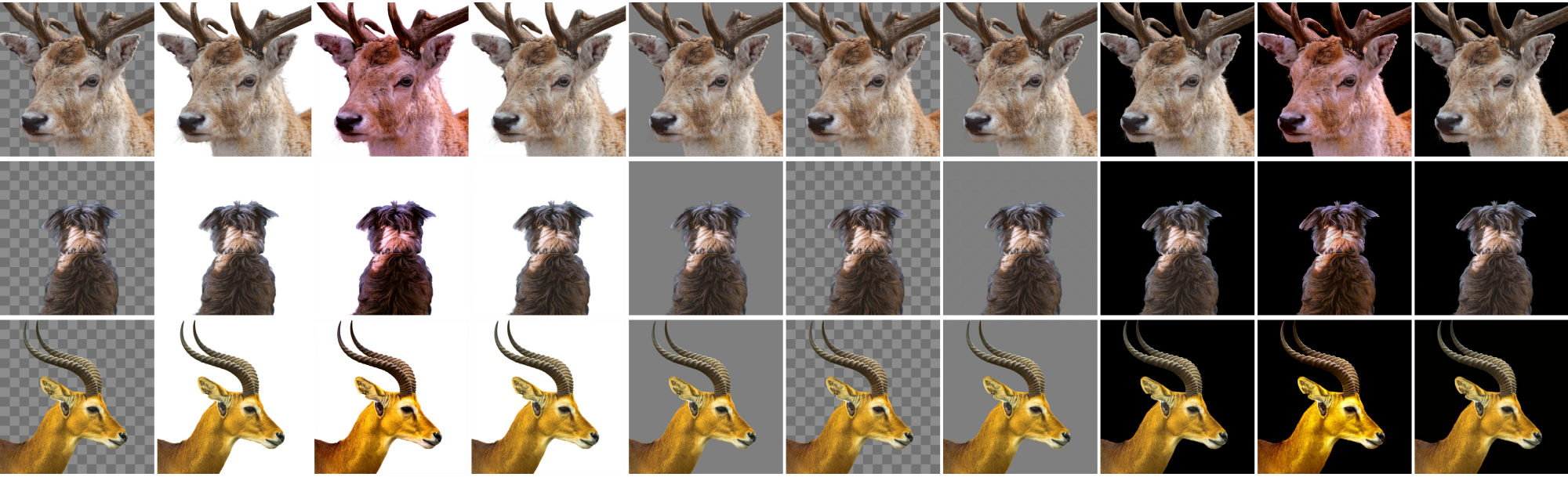}
    % \vspace{1mm}
    \begin{tabularx}{\linewidth}{XXXXXXXXXX}
        \centering\makebox[0.1\linewidth]{GT} &
        \centering\makebox[0.1\linewidth]{GT-white} &
        \centering\makebox[0.1\linewidth]{LD} &
        \centering\makebox[0.1\linewidth]{Ours} &
        \centering\makebox[0.1\linewidth]{GT-gray} &
        \centering\makebox[0.1\linewidth]{LD} &
        \centering\makebox[0.1\linewidth]{Ours} &
        \centering\makebox[0.1\linewidth]{GT-black} &
        \centering\makebox[0.1\linewidth]{LD} &
        \centering\makebox[0.1\linewidth]{Ours}
    \end{tabularx}
    % \vspace{-1.35em}    
    \caption{Qualitative results of image reconstruction on transparent images blended over solid color backgrounds. GT-white, GT-gray, and GT-black denote the ground-truth transparent image blended with white, gray, and black backgrounds, respectively.}
    \label{fig:blend}
\end{figure}

\textbf{VAE.} 
We present qualitative reconstruction results for transparent images in Figure~\ref{fig:qualitative_vae}. Each row displays composited outputs on three different backgrounds: checkerboard, black, and white, providing a clear view of both the RGB and the alpha channel fidelity. All samples are drawn from the AIM-500 test dataset, which contains images with rich transparency details. Note that this dataset was not used during training. Compared to layerdiffuse~\cite{layerdiffuse}, our model (Ours and Ours*) better preserves the original structure and transparency of the input images. In particular, LayerDiffuse~\cite{layerdiffuse} fails to reconstruct the alpha channel faithfully: it tends to shrink the transparency regions or reduce opacity, leading to results that diverge from the original image. This indicates a degradation in the representational capacity of its VAE component, as it cannot maintain consistent alpha values during decoding. Furthermore, when the transparent input image is composited with a solid-color background, the VAE in LayerDiffuse not only fails to preserve alpha consistency, but also exhibits noticeable degradation in RGB fidelity — for instance, as shown in Figure~\ref{fig:blend}, reconstructed images often display a reddish color cast, diverging from the original appearance.

\textbf{Diffusion.} We generate images using our fine-tuned diffusion model, as shown in Figure~\ref{fig:qualitative_diffusion}. The prompts used for image generation are derived by applying a captioning model~\cite{Qwen2.5-VL} to the original images in the AIM-500 dataset and the \benchmark{} test split. Notably, the training set contains only around 8K images, yet the model is able to generate semantically aligned images from the out-of-distribution AIM-500 dataset. The plausible generation highlights the strong generative prior learned by the diffusion model.

\subsection{Ablative Study}

\renewcommand{\valueoverpercentage}[3]{\makecell{#1\\[-.3em] \scriptsize{\textcolor{#2}{(#3)}}}}

\begin{table}[htbp]
  \centering
  \caption{Ablation study results.}
  \vspace{.1em}
  \label{tab:ablation_study_new}
  \resizebox{\textwidth}{!}{%
  \begin{tabular}{l ccccc ccccc}
    \toprule
    \multirow{3.5}{*}{\textbf{Method}} & \multicolumn{5}{c}{\textbf{AIM}} & \multicolumn{5}{c}{\textbf{Alpha}} \\
    \cmidrule(lr){2-6} \cmidrule(lr){7-11}
    & PSNR $\uparrow$ & SSIM $\uparrow$ & \makecell{LAION\\AES} $\uparrow$ & rFID $\downarrow$ & LPIPS $\downarrow$ & PSNR $\uparrow$ & SSIM $\uparrow$ & \makecell{LAION\\AES} $\uparrow$ & rFID $\downarrow$ & LPIPS $\downarrow$ \\
    \midrule
    AlphaVAE + FLUX & \uline{36.9439} & \uline{0.9737} & 4.9511 & \textbf{11.7884} & \textbf{0.0283} & \uline{38.1987} & \uline{0.9792} & 5.0616 & \uline{1.6600} & \textbf{0.0145} \\ [.5em]
    \quad w/o Norm KL & \valueoverpercentage{35.1629}{mypurple}{-1.7810} & \valueoverpercentage{0.9691}{mypurple}{-0.0046} & \valueoverpercentage{4.9538}{deepgreen}{+0.0027} & \valueoverpercentage{16.3582}{mypurple}{+4.5698} & \valueoverpercentage{0.0358}{mypurple}{+0.0075} & \valueoverpercentage{37.3880}{mypurple}{-0.8107} & \valueoverpercentage{0.9776}{mypurple}{-0.0016} & \valueoverpercentage{\textbf{5.0639}}{deepgreen}{+0.0023} & \valueoverpercentage{2.4898}{mypurple}{+0.8298} & \valueoverpercentage{0.0174}{mypurple}{+0.0029} \\ [.5em]
    \quad w/o Ref KL & \valueoverpercentage{\textbf{36.9473}}{deepgreen}{+0.0034} & \valueoverpercentage{\textbf{0.9738}}{deepgreen}{+0.0001} & \valueoverpercentage{\textbf{4.9555}}{deepgreen}{+0.0044} & \valueoverpercentage{\uline{12.0928}}{mypurple}{+0.3044} & \valueoverpercentage{\uline{0.0287}}{mypurple}{+0.0004} & \valueoverpercentage{38.1428}{mypurple}{-0.0559} & \valueoverpercentage{0.9790}{mypurple}{-0.0002} & \valueoverpercentage{\uline{5.0635}}{deepgreen}{+0.0019} & \valueoverpercentage{1.7945}{mypurple}{+0.1345} & \valueoverpercentage{0.0155}{mypurple}{+0.0010} \\ [.5em]
    \quad w/o GAN & \valueoverpercentage{36.8578}{mypurple}{-0.0861} & \valueoverpercentage{0.9734}{mypurple}{-0.0003} & \valueoverpercentage{4.9484}{mypurple}{-0.0027} & \valueoverpercentage{12.8674}{mypurple}{+1.0790} & \valueoverpercentage{0.0299}{mypurple}{+0.0016} & \valueoverpercentage{\textbf{38.4839}}{deepgreen}{+0.2852} & \valueoverpercentage{\textbf{0.9802}}{deepgreen}{+0.0010} & \valueoverpercentage{5.0556}{mypurple}{-0.0060} & \valueoverpercentage{1.7091}{mypurple}{+0.0491} & \valueoverpercentage{0.0154}{mypurple}{+0.0009} \\ [.5em]
    \quad w/o LPIPS & \valueoverpercentage{32.9796}{mypurple}{-3.9643} & \valueoverpercentage{0.9636}{mypurple}{-0.0101} & \valueoverpercentage{\uline{4.9547}}{deepgreen}{+0.0036} & \valueoverpercentage{22.3132}{mypurple}{+10.5248} & \valueoverpercentage{0.0457}{mypurple}{+0.0174} & \valueoverpercentage{36.5583}{mypurple}{-1.6404} & \valueoverpercentage{0.9763}{mypurple}{-0.0029} & \valueoverpercentage{5.0538}{mypurple}{-0.0078} & \valueoverpercentage{3.1332}{mypurple}{+1.4732} & \valueoverpercentage{0.0196}{mypurple}{+0.0051} \\ [.5em]
    \quad w/o ABMSE & \valueoverpercentage{36.8036}{mypurple}{-0.1403} & \valueoverpercentage{0.9709}{mypurple}{-0.0028} & \valueoverpercentage{4.9460}{mypurple}{-0.0051} & \valueoverpercentage{13.4679}{mypurple}{+1.6795} & \valueoverpercentage{0.0312}{mypurple}{+0.0029} & \valueoverpercentage{38.0847}{mypurple}{-0.1140} & \valueoverpercentage{0.9784}{mypurple}{-0.0008} & \valueoverpercentage{5.0589}{mypurple}{-0.0027} & \valueoverpercentage{\textbf{1.6599}}{deepgreen}{-0.0001} & \valueoverpercentage{\uline{0.0150}}{mypurple}{+0.0005} \\
    \midrule
    AlphaVAE + SDXL & \textbf{35.7446} & \textbf{0.9576} & \uline{4.9456} & \textbf{10.9178} & \textbf{0.0495} & \uline{35.5597} & \uline{0.9605} & \uline{5.0500} & 3.7305 & 0.0402 \\ [.5em]
    \quad w/o Norm KL & \valueoverpercentage{33.8361}{mypurple}{-1.9085} & \valueoverpercentage{0.9504}{mypurple}{-0.0072} & \valueoverpercentage{4.9277}{mypurple}{-0.0179} & \valueoverpercentage{18.9242}{mypurple}{+8.0064} & \valueoverpercentage{0.0607}{mypurple}{+0.0112} & \valueoverpercentage{35.4985}{mypurple}{-0.0612} & \valueoverpercentage{0.9604}{mypurple}{-0.0001} & \valueoverpercentage{5.0475}{mypurple}{-0.0025} & \valueoverpercentage{3.8617}{mypurple}{+0.1312} & \valueoverpercentage{0.0401}{deepgreen}{-0.0001} \\ [.5em]
    \quad w/o Ref KL & \valueoverpercentage{33.9751}{mypurple}{-1.7695} & \valueoverpercentage{0.9509}{mypurple}{-0.0067} & \valueoverpercentage{4.9383}{mypurple}{-0.0073} & \valueoverpercentage{18.2684}{mypurple}{+7.3506} & \valueoverpercentage{0.0587}{mypurple}{+0.0092} & \valueoverpercentage{35.4917}{mypurple}{-0.0680} & \valueoverpercentage{0.9602}{mypurple}{-0.0003} & \valueoverpercentage{5.0484}{mypurple}{-0.0016} & \valueoverpercentage{3.7902}{mypurple}{+0.0597} & \valueoverpercentage{\textbf{0.0393}}{deepgreen}{-0.0009} \\ [.5em]
    \quad w/o GAN & \valueoverpercentage{34.0428}{mypurple}{-1.7018} & \valueoverpercentage{0.9507}{mypurple}{-0.0069} & \valueoverpercentage{4.9338}{mypurple}{-0.0118} & \valueoverpercentage{18.5222}{mypurple}{+7.6044} & \valueoverpercentage{0.0598}{mypurple}{+0.0103} & \valueoverpercentage{\textbf{35.6887}}{deepgreen}{+0.1290} & \valueoverpercentage{\textbf{0.9610}}{deepgreen}{+0.0005} & \valueoverpercentage{5.0467}{mypurple}{-0.0033} & \valueoverpercentage{\textbf{3.6883}}{deepgreen}{-0.0422} & \valueoverpercentage{0.0405}{mypurple}{+0.0003} \\ [.5em]
    \quad w/o LPIPS & \valueoverpercentage{33.9153}{mypurple}{-1.8293} & \valueoverpercentage{0.9511}{mypurple}{-0.0065} & \valueoverpercentage{4.9348}{mypurple}{-0.0108} & \valueoverpercentage{18.6036}{mypurple}{+7.6858} & \valueoverpercentage{0.0592}{mypurple}{+0.0097} & \valueoverpercentage{35.4946}{mypurple}{-0.0651} & \valueoverpercentage{0.9604}{mypurple}{-0.0001} & \valueoverpercentage{5.0494}{mypurple}{-0.0006} & \valueoverpercentage{3.7675}{mypurple}{+0.0370} & \valueoverpercentage{\uline{0.0396}}{deepgreen}{-0.0006} \\ [.5em]
    \quad w/o ABMSE & \valueoverpercentage{\uline{34.5586}}{mypurple}{-1.1860} & \valueoverpercentage{\uline{0.9512}}{mypurple}{-0.0064} & \valueoverpercentage{\textbf{4.9471}}{deepgreen}{+0.0015} & \valueoverpercentage{\uline{16.0156}}{mypurple}{+5.0978} & \valueoverpercentage{\uline{0.0559}}{mypurple}{+0.0064} & \valueoverpercentage{35.3997}{mypurple}{-0.1600} & \valueoverpercentage{0.9587}{mypurple}{-0.0018} & \valueoverpercentage{\textbf{5.0603}}{deepgreen}{+0.0103} & \valueoverpercentage{\uline{3.7250}}{deepgreen}{-0.0055} & \valueoverpercentage{\textbf{0.0393}}{deepgreen}{-0.0009} \\ [.5em]
    \bottomrule
  \end{tabular}}
\end{table}

Following the evaluation protocols described in Section~\ref{subsec:evaluation_protocol}, we systematically assess the performance of various VAE configurations using PSNR, SSIM, LAION, rFID, and LPIPS metrics. Overall, our proposed method consistently achieves the best results across all evaluation criteria.

Notably, within the FLUX-VAE framework, the configuration without the reference KL term ("w/o ref KL") performs particularly well on the AIM-500 dataset. The reference KL term is specifically designed to incorporate information from the alpha channel into the latent representation without significantly disrupting the original latent distribution. As shown in Table~\ref{tab:ablation_study_new}, this design helps to better preserve generation quality during subsequent training in the diffusion model.

\section{Conclusion}
\label{sec:conclusion}

In this paper, we address the long-standing challenge of transparent image evaluation. 
We first introduce \benchmark{}, the first comprehensive RGBA benchmark that seamlessly extends standard RGB metrics, such as PSNR, SSIM, and FID, via alpha blending. 
And we present \method{}, a unified end-to-end RGBA VAE architecture that augments a pretrained three-channel VAE with a dedicated alpha channel and specialized losses to preserve latent fidelity and achieve high-quality reconstruction. 
Through extensive experiments on reconstruction and image generation tasks, we demonstrate that, with only 8K training images, our method outperforms state-of-the-art approaches trained on orders of magnitude more data in both VAE reconstruction and transparent image generation when fine-tuned within a latent diffusion framework.
Our work not only fills a critical gap in evaluation protocols and model design for RGBA image reconstruction and synthesis but also lays the groundwork for future extensions to dynamic, multi-layer video generation and interactive editing applications.

\clearpage
\newpage
\bibliographystyle{unsrt}
{\small\bibliography{main}}

% \input{checklist}

%%%%%%%%%%%%%%%%%%%%%%%%%%%%%%%%%%%%%%%%%%%%%%%%%%%%%%%%%%%%

\appendix
\newpage
\setcounter{page}{1}

\begin{center}
  {\LARGE\bfseries Appendix}  % \Huge 约 24 pt，\bfseries 加粗
\end{center}
\section{Formula Derivation}
\label{app:formula}

In this section, we derive Equation~\eqref{eq:reconstruction_loss}.  
Assuming the three channels of the background color $b\!\in\!\mathbb{R}^3$ are independent and identically distributed, we can work on a single pixel without lack of generality.  
Define
\begin{equation}
\label{eq:defs}
\begin{cases}
\displaystyle
\mathcal{L}_{\text{rec}}(\mathcal{E},\mathcal{D})
      =\mathbb{E}_{b}\!\Bigl[\bigl\lVert\mathcal{A}(\hat{x},b)-\mathcal{A}(x,b)\bigr\rVert_2^2\Bigr],\\[6pt]
\displaystyle
\mathcal{A}(\hat{x},b)-\mathcal{A}(x,b)=P-\Delta_\alpha b,\\[6pt]
\displaystyle
P=\hat{x}_{\text{rgb}}\odot\hat{x}_{\alpha}-x_{\text{rgb}}\odot x_{\alpha}\in\mathbb{R}^3,\\[6pt]
\displaystyle
\Delta_\alpha=\hat{x}_{\alpha}-x_{\alpha}\in\mathbb{R}.
\end{cases}
\end{equation}

% Main derivation
Substituting the second line of~\eqref{eq:defs} into the first gives
\begin{align}
\mathcal{L}_{\text{rec}}(\mathcal{E},\mathcal{D})
&=\mathbb{E}_{b}\!\Bigl[\bigl\lVert P-\Delta_\alpha b\bigr\rVert_2^2\Bigr]                                                     \notag\\
&=\mathbb{E}_{b}\!\Bigl[(P-\Delta_\alpha b)(P-\Delta_\alpha b)^{\!\top}\Bigr]                                                  \notag\\
&=\mathbb{E}_{b}\!\Bigl[PP^{\!\top}-2\Delta_\alpha\,bP^{\!\top}-\Delta_\alpha^{2}bb^{\!\top}\Bigr]                             \notag\\
&=\lVert P\rVert_2^{\,2}-2\Delta_\alpha\,\langle\mathbb{E}[b],P\rangle-\Delta_\alpha^{2}\,\mathbb{E}\!\bigl[bb^{\!\top}\bigr].
\end{align}
Because $\mathbb{E}\!\bigl[bb^{\!\top}\bigr]=\mathbb{E}[b_1^2+b_2^2+b_3^2]=\lVert\mathbb{E}[b^{2}]\rVert_1$,
Equation~\eqref{eq:reconstruction_loss} follows immediately.

% Practical form
For implementation, we use the equivalent channel-wise form
\begin{align}
\mathcal{L}_{\text{rec}}(\mathcal{E},\mathcal{D})
&=\lVert P\rVert_2^{\,2}-2\Delta_\alpha\,\langle\mathbb{E}[b],P\rangle-\Delta_\alpha^{2}\,\lVert\mathbb{E}[b^{2}]\rVert_1                                   \notag\\
&=\lVert P^{2}\rVert_1-\mathbb{E}_{\text{ch}}\!\bigl[2\Delta_\alpha\,\mathbb{E}[b]\odot P\bigr]-\Delta_\alpha^{2}\,\lVert\mathbb{E}[b^{2}]\rVert_1             \notag\\
&=\mathbb{E}_{\text{ch}}\!\Bigl[P^{2}-2\Delta_\alpha\,\mathbb{E}[b]\odot P-\Delta_\alpha^{2}\,\mathbb{E}[b^{2}]\Bigr],
\end{align}
where $\mathbb{E}_{\text{ch}}$ denotes the expectation over the RGB channels.
%--------------------------------------------------------------------------- 

% \section{Technical Appendices and Supplementary Material}
% Technical appendices with additional results, figures, graphs, and proofs may be submitted with the paper submission before the full submission deadline (see above), or as a separate PDF in the ZIP file below before the supplementary material deadline. There is no page limit for the technical appendices.

\begin{figure}[th]
    \centering
    \includegraphics[width=\linewidth]{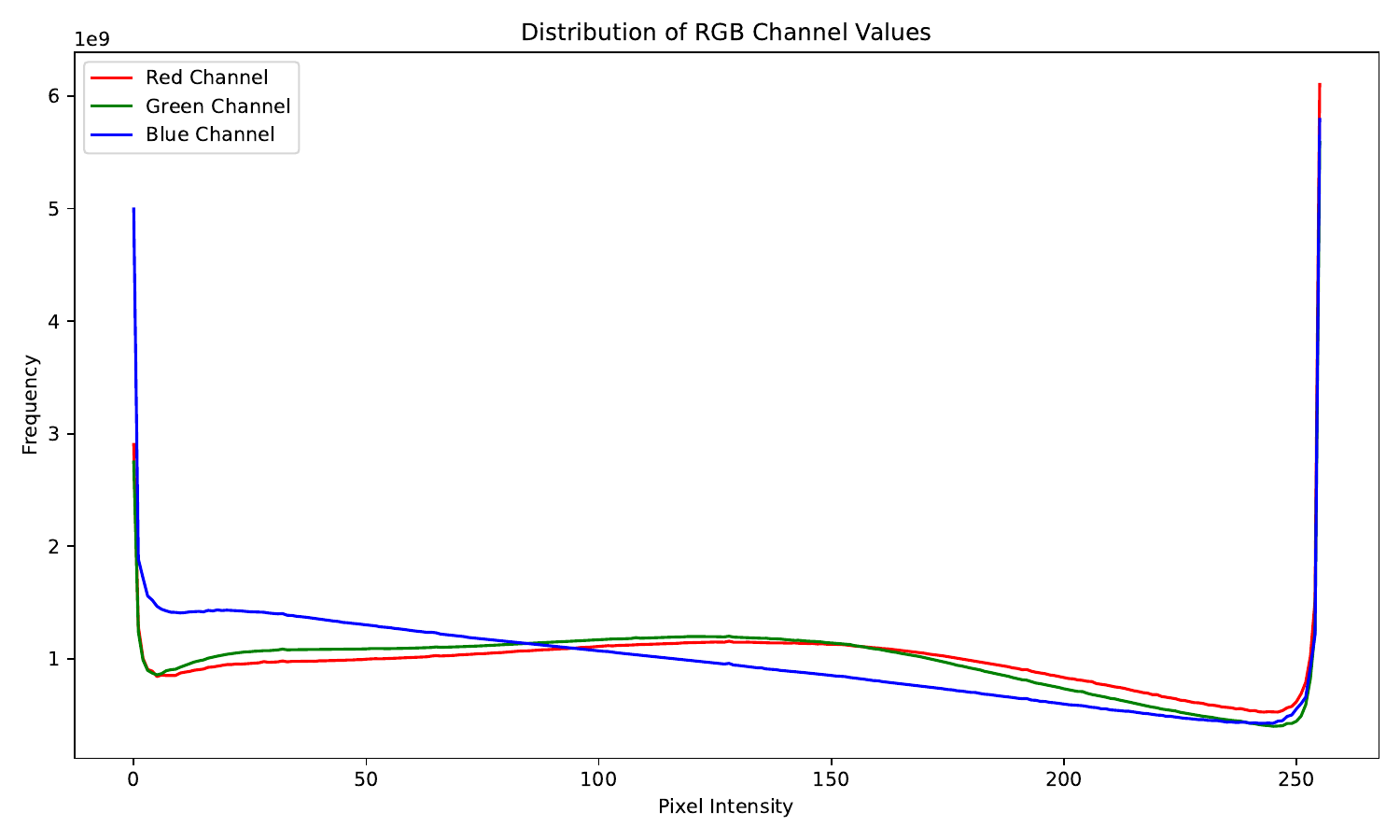}
    \caption{Color distribution of pixels in ImageNet train split. The pixel values are scaled to $[0,255]$.}
    \label{fig:color-distribution}
\end{figure}

\section{Details of Hyperparameters.}
\label{app:params}

In our reconstruction loss, the background pixel distribution $b$ contributes two pre-computable terms: its first raw moment, $\mathbb{E}[b]$, and its second raw moment, $\mathbb{E}[b^{2}]$.
Figure \ref{fig:color-distribution} visualizes the empirical distribution of $b$ on the ImageNet training split.
Throughout training we rescale pixel intensities to the range $[-1,1]$, yielding
\begin{equation}
    \mathbb{E}[b]=(-0.0357,\,-0.0811,\,-0.1797)
    \quad\text{and}\quad
    \mathbb{E}[b^{2}]=(0.3163,\,0.3060,\,0.3634).
\end{equation}

In addition to the background pixel statistics, we detail the weighting of various loss terms introduced in Section~\ref{sec:method}. Specifically, we use a reconstruction loss with a weight of 1.0, a perceptual loss (computed using LPIPS) with a weight of 0.5, and a composite regularization loss comprising two KL divergence terms: the primary latent KL loss is weighted by $10^{-6}$, while the reference KL loss is assigned a much smaller weight of $10^{-16}$. For adversarial training, we enable the GAN loss after 4000 steps and set the generator loss weight to 1.0.

\section{Limitations}
\label{app:limitation}

% We explore the LoRA paradigm for parameter-efficient fine-tuning.
% Actually, there are some other ways to fine-tune a pre-trained model, such as full-parameter fine-tuning, Control-Net, etc.
% Due to computation cost, we have not tested these ways.
% Future work may experiment on this.

Our study focuses exclusively on parameter-efficient fine-tuning with LoRA.
Other fine-tuning methods, such as full-parameter fine-tuning and task-specific adaptation modules like ControlNet,
are not evaluated because they require significantly more computation.
Investigating these approaches is a valuable avenue for future work.

\section{More Qualitative Analysis}
\label{app:samples}

\begin{figure}[t]
    \centering
    \includegraphics[width=\linewidth]{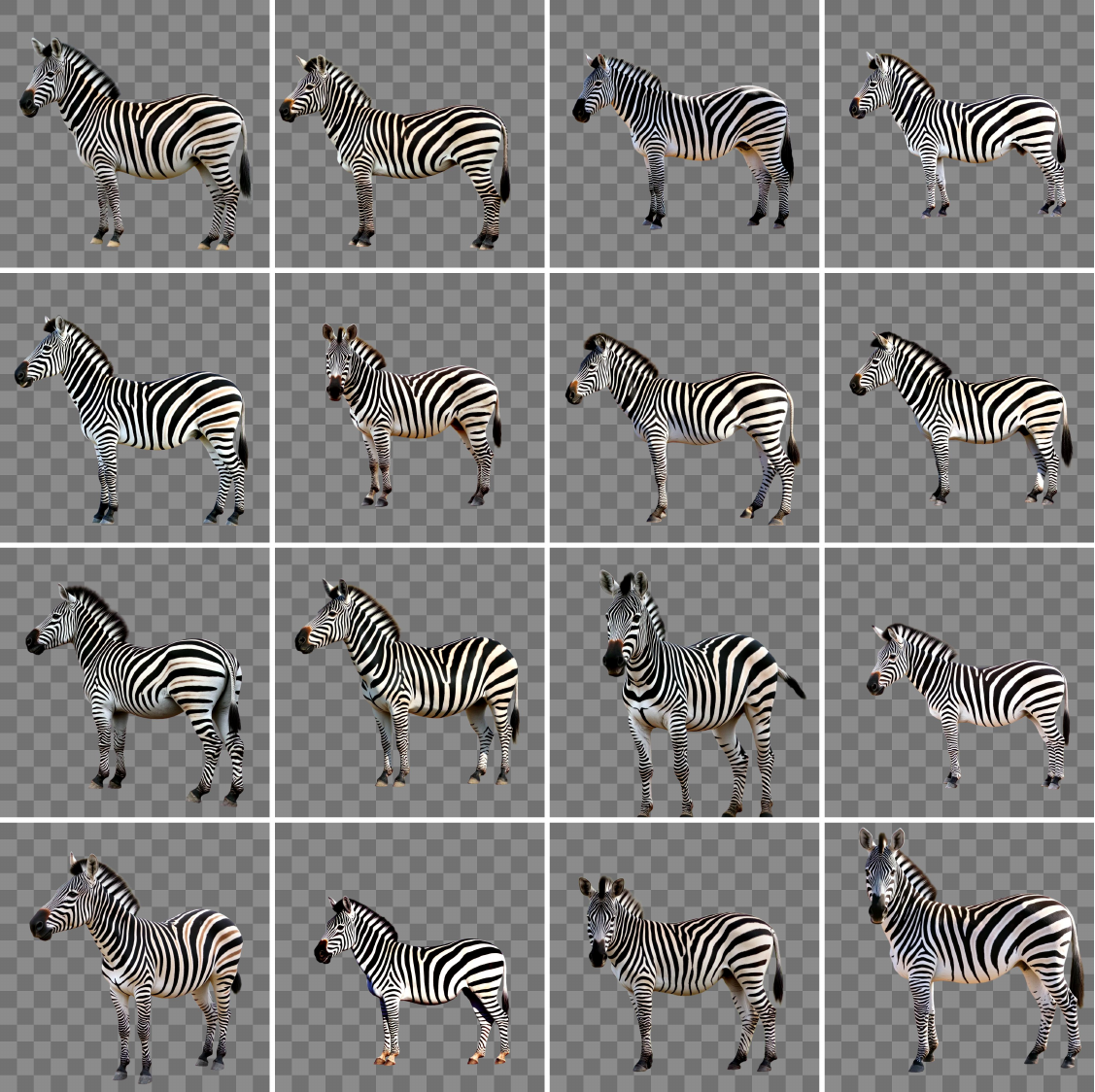}  
    \caption{A zebra stands with its head turned to the left, showcasing its distinctive black and white stripes.}
    \label{fig:appendix_1}
\end{figure}

\begin{figure}[t]
    \centering
    \includegraphics[width=\linewidth]{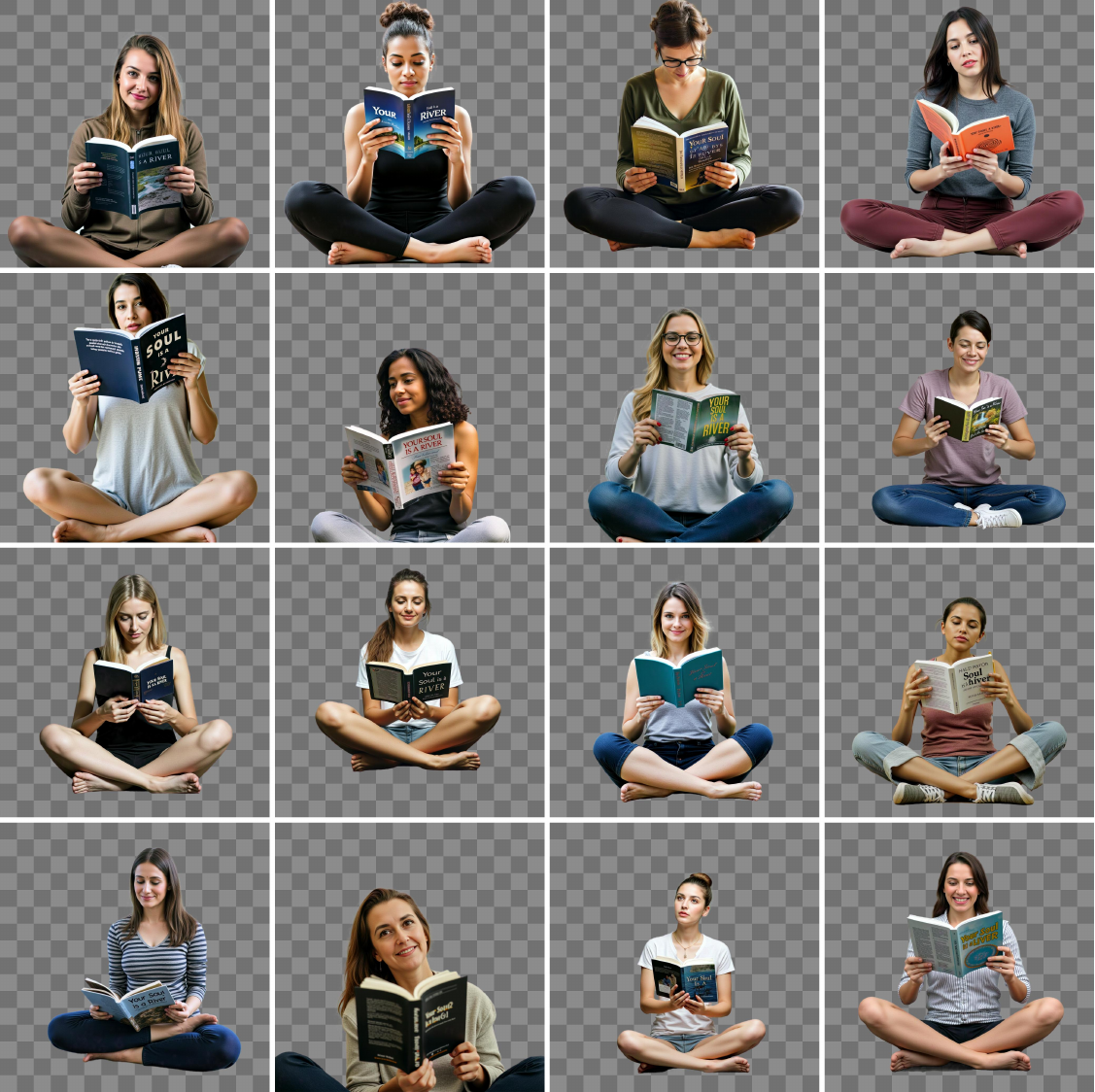}  
    \caption{A woman sits cross-legged, reading a book titled "Your Soul is a River."}
    \label{fig:appendix_2}
\end{figure}

\begin{figure}[t]
    \centering
    \includegraphics[width=\linewidth]{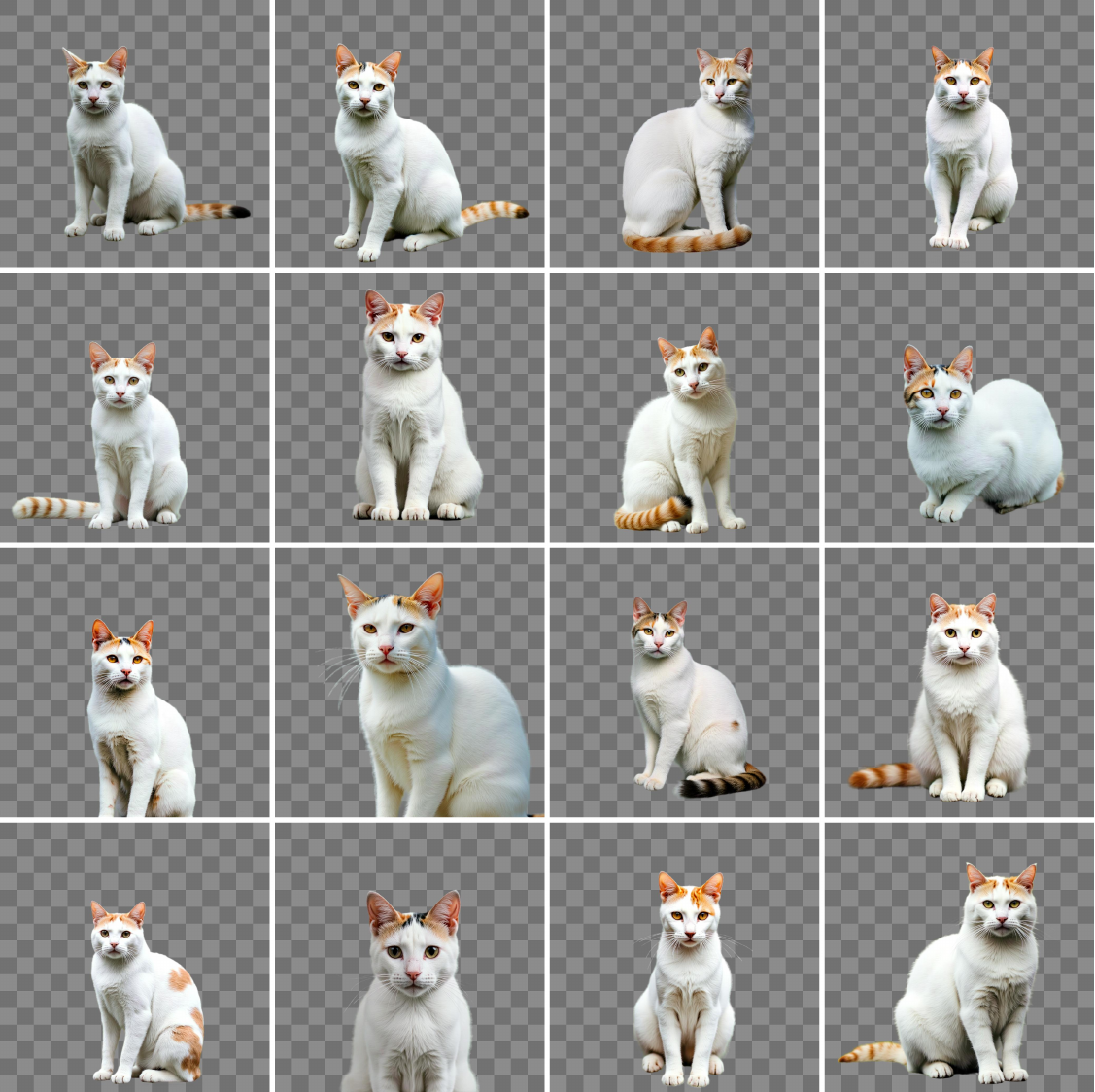}  
    \caption{A white cat with orange and black markings sits calmly, gazing forward.}
    \label{fig:appendix_3}
\end{figure}

\begin{figure}[t]
    \centering
    \includegraphics[width=\linewidth]{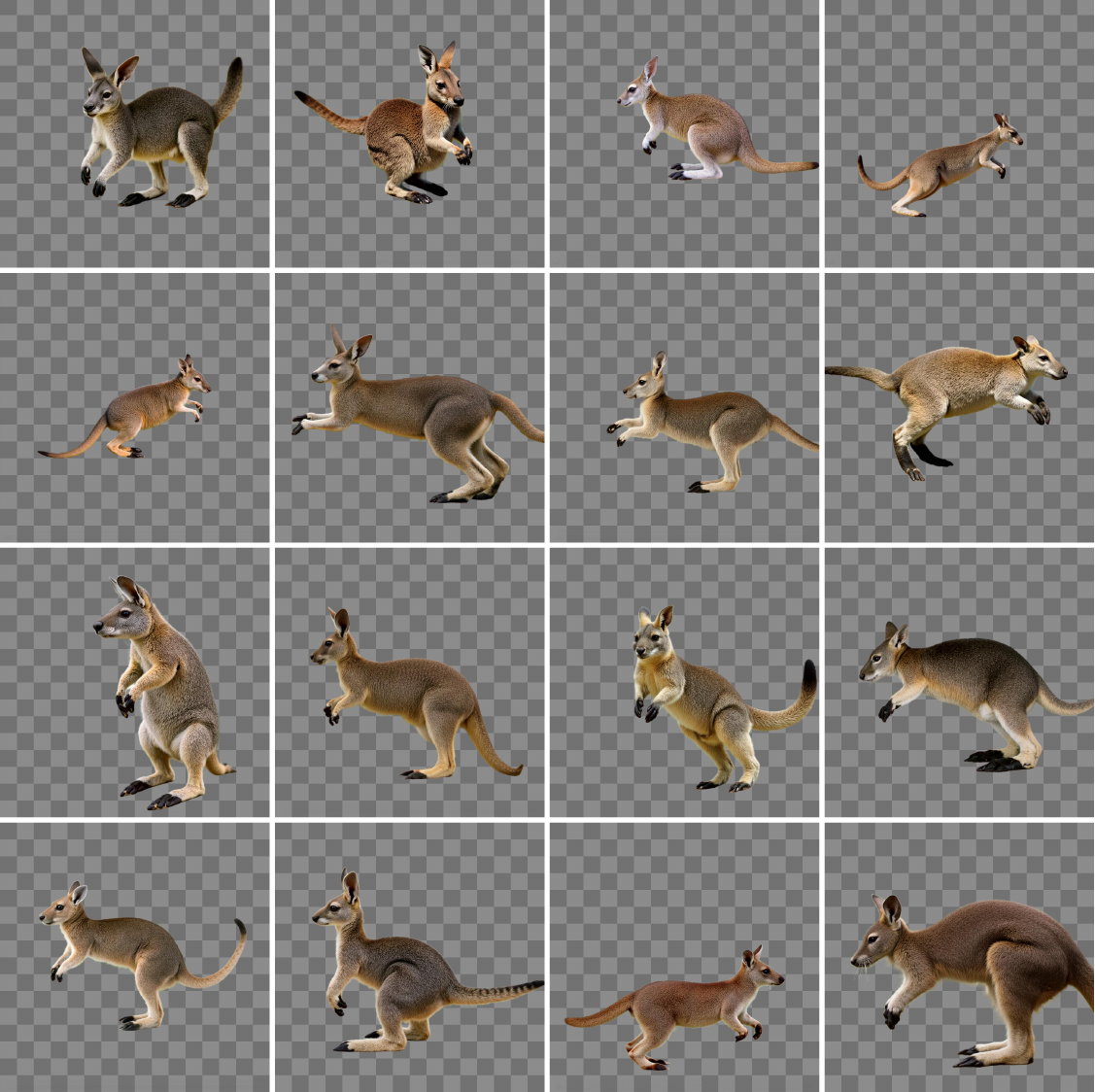}  
    \caption{A wallaby is captured mid-hop, showcasing its agile movement and distinctive features.}
    \label{fig:appendix_4}
\end{figure}

\begin{figure}[t]
    \centering
    \includegraphics[width=\linewidth]{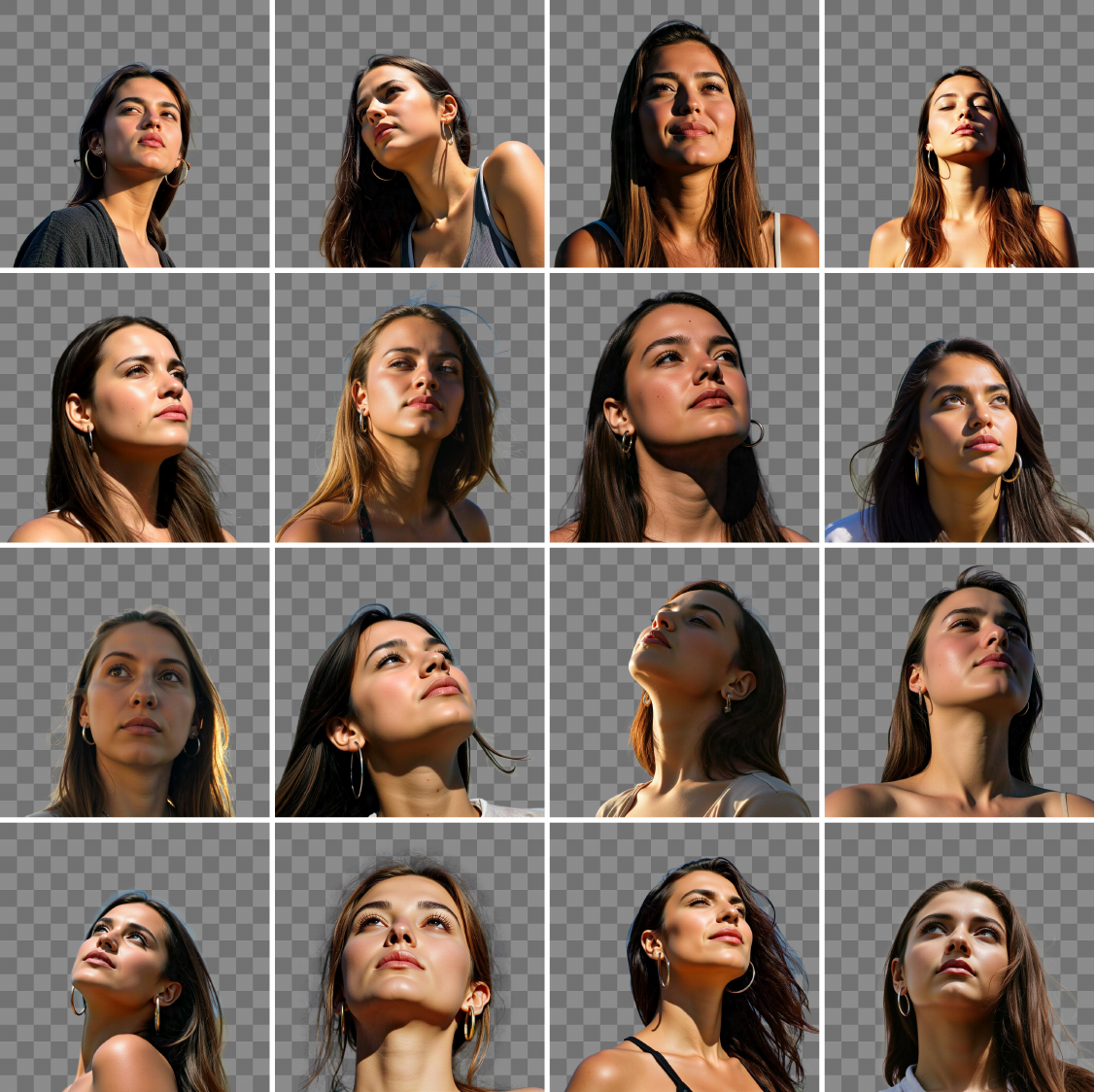}  
    \caption{A person with long hair and hoop earrings looks upward, bathed in sunlight.}
    \label{fig:appendix_5}
\end{figure}

\begin{figure}[t]
    \centering
    \includegraphics[width=\linewidth]{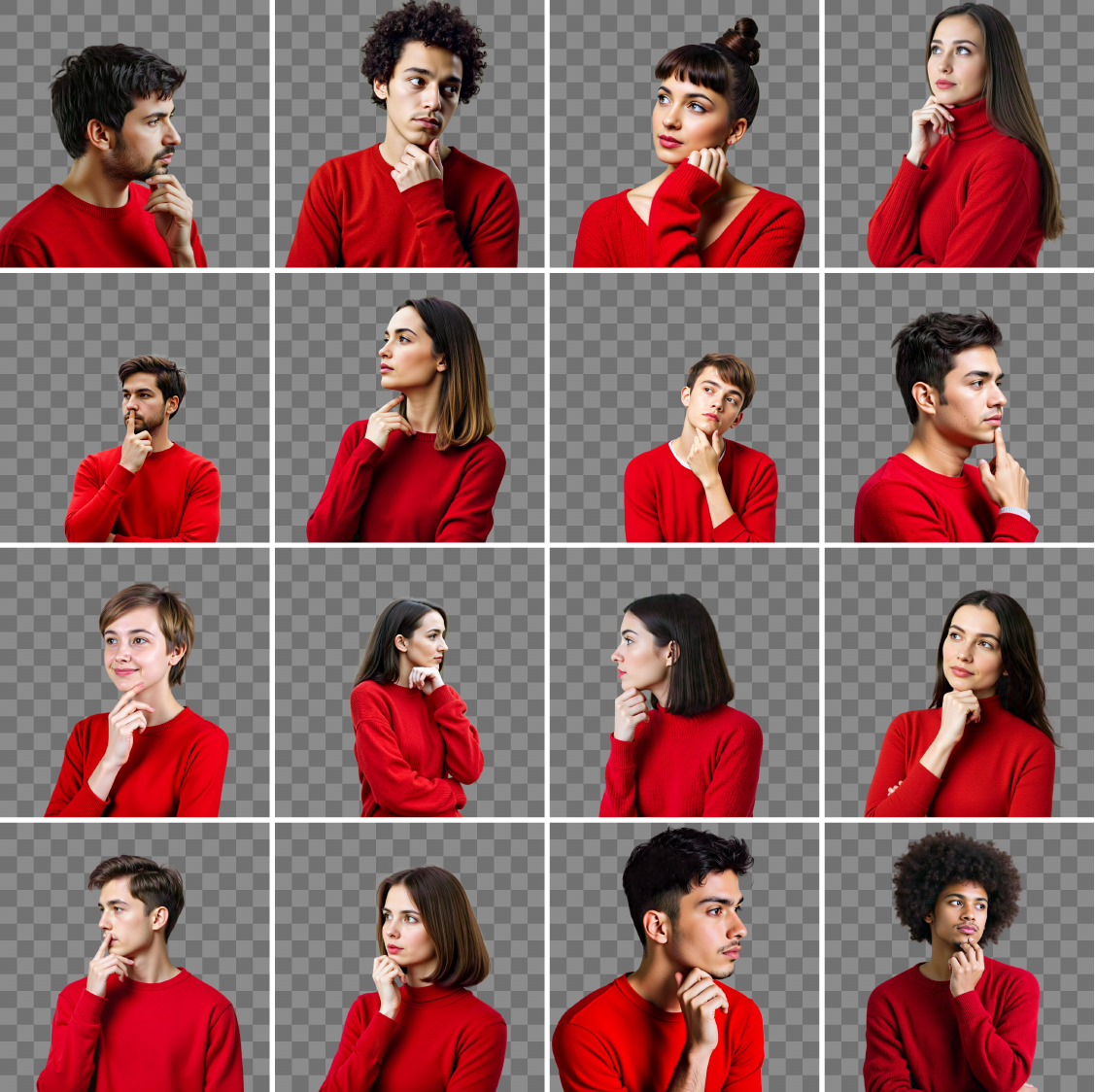}  
    \caption{A person wearing a red sweater is looking to the side with their hand on their chin.}
    \label{fig:appendix_6}
\end{figure}

\begin{figure}[t]
    \centering
    \includegraphics[width=\linewidth]{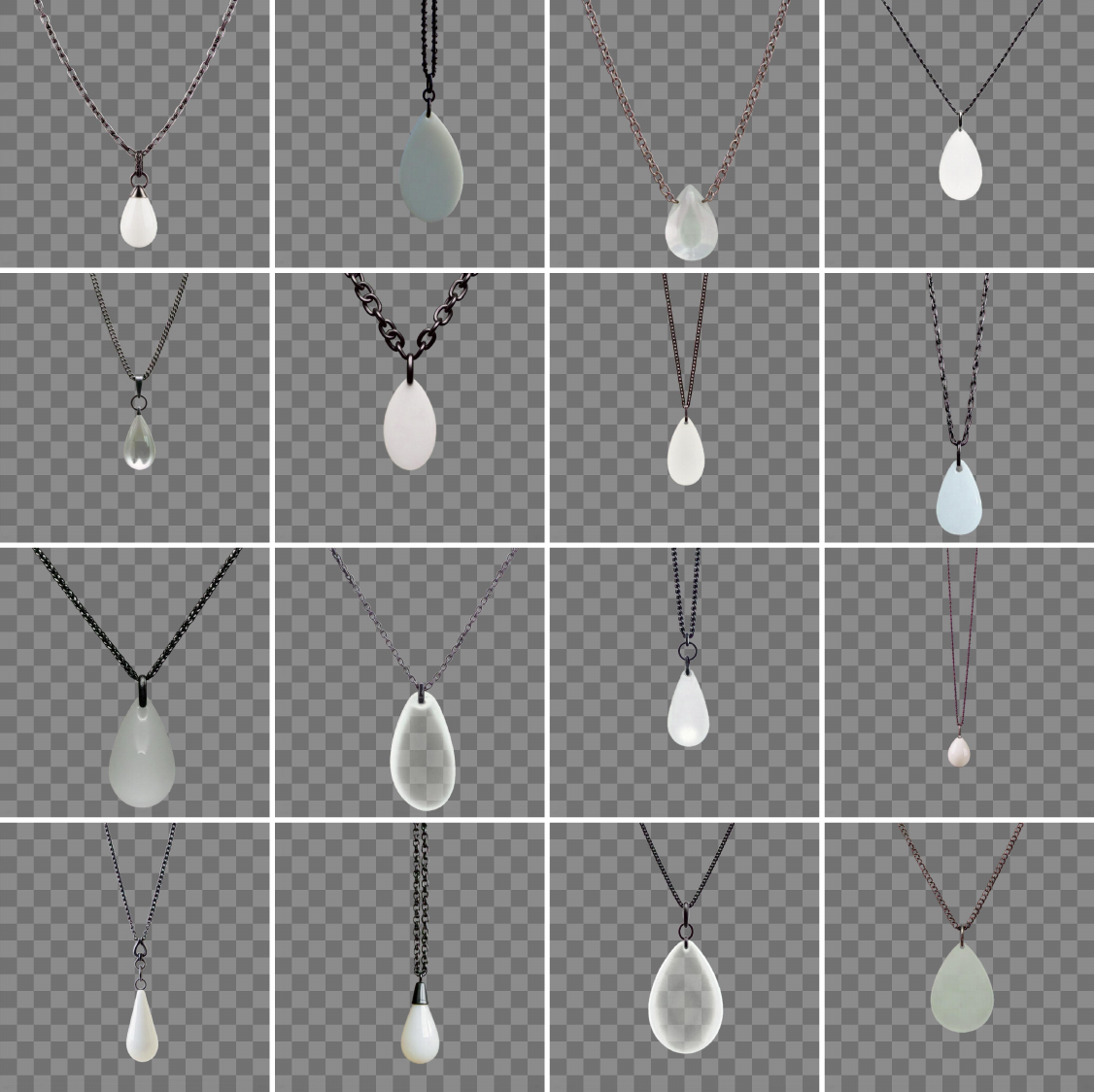}  
    \caption{A necklace with a black chain and a white, teardrop-shaped pendant hangs against a white backdrop.}
    \label{fig:appendix_7}
\end{figure}

\begin{figure}[t]
    \centering
    \includegraphics[width=\linewidth]{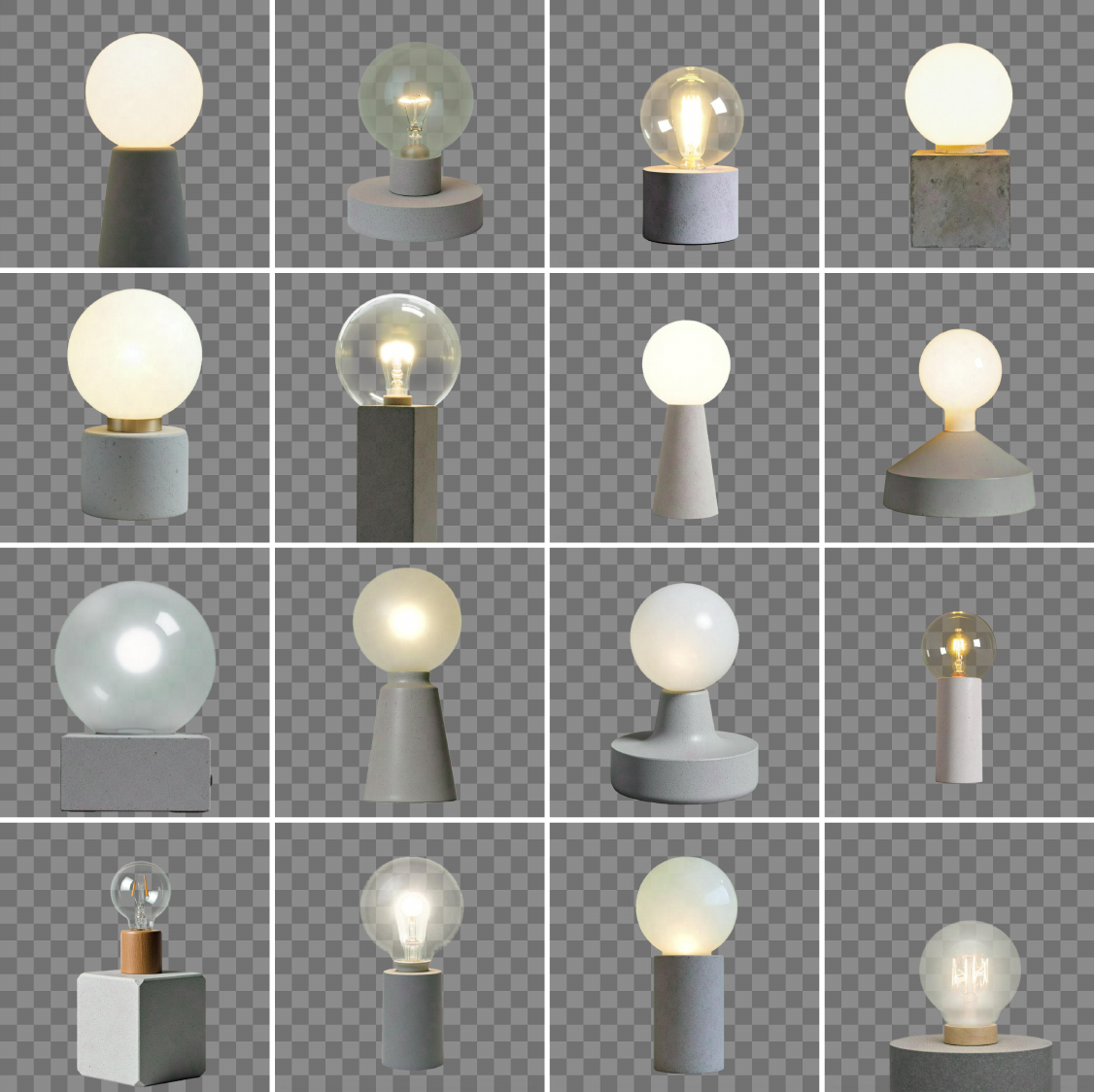}  
    \caption{A modern table lamp with a concrete base and a round, frosted glass bulb.}
    \label{fig:appendix_8}
\end{figure}

\begin{figure}[t]
    \centering
    \includegraphics[width=\linewidth]{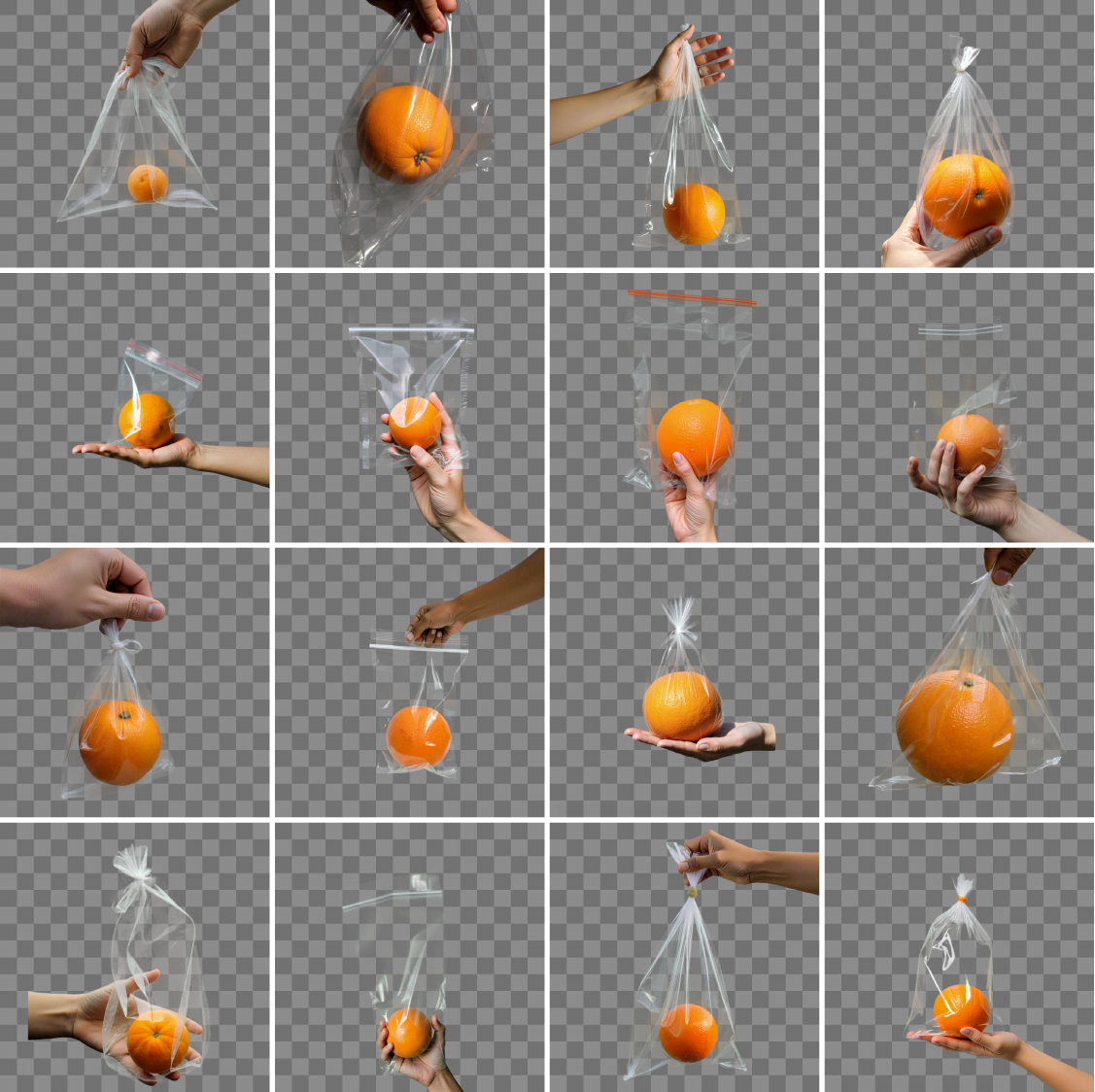}  
    \caption{A hand holds a plastic bag containing an orange.}
    \label{fig:appendix_9}
\end{figure}

\begin{figure}[t]
    \centering
    \includegraphics[width=\linewidth]{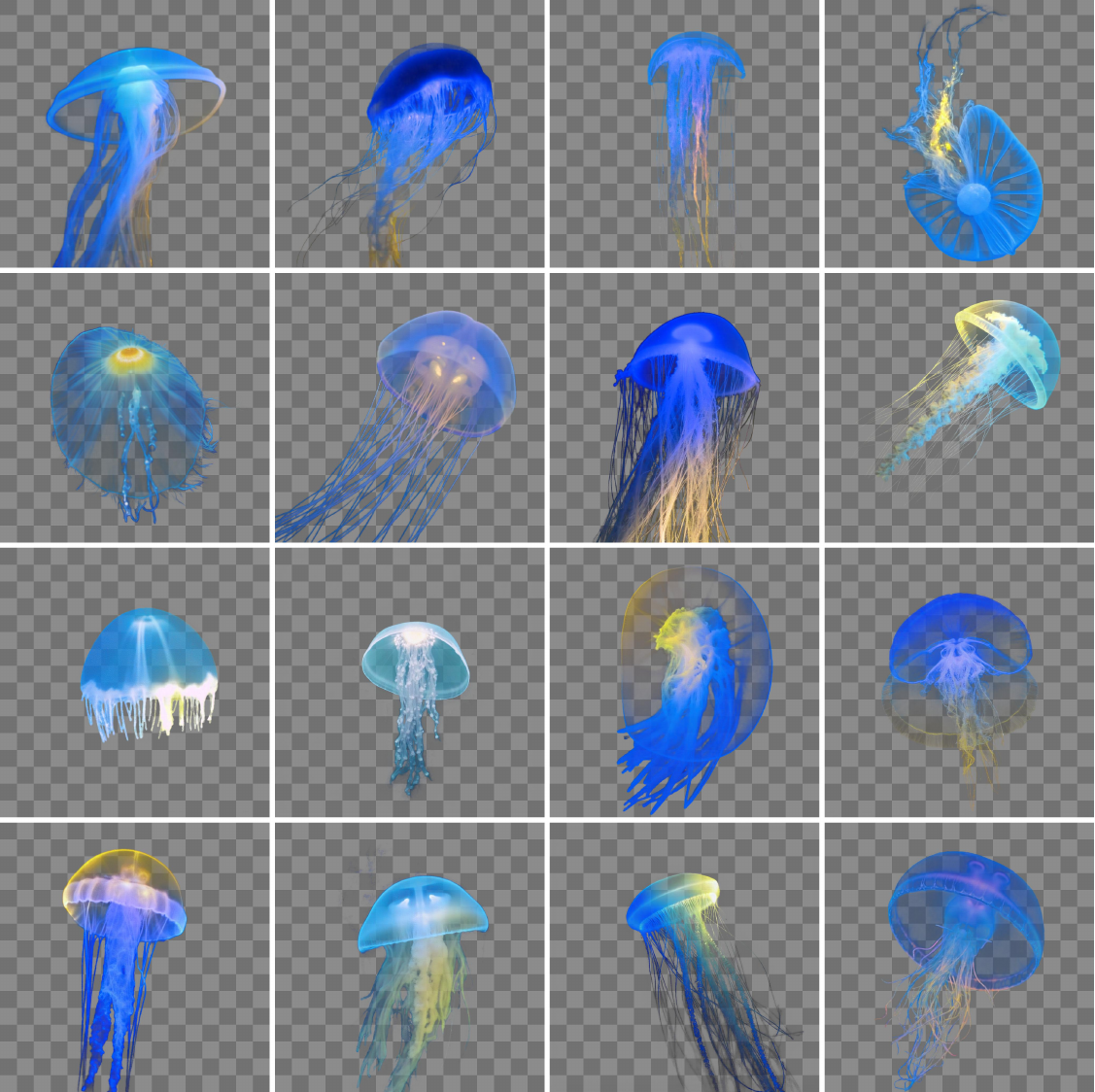}  
    \caption{A glowing jellyfish with translucent blue and yellow hues is shown.}
    \label{fig:appendix_10}
\end{figure}

\begin{figure}[t]
    \centering
    \includegraphics[width=\linewidth]{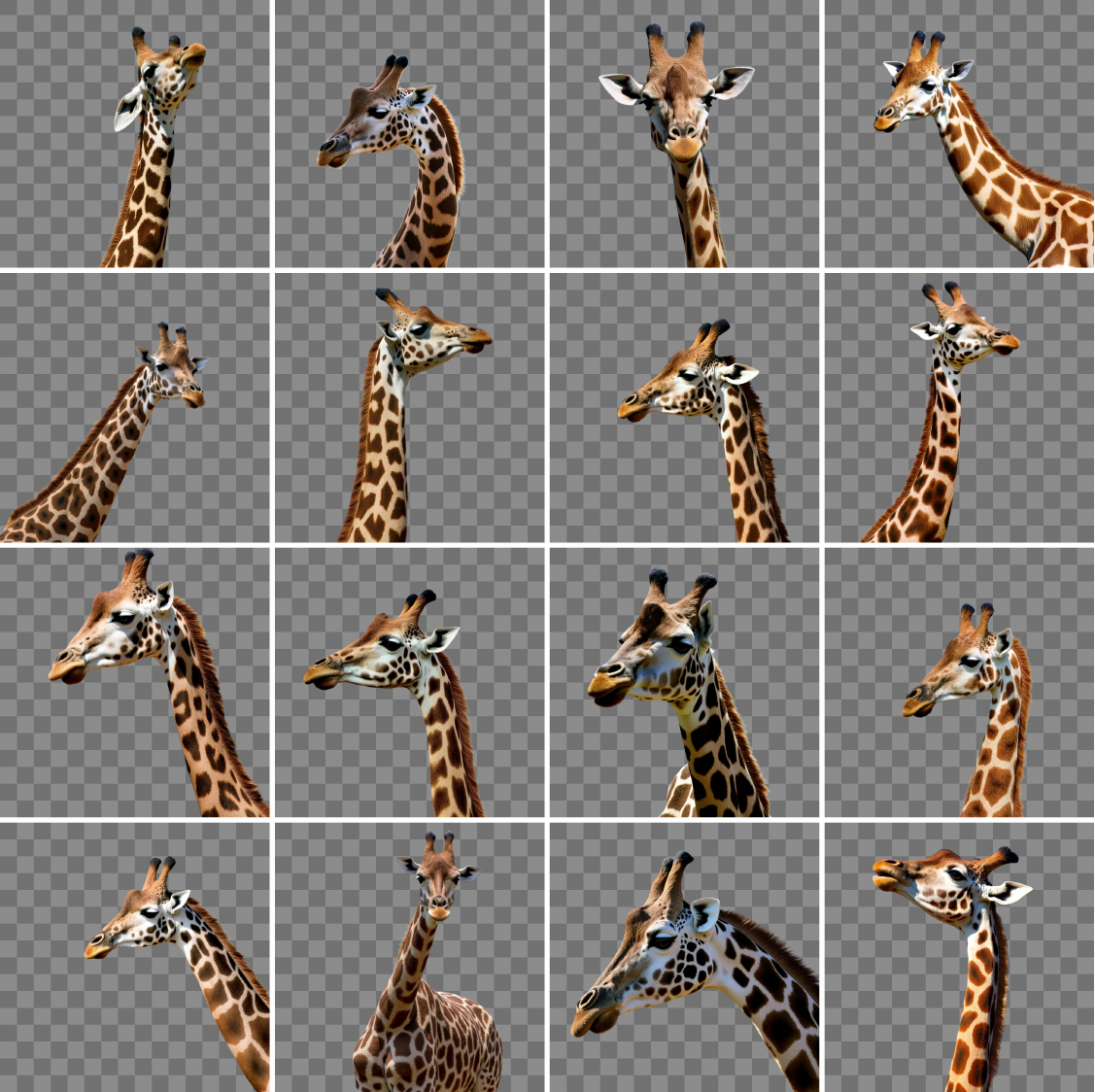}  
    \caption{A giraffe with its head turned to the side, showcasing its long neck and distinctive coat pattern.}
    \label{fig:appendix_11}
\end{figure}

\begin{figure}[t]
    \centering
    \includegraphics[width=\linewidth]{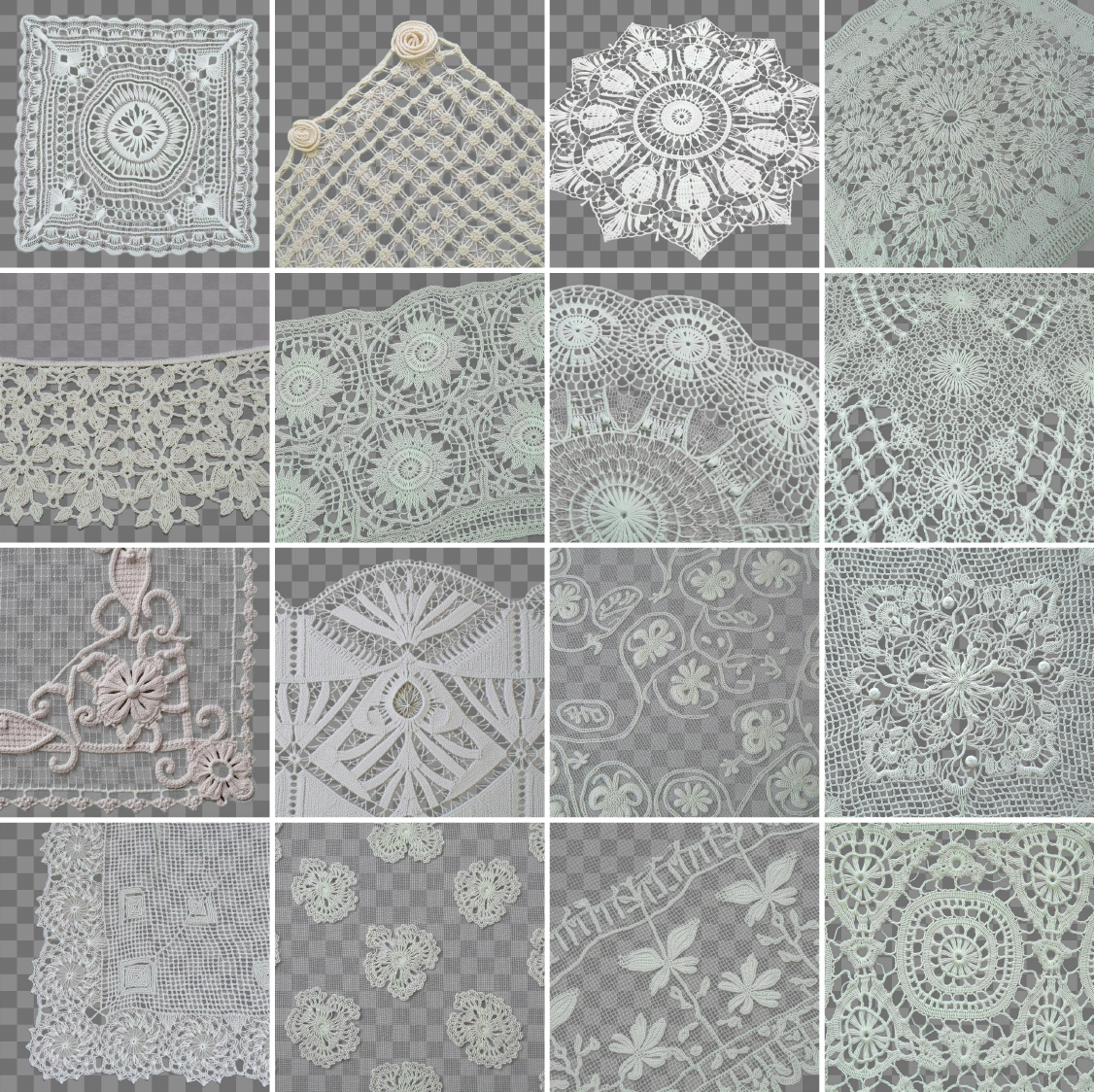}  
    \caption{A detailed, intricate crocheted pattern with floral motifs and openwork design.}
    \label{fig:appendix_12}
\end{figure}

\begin{figure}[t]
    \centering
    \includegraphics[width=\linewidth]{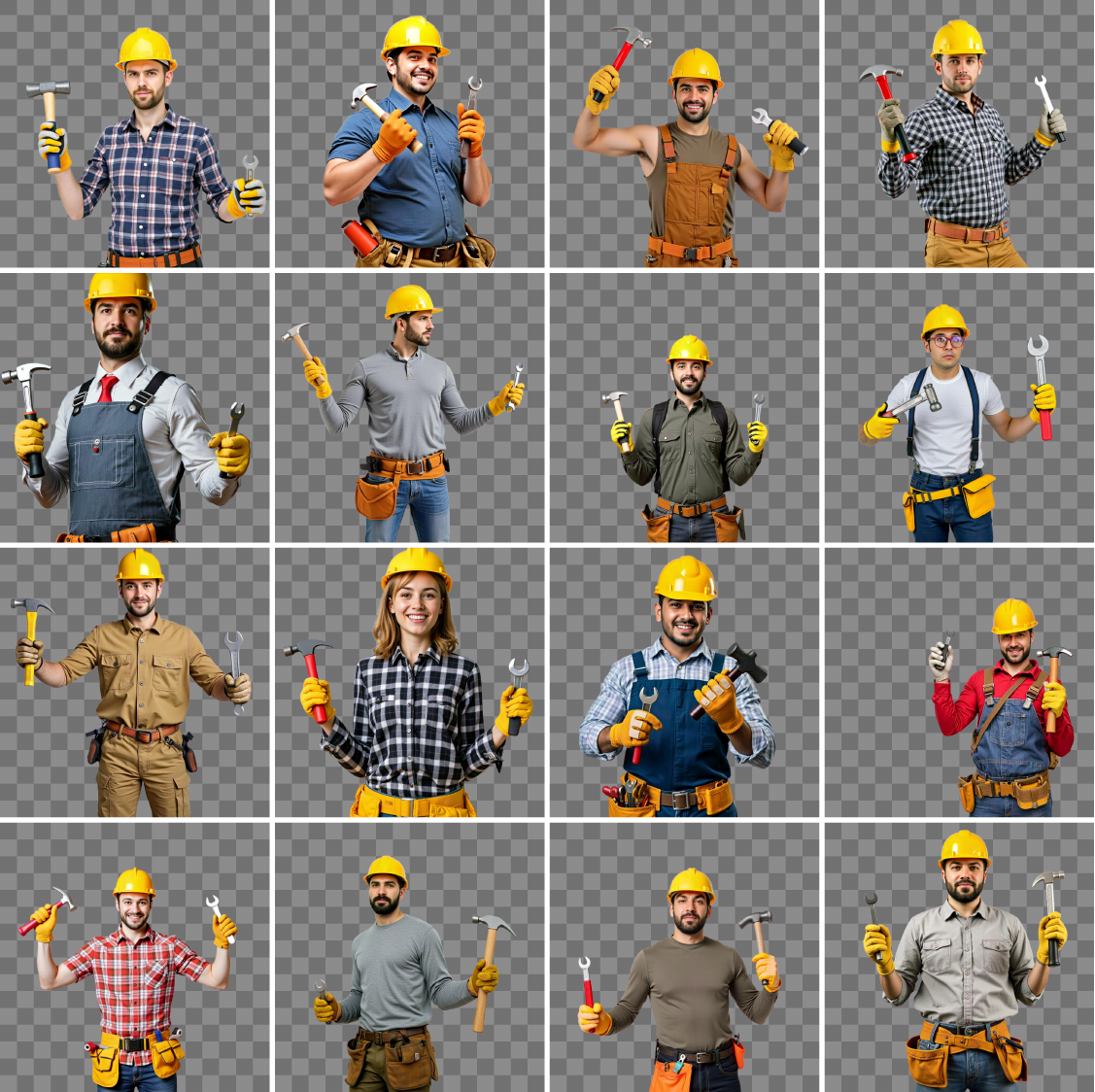}  
    \caption{A construction worker wearing a yellow hard hat, gloves, and tool belt holds a hammer and wrench.}
    \label{fig:appendix_13}
\end{figure}

\begin{figure}[t]
    \centering
    \includegraphics[width=\linewidth]{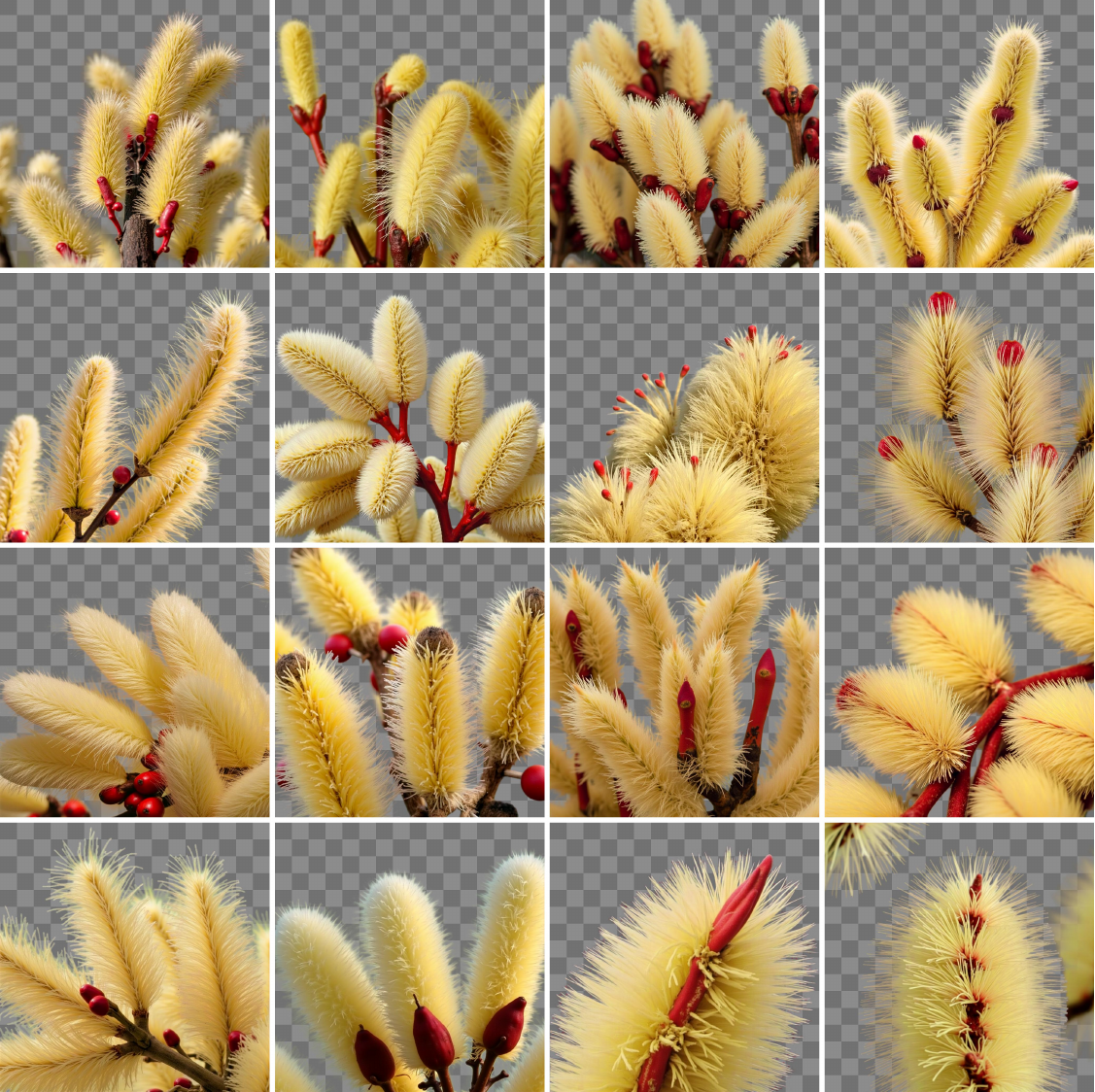}  
    \caption{A close-up of fluffy, yellowish-white catkins with red buds.}
    \label{fig:appendix_14}
\end{figure}

\begin{figure}[t]
    \centering
    \includegraphics[width=\linewidth]{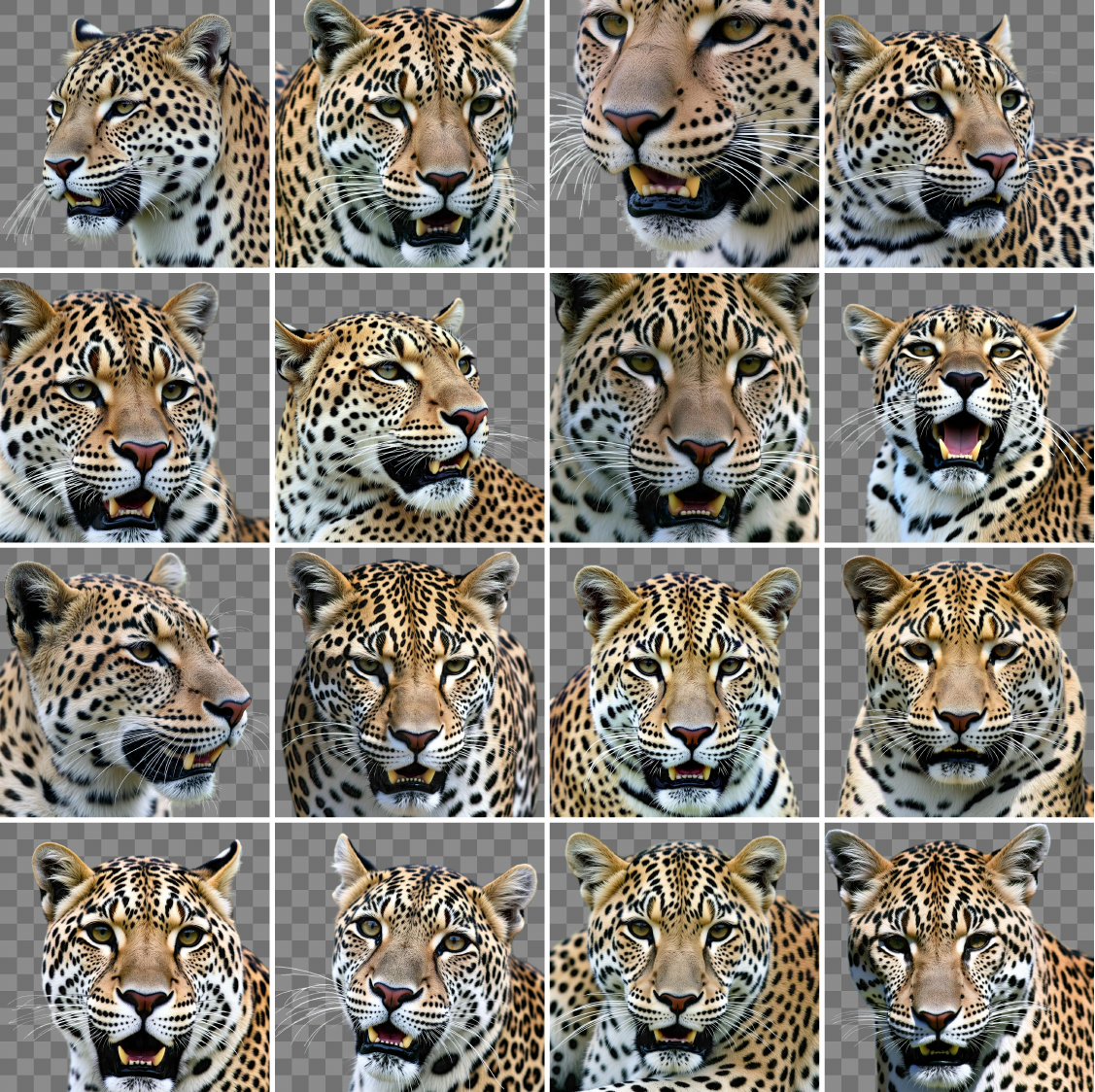}  
    \caption{A close-up of a leopard's face with its mouth slightly open, showcasing its sharp teeth and distinctive spotted fur.}
    \label{fig:appendix_15}
\end{figure}

\begin{figure}[t]
    \centering
    \includegraphics[width=\linewidth]{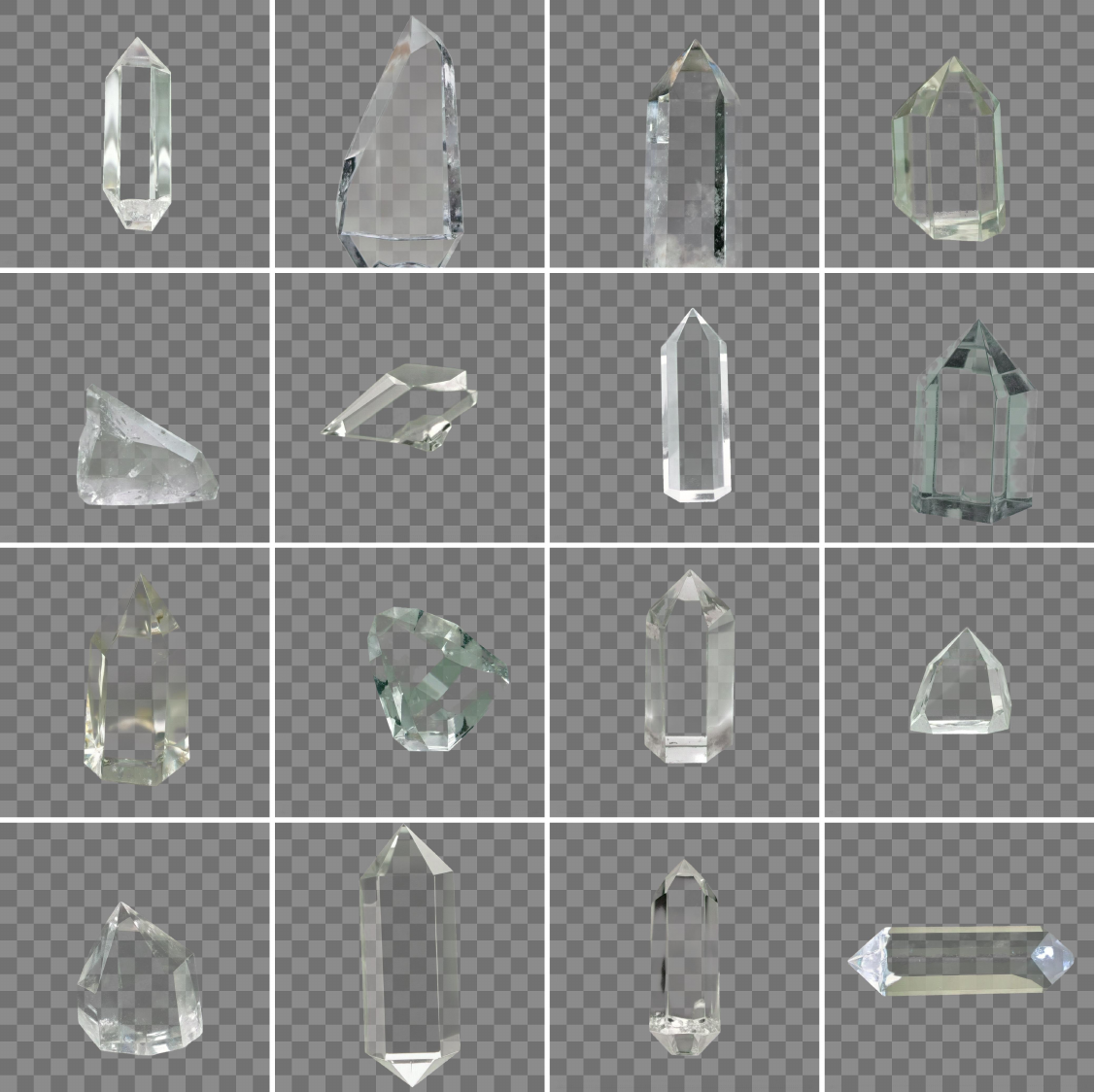}  
    \caption{A clear, faceted crystal with smooth edges and a pointed tip is shown.}
    \label{fig:appendix_16}
\end{figure}

\begin{figure}[t]
    \centering
    \includegraphics[width=\linewidth]{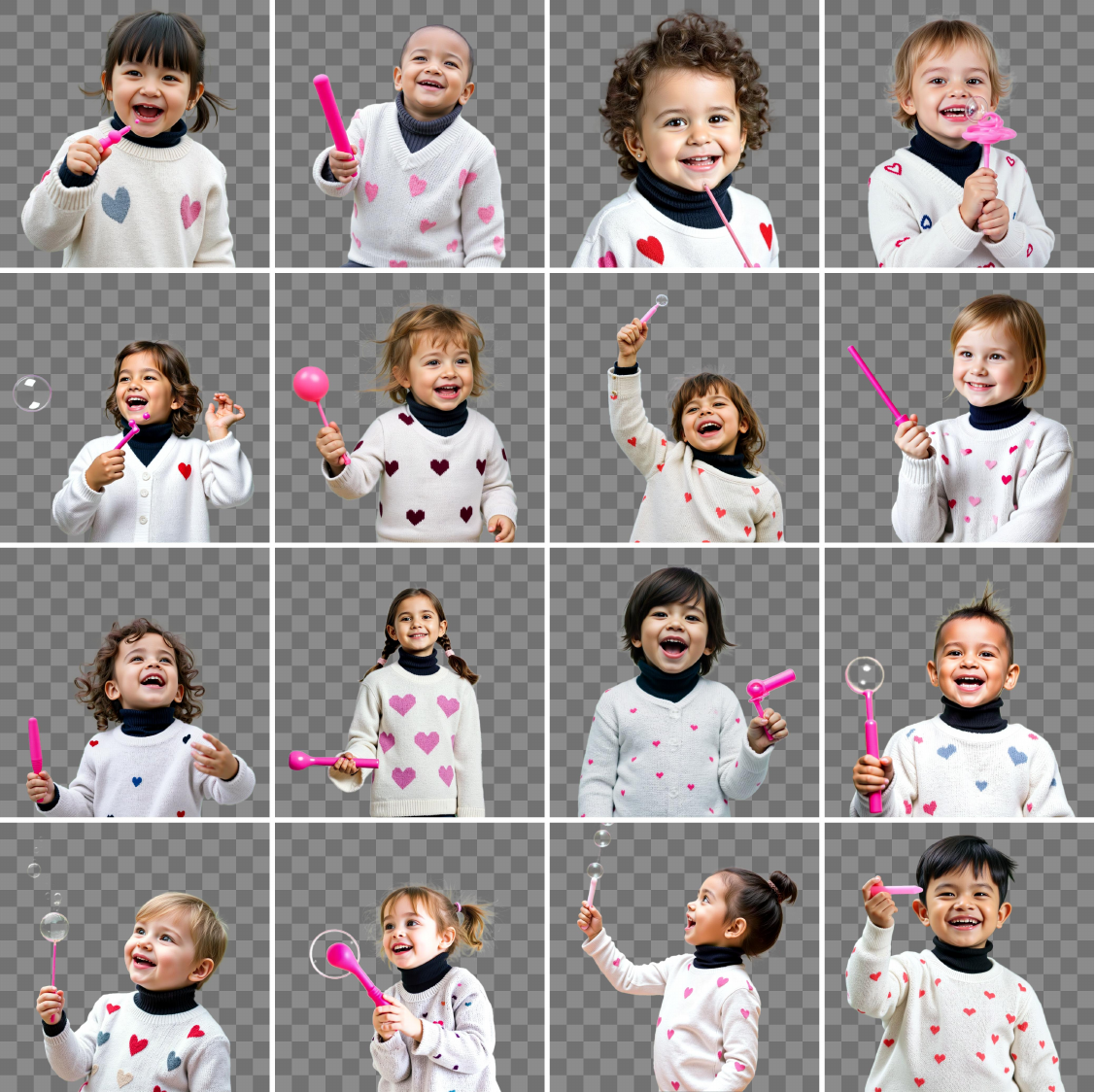}  
    \caption{A child is joyfully playing with a pink bubble wand, wearing a white sweater with heart patterns and a dark turtleneck.}
    \label{fig:appendix_17}
\end{figure}

\begin{figure}[t]
    \centering
    \includegraphics[width=\linewidth]{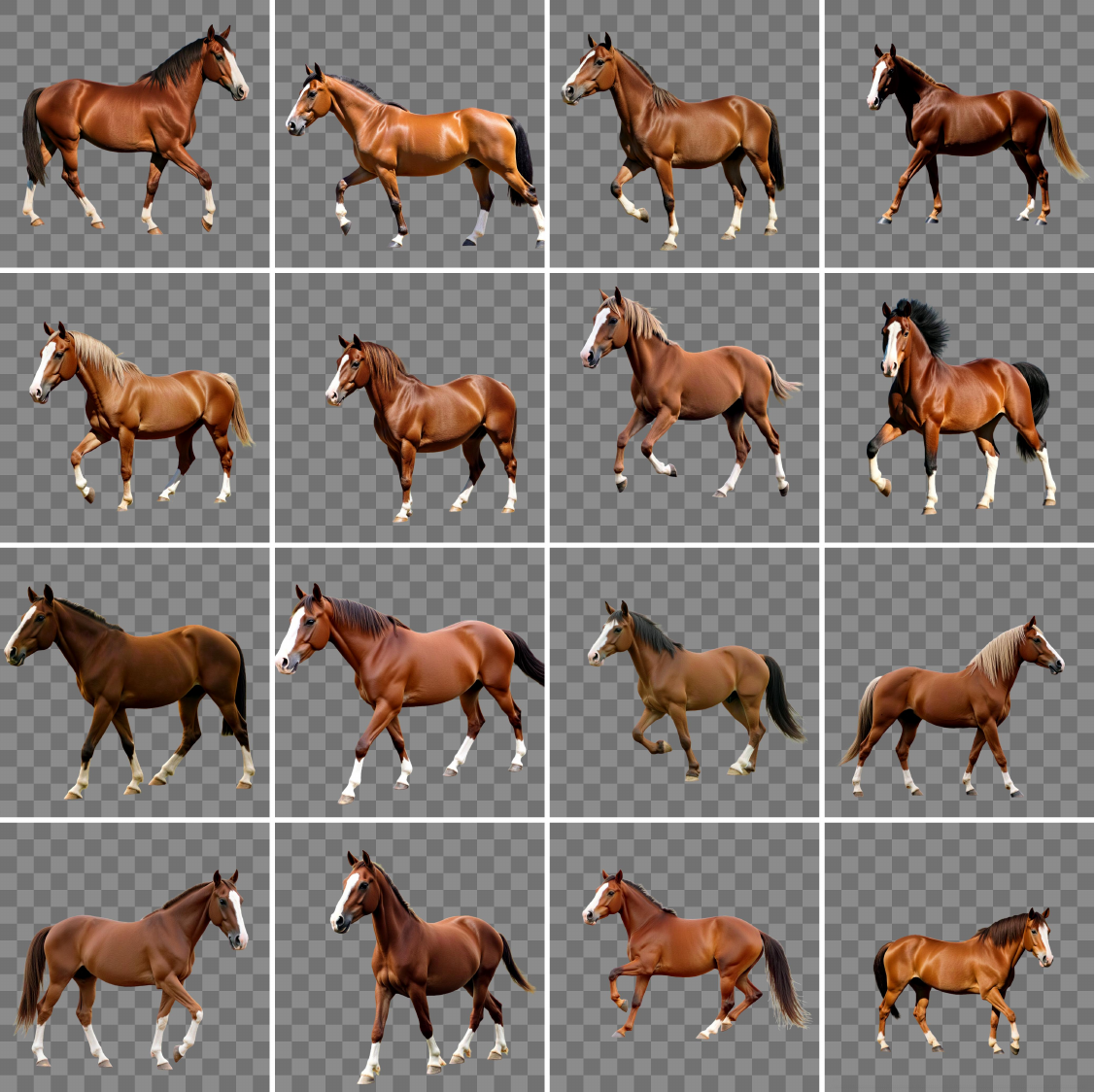}  
    \caption{A brown horse with white markings on its face and legs is captured mid-stride.}
    \label{fig:appendix_18}
\end{figure}

\begin{figure}[t]
    \centering
    \includegraphics[width=\linewidth]{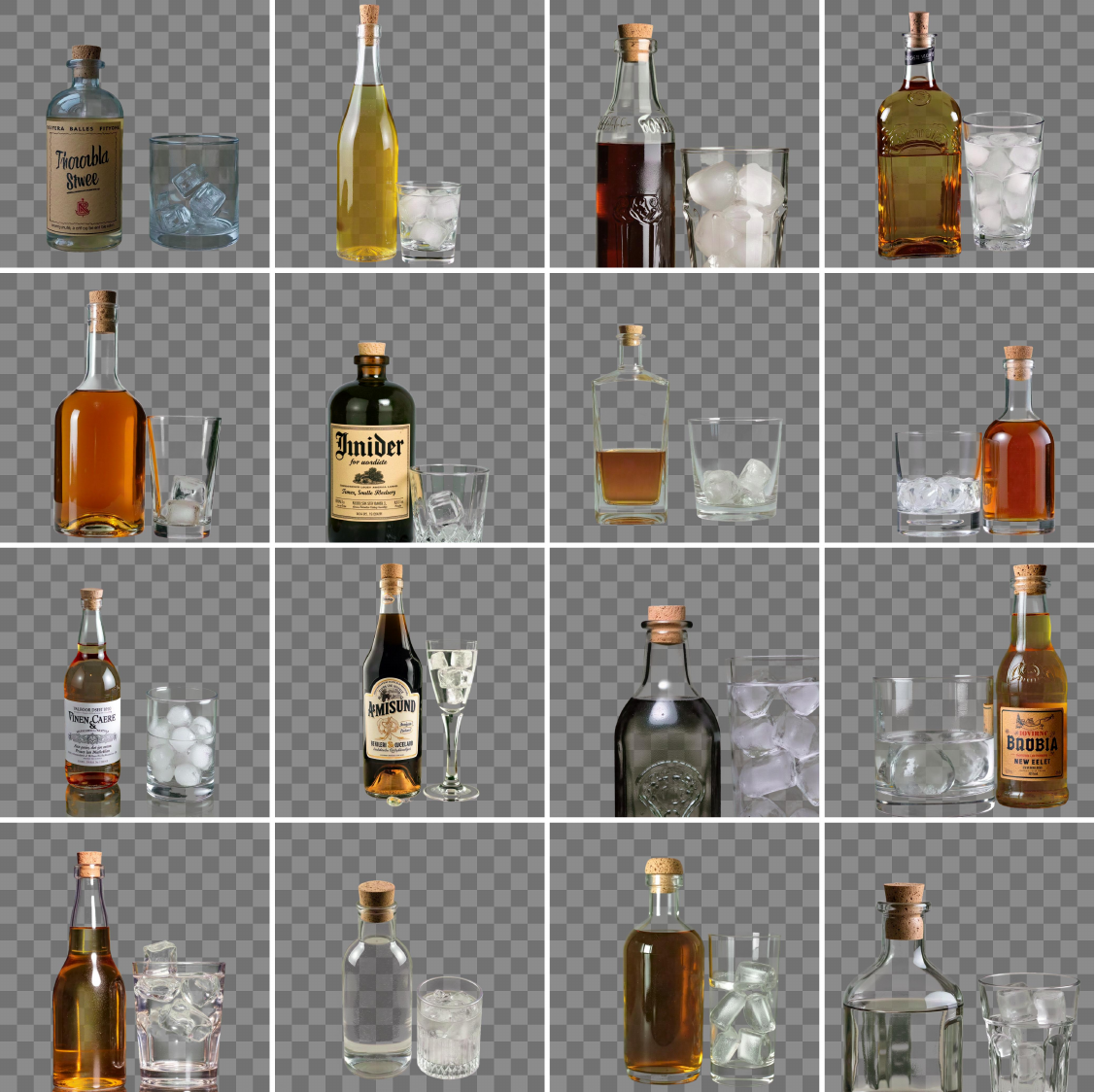}  
    \caption{A bottle with a cork and a glass with ice cubes are placed together.}
    \label{fig:appendix_19}
\end{figure}

\begin{figure}[t]
    \centering
    \includegraphics[width=\linewidth]{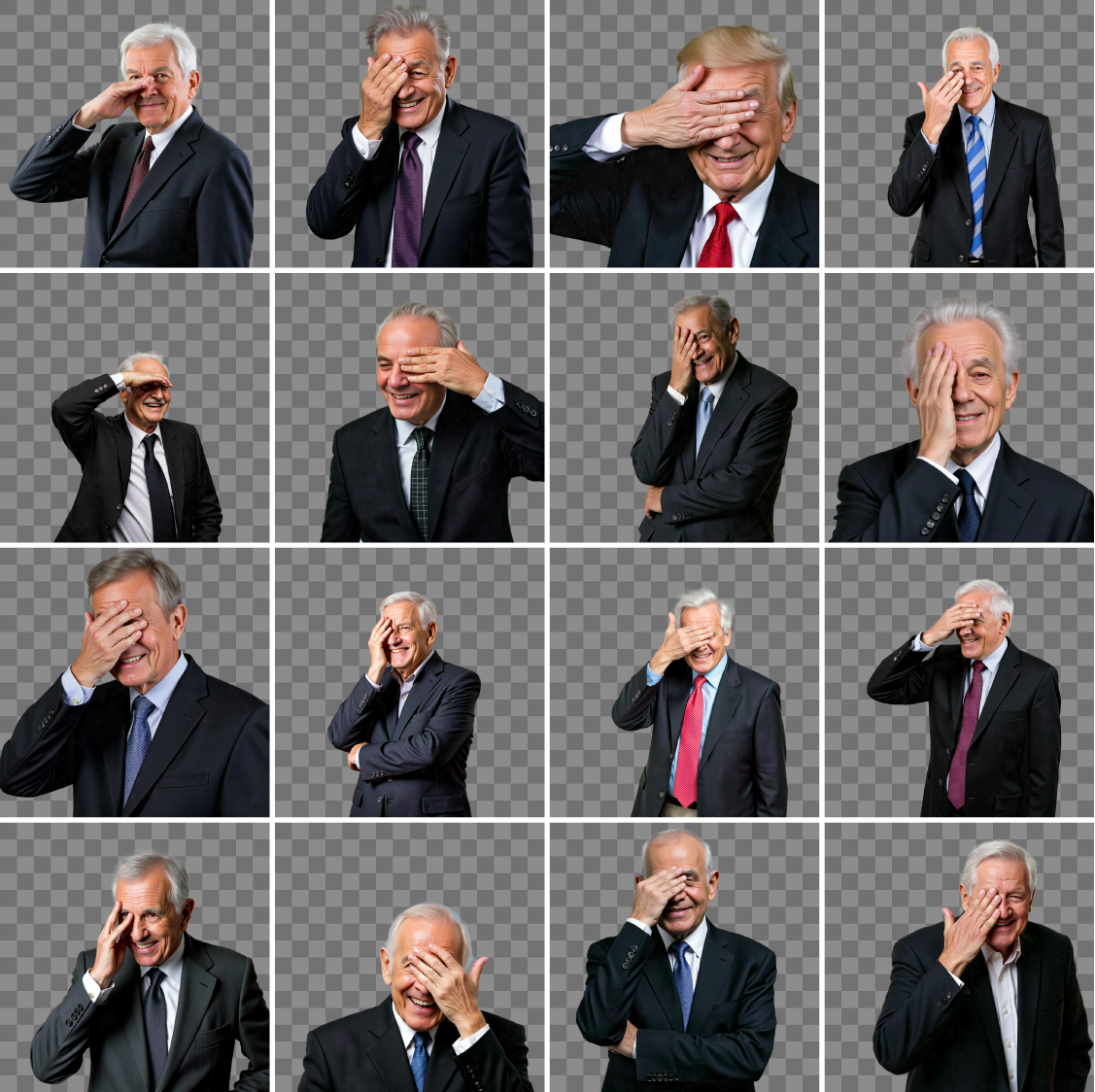}  
    \caption{An elderly man in a suit covers his eyes with his hand, smiling.}
    \label{fig:appendix_20}
\end{figure}

We present additional qualitative results from Figure~\ref{fig:appendix_1} to Figure~\ref{fig:appendix_20}, with all images rendered at a resolution of $1024\times1024$.

\section{Details of Experimental Results}
\label{app:tables}

We present details of the experimental results from Table~\ref{tab:transposed_metrics_1} to Table~\ref{tab:subset_transposed_metrics_26}.

\clearpage
\begin{table}[htbp]
  \centering
  \caption{Metric scores of Alpha test split across color categories on VAE of LayerDiffuse.}
  \label{tab:transposed_metrics_1}
  \resizebox{\textwidth}{!}{%
\begin{tabular}{l c c c c c c c c c c}
    \toprule
    \textbf{Metric} & \texttt{black} & \texttt{gray} & \texttt{white} & \texttt{red} & \texttt{yellow} & \texttt{green} & \texttt{cyan} & \texttt{blue} & \texttt{magenta} & \texttt{overall} \\
    \midrule
    PSNR & 32.2987 & 33.4605 & 32.2383 & 32.4273 & 32.3152 & 32.4010 & 32.3259 & 32.2476 & 32.3637 & 32.4531 \\
    SSIM & 0.9424 & 0.9497 & 0.9453 & 0.9481 & 0.9464 & 0.9498 & 0.9486 & 0.9457 & 0.9496 & 0.9473 \\
    LAION AES & 5.3193 & 5.2509 & 5.1818 & 5.0469 & 4.8619 & 4.9056 & 4.9785 & 4.9216 & 4.8089 & 5.0306 \\
    rFID & 7.5630 & 5.6691 & 7.3526 & 6.3528 & 6.7486 & 6.2935 & 6.1922 & 6.1664 & 6.0109 & 6.4832 \\
    LPIPS & 0.0368 & 0.0305 & 0.0355 & 0.0322 & 0.0330 & 0.0307 & 0.0314 & 0.0308 & 0.0308 & 0.0324 \\
    \bottomrule
  \end{tabular}}
\end{table}

\begin{table}[htbp]
  \centering
  \caption{Metric scores of Alpha test split across color categories on VAE of FLUX (ours).}
  \label{tab:transposed_metrics_2}
  \resizebox{\textwidth}{!}{%
\begin{tabular}{l c c c c c c c c c c}
    \toprule
    \textbf{Metric} & \texttt{black} & \texttt{gray} & \texttt{white} & \texttt{red} & \texttt{yellow} & \texttt{green} & \texttt{cyan} & \texttt{blue} & \texttt{magenta} & \texttt{overall} \\
    \midrule
    PSNR & 38.1086 & 39.3765 & 37.8794 & 38.2636 & 37.9722 & 38.2767 & 38.0581 & 37.8197 & 38.0331 & 38.1987 \\
    SSIM & 0.9763 & 0.9802 & 0.9784 & 0.9796 & 0.9792 & 0.9804 & 0.9801 & 0.9785 & 0.9803 & 0.9792 \\
    LAION AES & 5.3554 & 5.2547 & 5.2045 & 5.0825 & 4.9082 & 4.9341 & 5.0104 & 4.9599 & 4.8444 & 5.0616 \\
    rFID & 1.8807 & 1.9358 & 2.6166 & 1.3276 & 1.9337 & 1.2855 & 1.6091 & 1.2121 & 1.1391 & 1.6600 \\
    LPIPS & 0.0187 & 0.0161 & 0.0181 & 0.0139 & 0.0139 & 0.0116 & 0.0128 & 0.0135 & 0.0122 & 0.0145 \\
    \bottomrule
  \end{tabular}}
\end{table}

\begin{table}[htbp]
  \centering
  \caption{Metric scores of Alpha test split across color categories on VAE of FLUX (w/o Norm KL).}
  \label{tab:transposed_metrics_3}
  \resizebox{\textwidth}{!}{%
\begin{tabular}{l c c c c c c c c c c}
    \toprule
    \textbf{Metric} & \texttt{black} & \texttt{gray} & \texttt{white} & \texttt{red} & \texttt{yellow} & \texttt{green} & \texttt{cyan} & \texttt{blue} & \texttt{magenta} & \texttt{overall} \\
    \midrule
    PSNR & 37.3068 & 38.8858 & 36.9265 & 37.4930 & 37.1216 & 37.4400 & 37.1503 & 36.9786 & 37.1892 & 37.3880 \\
    SSIM & 0.9732 & 0.9789 & 0.9761 & 0.9785 & 0.9772 & 0.9792 & 0.9787 & 0.9770 & 0.9792 & 0.9776 \\
    LAION AES & 5.3538 & 5.2504 & 5.1904 & 5.0873 & 4.9319 & 4.9275 & 5.0210 & 4.9620 & 4.8512 & 5.0639 \\
    rFID & 2.7818 & 2.8000 & 3.9974 & 1.9073 & 3.0712 & 1.9013 & 2.5008 & 1.7790 & 1.6696 & 2.4898 \\
    LPIPS & 0.0216 & 0.0191 & 0.0217 & 0.0167 & 0.0169 & 0.0141 & 0.0155 & 0.0160 & 0.0146 & 0.0174 \\
    \bottomrule
  \end{tabular}}
\end{table}

\begin{table}[htbp]
  \centering
  \caption{Metric scores of Alpha test split across color categories on VAE of FLUX (w/o Ref KL).}
  \label{tab:transposed_metrics_4}
  \resizebox{\textwidth}{!}{%
\begin{tabular}{l c c c c c c c c c c}
    \toprule
    \textbf{Metric} & \texttt{black} & \texttt{gray} & \texttt{white} & \texttt{red} & \texttt{yellow} & \texttt{green} & \texttt{cyan} & \texttt{blue} & \texttt{magenta} & \texttt{overall} \\
    \midrule
    PSNR & 38.0446 & 39.3336 & 37.8078 & 38.1920 & 37.8873 & 38.2212 & 38.0218 & 37.7860 & 37.9905 & 38.1428 \\
    SSIM & 0.9758 & 0.9800 & 0.9781 & 0.9794 & 0.9789 & 0.9802 & 0.9799 & 0.9782 & 0.9801 & 0.9790 \\
    LAION AES & 5.3596 & 5.2505 & 5.2010 & 5.0858 & 4.9156 & 4.9281 & 5.0128 & 4.9661 & 4.8519 & 5.0635 \\
    rFID & 2.0329 & 2.0735 & 2.9343 & 1.3408 & 2.2037 & 1.3283 & 1.7552 & 1.3025 & 1.1790 & 1.7945 \\
    LPIPS & 0.0199 & 0.0170 & 0.0191 & 0.0149 & 0.0148 & 0.0124 & 0.0135 & 0.0145 & 0.0130 & 0.0155 \\
    \bottomrule
  \end{tabular}}
\end{table}

\begin{table}[htbp]
  \centering
  \caption{Metric scores of Alpha test split across color categories on VAE of FLUX (w/o GAN).}
  \label{tab:transposed_metrics_5}
  \resizebox{\textwidth}{!}{%
\begin{tabular}{l c c c c c c c c c c}
    \toprule
    \textbf{Metric} & \texttt{black} & \texttt{gray} & \texttt{white} & \texttt{red} & \texttt{yellow} & \texttt{green} & \texttt{cyan} & \texttt{blue} & \texttt{magenta} & \texttt{overall} \\
    \midrule
    PSNR & 38.3982 & 39.7655 & 38.1507 & 38.5581 & 38.2452 & 38.5394 & 38.3143 & 38.0783 & 38.3053 & 38.4839 \\
    SSIM & 0.9773 & 0.9811 & 0.9793 & 0.9807 & 0.9801 & 0.9814 & 0.9811 & 0.9795 & 0.9813 & 0.9802 \\
    LAION AES & 5.3435 & 5.2575 & 5.1959 & 5.0754 & 4.9003 & 4.9262 & 5.0056 & 4.9532 & 4.8428 & 5.0556 \\
    rFID & 1.8812 & 2.1725 & 2.6998 & 1.2819 & 1.9951 & 1.3422 & 1.6618 & 1.1954 & 1.1517 & 1.7091 \\
    LPIPS & 0.0195 & 0.0169 & 0.0193 & 0.0148 & 0.0150 & 0.0125 & 0.0138 & 0.0143 & 0.0129 & 0.0154 \\
    \bottomrule
  \end{tabular}}
\end{table}

\begin{table}[htbp]
  \centering
  \caption{Metric scores of Alpha test split across color categories on VAE of FLUX (w/o LPIPS).}
  \label{tab:transposed_metrics_6}
  \resizebox{\textwidth}{!}{%
\begin{tabular}{l c c c c c c c c c c}
    \toprule
    \textbf{Metric} & \texttt{black} & \texttt{gray} & \texttt{white} & \texttt{red} & \texttt{yellow} & \texttt{green} & \texttt{cyan} & \texttt{blue} & \texttt{magenta} & \texttt{overall} \\
    \midrule
    PSNR & 36.5276 & 38.4143 & 36.0067 & 36.6541 & 36.2457 & 36.5500 & 36.2184 & 36.1114 & 36.2966 & 36.5583 \\
    SSIM & 0.9710 & 0.9779 & 0.9747 & 0.9772 & 0.9761 & 0.9782 & 0.9777 & 0.9754 & 0.9781 & 0.9763 \\
    LAION AES & 5.3500 & 5.2426 & 5.1774 & 5.0790 & 4.9108 & 4.9194 & 5.0064 & 4.9554 & 4.8430 & 5.0538 \\
    rFID & 3.6630 & 3.1768 & 4.4897 & 2.6961 & 3.6317 & 2.6668 & 3.0846 & 2.4477 & 2.3424 & 3.1332 \\
    LPIPS & 0.0236 & 0.0201 & 0.0231 & 0.0193 & 0.0194 & 0.0167 & 0.0182 & 0.0184 & 0.0172 & 0.0196 \\
    \bottomrule
  \end{tabular}}
\end{table}

\begin{table}[htbp]
  \centering
  \caption{Metric scores of Alpha test split across color categories on VAE of FLUX (w/o ABMSE).}
  \label{tab:transposed_metrics_7}
  \resizebox{\textwidth}{!}{%
\begin{tabular}{l c c c c c c c c c c}
    \toprule
    \textbf{Metric} & \texttt{black} & \texttt{gray} & \texttt{white} & \texttt{red} & \texttt{yellow} & \texttt{green} & \texttt{cyan} & \texttt{blue} & \texttt{magenta} & \texttt{overall} \\
    \midrule
    PSNR & 38.0177 & 39.1793 & 37.7778 & 38.1497 & 37.8376 & 38.1371 & 37.9670 & 37.7438 & 37.9519 & 38.0847 \\
    SSIM & 0.9758 & 0.9791 & 0.9775 & 0.9789 & 0.9783 & 0.9796 & 0.9793 & 0.9779 & 0.9795 & 0.9784 \\
    LAION AES & 5.3491 & 5.2584 & 5.2022 & 5.0804 & 4.9057 & 4.9268 & 5.0043 & 4.9578 & 4.8455 & 5.0589 \\
    rFID & 1.8418 & 2.0215 & 2.6620 & 1.2887 & 1.8918 & 1.3379 & 1.5880 & 1.1761 & 1.1315 & 1.6599 \\
    LPIPS & 0.0189 & 0.0172 & 0.0188 & 0.0142 & 0.0145 & 0.0120 & 0.0131 & 0.0136 & 0.0124 & 0.0150 \\
    \bottomrule
  \end{tabular}}
\end{table}

\begin{table}[htbp]
  \centering
  \caption{Metric scores of Alpha test split across color categories on VAE of SDXL (ours).}
  \label{tab:transposed_metrics_8}
  \resizebox{\textwidth}{!}{%
\begin{tabular}{l c c c c c c c c c c}
    \toprule
    \textbf{Metric} & \texttt{black} & \texttt{gray} & \texttt{white} & \texttt{red} & \texttt{yellow} & \texttt{green} & \texttt{cyan} & \texttt{blue} & \texttt{magenta} & \texttt{overall} \\
    \midrule
    PSNR & 35.5273 & 36.5843 & 35.1748 & 35.6110 & 35.3228 & 35.6430 & 35.4329 & 35.3176 & 35.4235 & 35.5597 \\
    SSIM & 0.9560 & 0.9618 & 0.9591 & 0.9613 & 0.9602 & 0.9621 & 0.9617 & 0.9600 & 0.9620 & 0.9605 \\
    LAION AES & 5.3394 & 5.2481 & 5.1973 & 5.0617 & 4.9121 & 4.9081 & 5.0042 & 4.9470 & 4.8323 & 5.0500 \\
    rFID & 4.2172 & 4.0979 & 4.9661 & 3.2091 & 4.2911 & 3.2912 & 3.6115 & 3.0034 & 2.8873 & 3.7305 \\
    LPIPS & 0.0464 & 0.0446 & 0.0469 & 0.0390 & 0.0396 & 0.0352 & 0.0374 & 0.0371 & 0.0355 & 0.0402 \\
    \bottomrule
  \end{tabular}}
\end{table}

\begin{table}[htbp]
  \centering
  \caption{Metric scores of Alpha test split across color categories on VAE of SDXL (w/o Norm KL).}
  \label{tab:transposed_metrics_9}
  \resizebox{\textwidth}{!}{%
\begin{tabular}{l c c c c c c c c c c}
    \toprule
    \textbf{Metric} & \texttt{black} & \texttt{gray} & \texttt{white} & \texttt{red} & \texttt{yellow} & \texttt{green} & \texttt{cyan} & \texttt{blue} & \texttt{magenta} & \texttt{overall} \\
    \midrule
    PSNR & 35.4681 & 36.5361 & 35.1141 & 35.5533 & 35.2700 & 35.5798 & 35.3618 & 35.2497 & 35.3532 & 35.4985 \\
    SSIM & 0.9564 & 0.9617 & 0.9591 & 0.9611 & 0.9602 & 0.9620 & 0.9616 & 0.9598 & 0.9619 & 0.9604 \\
    LAION AES & 5.3420 & 5.2439 & 5.1978 & 5.0637 & 4.9034 & 4.9057 & 4.9980 & 4.9451 & 4.8281 & 5.0475 \\
    rFID & 4.2193 & 4.3088 & 5.3154 & 3.3007 & 4.4657 & 3.3526 & 3.7425 & 3.1065 & 2.9439 & 3.8617 \\
    LPIPS & 0.0463 & 0.0446 & 0.0466 & 0.0390 & 0.0395 & 0.0352 & 0.0374 & 0.0371 & 0.0355 & 0.0401 \\
    \bottomrule
  \end{tabular}}
\end{table}

\begin{table}[htbp]
  \centering
  \caption{Metric scores of Alpha test split across color categories on VAE of SDXL (w/o Ref KL).}
  \label{tab:transposed_metrics_10}
  \resizebox{\textwidth}{!}{%
\begin{tabular}{l c c c c c c c c c c}
    \toprule
    \textbf{Metric} & \texttt{black} & \texttt{gray} & \texttt{white} & \texttt{red} & \texttt{yellow} & \texttt{green} & \texttt{cyan} & \texttt{blue} & \texttt{magenta} & \texttt{overall} \\
    \midrule
    PSNR & 35.4490 & 36.4958 & 35.1270 & 35.5450 & 35.2833 & 35.5766 & 35.3628 & 35.2377 & 35.3482 & 35.4917 \\
    SSIM & 0.9557 & 0.9615 & 0.9590 & 0.9610 & 0.9602 & 0.9618 & 0.9615 & 0.9596 & 0.9617 & 0.9602 \\
    LAION AES & 5.3399 & 5.2404 & 5.1905 & 5.0652 & 4.9079 & 4.9075 & 5.0016 & 4.9487 & 4.8343 & 5.0484 \\
    rFID & 4.1823 & 4.3080 & 5.1787 & 3.2270 & 4.2665 & 3.2916 & 3.6771 & 3.0593 & 2.9209 & 3.7902 \\
    LPIPS & 0.0452 & 0.0439 & 0.0456 & 0.0381 & 0.0386 & 0.0344 & 0.0366 & 0.0362 & 0.0348 & 0.0393 \\
    \bottomrule
  \end{tabular}}
\end{table}

\begin{table}[htbp]
  \centering
  \caption{Metric scores of Alpha test split across color categories on VAE of SDXL (w/o GAN).}
  \label{tab:transposed_metrics_11}
  \resizebox{\textwidth}{!}{%
\begin{tabular}{l c c c c c c c c c c}
    \toprule
    \textbf{Metric} & \texttt{black} & \texttt{gray} & \texttt{white} & \texttt{red} & \texttt{yellow} & \texttt{green} & \texttt{cyan} & \texttt{blue} & \texttt{magenta} & \texttt{overall} \\
    \midrule
    PSNR & 35.6458 & 36.7020 & 35.3358 & 35.7219 & 35.4739 & 35.7754 & 35.5738 & 35.4336 & 35.5357 & 35.6887 \\
    SSIM & 0.9571 & 0.9622 & 0.9598 & 0.9617 & 0.9609 & 0.9625 & 0.9622 & 0.9604 & 0.9624 & 0.9610 \\
    LAION AES & 5.3344 & 5.2463 & 5.1910 & 5.0589 & 4.8991 & 4.9111 & 5.0043 & 4.9456 & 4.8293 & 5.0467 \\
    rFID & 4.0130 & 4.3954 & 5.0296 & 3.1645 & 4.0869 & 3.2718 & 3.4708 & 2.9374 & 2.8250 & 3.6883 \\
    LPIPS & 0.0468 & 0.0448 & 0.0468 & 0.0394 & 0.0398 & 0.0355 & 0.0377 & 0.0374 & 0.0359 & 0.0405 \\
    \bottomrule
  \end{tabular}}
\end{table}

\begin{table}[htbp]
  \centering
  \caption{Metric scores of Alpha test split across color categories on VAE of SDXL (w/o LPIPS).}
  \label{tab:transposed_metrics_12}
  \resizebox{\textwidth}{!}{%
\begin{tabular}{l c c c c c c c c c c}
    \toprule
    \textbf{Metric} & \texttt{black} & \texttt{gray} & \texttt{white} & \texttt{red} & \texttt{yellow} & \texttt{green} & \texttt{cyan} & \texttt{blue} & \texttt{magenta} & \texttt{overall} \\
    \midrule
    PSNR & 35.4575 & 36.5327 & 35.1170 & 35.6070 & 35.2386 & 35.4740 & 35.3096 & 35.2734 & 35.4419 & 35.4946 \\
    SSIM & 0.9561 & 0.9616 & 0.9591 & 0.9612 & 0.9602 & 0.9619 & 0.9615 & 0.9598 & 0.9619 & 0.9604 \\
    LAION AES & 5.3446 & 5.2464 & 5.1980 & 5.0667 & 4.9047 & 4.9006 & 4.9996 & 4.9498 & 4.8342 & 5.0494 \\
    rFID & 4.1360 & 4.2773 & 5.1766 & 3.1705 & 4.2642 & 3.4594 & 3.6515 & 2.9238 & 2.8485 & 3.7675 \\
    LPIPS & 0.0455 & 0.0441 & 0.0461 & 0.0382 & 0.0392 & 0.0350 & 0.0372 & 0.0364 & 0.0348 & 0.0396 \\
    \bottomrule
  \end{tabular}}
\end{table}

\begin{table}[htbp]
  \centering
  \caption{Metric scores of Alpha test split across color categories on VAE of SDXL (w/o ABMSE).}
  \label{tab:transposed_metrics_13}
  \resizebox{\textwidth}{!}{%
\begin{tabular}{l c c c c c c c c c c}
    \toprule
    \textbf{Metric} & \texttt{black} & \texttt{gray} & \texttt{white} & \texttt{red} & \texttt{yellow} & \texttt{green} & \texttt{cyan} & \texttt{blue} & \texttt{magenta} & \texttt{overall} \\
    \midrule
    PSNR & 35.3931 & 36.1835 & 35.0748 & 35.4523 & 35.1950 & 35.4707 & 35.3176 & 35.2164 & 35.2939 & 35.3997 \\
    SSIM & 0.9562 & 0.9594 & 0.9566 & 0.9596 & 0.9581 & 0.9600 & 0.9596 & 0.9585 & 0.9602 & 0.9587 \\
    LAION AES & 5.3486 & 5.2545 & 5.1987 & 5.0779 & 4.9230 & 4.9194 & 5.0173 & 4.9556 & 4.8473 & 5.0603 \\
    rFID & 3.8295 & 4.4907 & 5.2114 & 3.1326 & 4.2212 & 3.3066 & 3.6059 & 2.8809 & 2.8459 & 3.7250 \\
    LPIPS & 0.0442 & 0.0458 & 0.0477 & 0.0376 & 0.0388 & 0.0339 & 0.0361 & 0.0355 & 0.0344 & 0.0393 \\
    \bottomrule
  \end{tabular}}
\end{table}

%%%%%%%%%%%%%%%%

\begin{table}[htbp]
  \centering
  \caption{Metric scores across subtypes of AIM and color categories on VAE of LayerDiffuse.}
  \label{tab:subset_transposed_metrics_14}
  \resizebox{\textwidth}{!}{%
\begin{tabular}{c c c c c c c c c c c c}
    \toprule
    \multirow{2.5}{*}{\textbf{Subtype}} & \multirow{2.5}{*}{\textbf{Metric}} & \multicolumn{10}{c}{\textbf{Colors}} \\ \cmidrule{3-12}
     &  & \texttt{black} & \texttt{gray} & \texttt{white} & \texttt{red} & \texttt{yellow} & \texttt{green} & \texttt{cyan} & \texttt{blue} & \texttt{magenta} & \texttt{overall} \\
    \midrule
    \multirow{5}{*}{\texttt{portrait}} & \multicolumn{1}{|l}{PSNR} & 33.0302 & 35.1074 & 32.9291 & 33.3363 & 33.1247 & 33.1945 & 33.0419 & 32.9910 & 33.1967 & 33.3280 \\
    & \multicolumn{1}{|l}{SSIM} & 0.9484 & 0.9582 & 0.9535 & 0.9563 & 0.9544 & 0.9581 & 0.9567 & 0.9532 & 0.9579 & 0.9552 \\
    & \multicolumn{1}{|l}{LAION AES} & 5.0925 & 5.0978 & 5.0091 & 5.0252 & 4.8070 & 4.9609 & 4.9370 & 4.8550 & 4.8635 & 4.9609 \\
    & \multicolumn{1}{|l}{rFID} & 22.1295 & 16.8871 & 17.0631 & 15.9141 & 17.0627 & 13.7088 & 14.4707 & 15.4059 & 14.2083 & 16.3167 \\
    & \multicolumn{1}{|l}{LPIPS} & 0.0431 & 0.0348 & 0.0412 & 0.0374 & 0.0382 & 0.0345 & 0.0356 & 0.0351 & 0.0348 & 0.0372 \\
    \cmidrule{1-12}
    \multirow{5}{*}{\texttt{transparent}} & \multicolumn{1}{|l}{PSNR} & 28.7389 & 37.2007 & 28.2495 & 29.2511 & 28.7266 & 29.0668 & 28.7697 & 28.7344 & 28.9971 & 29.7483 \\
\bottomrule
  \end{tabular}}
\end{table}

\begin{table}[htbp]
  \centering
  \caption{Metric scores across subtypes of AIM and color categories on VAE of FLUX (ours).}
  \label{tab:subset_transposed_metrics_15}
  \resizebox{\textwidth}{!}{%
\begin{tabular}{c c c c c c c c c c c c}
    \toprule
    \multirow{2.5}{*}{\textbf{Subtype}} & \multirow{2.5}{*}{\textbf{Metric}} & \multicolumn{10}{c}{\textbf{Colors}} \\ \cmidrule{3-12}
     &  & \texttt{black} & \texttt{gray} & \texttt{white} & \texttt{red} & \texttt{yellow} & \texttt{green} & \texttt{cyan} & \texttt{blue} & \texttt{magenta} & \texttt{overall} \\
    \midrule
    \multirow{5}{*}{\texttt{portrait}} & \multicolumn{1}{|l}{PSNR} & 40.2862 & 42.8302 & 40.0708 & 40.5722 & 40.2461 & 40.4436 & 40.0881 & 39.7575 & 40.1068 & 40.4891 \\
    & \multicolumn{1}{|l}{SSIM} & 0.9846 & 0.9891 & 0.9876 & 0.9883 & 0.9881 & 0.9893 & 0.9890 & 0.9868 & 0.9890 & 0.9880 \\
    & \multicolumn{1}{|l}{LAION AES} & 5.1643 & 5.1615 & 5.0640 & 5.1189 & 4.9139 & 5.0389 & 5.0244 & 4.9664 & 4.9457 & 5.0442 \\
    & \multicolumn{1}{|l}{rFID} & 4.7230 & 3.1646 & 3.5066 & 2.4490 & 3.1405 & 2.0564 & 2.3312 & 2.7373 & 2.0353 & 2.9049 \\
    & \multicolumn{1}{|l}{LPIPS} & 0.0161 & 0.0119 & 0.0139 & 0.0111 & 0.0107 & 0.0087 & 0.0097 & 0.0105 & 0.0092 & 0.0113 \\
    \cmidrule{1-12}
    \multirow{5}{*}{\texttt{transparent}} & \multicolumn{1}{|l}{PSNR} & 32.7464 & 35.9282 & 30.5379 & 31.9773 & 31.1065 & 31.9161 & 30.9990 & 31.0960 & 30.9451 & 31.9169 \\
\bottomrule
  \end{tabular}}
\end{table}

\begin{table}[htbp]
  \centering
  \caption{Metric scores across subtypes of AIM and color categories on VAE of FLUX (w/o Norm KL).}
  \label{tab:subset_transposed_metrics_16}
  \resizebox{\textwidth}{!}{%
\begin{tabular}{c c c c c c c c c c c c}
    \toprule
    \multirow{2.5}{*}{\textbf{Subtype}} & \multirow{2.5}{*}{\textbf{Metric}} & \multicolumn{10}{c}{\textbf{Colors}} \\ \cmidrule{3-12}
     &  & \texttt{black} & \texttt{gray} & \texttt{white} & \texttt{red} & \texttt{yellow} & \texttt{green} & \texttt{cyan} & \texttt{blue} & \texttt{magenta} & \texttt{overall} \\
    \midrule
    \multirow{5}{*}{\texttt{portrait}} & \multicolumn{1}{|l}{PSNR} & 38.5509 & 41.8724 & 38.2719 & 39.0076 & 38.6528 & 38.7745 & 38.3025 & 38.0592 & 38.4824 & 38.8860 \\
    & \multicolumn{1}{|l}{SSIM} & 0.9818 & 0.9876 & 0.9849 & 0.9869 & 0.9859 & 0.9880 & 0.9874 & 0.9850 & 0.9878 & 0.9861 \\
    & \multicolumn{1}{|l}{LAION AES} & 5.1598 & 5.1631 & 5.0611 & 5.1102 & 4.9317 & 5.0489 & 5.0314 & 4.9646 & 4.9550 & 5.0473 \\
    & \multicolumn{1}{|l}{rFID} & 6.2641 & 6.0452 & 6.3839 & 3.5284 & 5.4736 & 3.2585 & 4.1544 & 4.2184 & 3.1191 & 4.7162 \\
    & \multicolumn{1}{|l}{LPIPS} & 0.0191 & 0.0152 & 0.0177 & 0.0135 & 0.0135 & 0.0108 & 0.0123 & 0.0127 & 0.0112 & 0.0140 \\
    \cmidrule{1-12}
    \multirow{5}{*}{\texttt{transparent}} & \multicolumn{1}{|l}{PSNR} & 29.7288 & 33.5282 & 27.7907 & 29.1354 & 28.3555 & 28.8551 & 28.0053 & 28.1877 & 28.0979 & 29.0761 \\
\bottomrule
  \end{tabular}}
\end{table}

\begin{table}[htbp]
  \centering
  \caption{Metric scores across subtypes of AIM and color categories on VAE of FLUX (w/o Ref KL).}
  \label{tab:subset_transposed_metrics_17}
  \resizebox{\textwidth}{!}{%
\begin{tabular}{c c c c c c c c c c c c}
    \toprule
    \multirow{2.5}{*}{\textbf{Subtype}} & \multirow{2.5}{*}{\textbf{Metric}} & \multicolumn{10}{c}{\textbf{Colors}} \\ \cmidrule{3-12}
     &  & \texttt{black} & \texttt{gray} & \texttt{white} & \texttt{red} & \texttt{yellow} & \texttt{green} & \texttt{cyan} & \texttt{blue} & \texttt{magenta} & \texttt{overall} \\
    \midrule
    \multirow{5}{*}{\texttt{portrait}} & \multicolumn{1}{|l}{PSNR} & 40.1312 & 42.6648 & 39.9200 & 40.3855 & 40.0629 & 40.3297 & 40.0098 & 39.6624 & 39.9857 & 40.3502 \\
    & \multicolumn{1}{|l}{SSIM} & 0.9840 & 0.9887 & 0.9870 & 0.9878 & 0.9876 & 0.9889 & 0.9885 & 0.9863 & 0.9886 & 0.9875 \\
    & \multicolumn{1}{|l}{LAION AES} & 5.1771 & 5.1678 & 5.0666 & 5.1155 & 4.9246 & 5.0426 & 5.0274 & 4.9635 & 4.9558 & 5.0490 \\
    & \multicolumn{1}{|l}{rFID} & 4.8658 & 3.7301 & 4.1279 & 2.7334 & 3.6530 & 2.2121 & 2.6575 & 2.8246 & 2.0897 & 3.2105 \\
    & \multicolumn{1}{|l}{LPIPS} & 0.0178 & 0.0133 & 0.0153 & 0.0122 & 0.0117 & 0.0095 & 0.0106 & 0.0116 & 0.0102 & 0.0125 \\
    \cmidrule{1-12}
    \multirow{5}{*}{\texttt{transparent}} & \multicolumn{1}{|l}{PSNR} & 32.7316 & 36.2265 & 30.8402 & 32.1378 & 31.3515 & 32.2155 & 31.4091 & 31.3970 & 31.2557 & 32.1739 \\
\bottomrule
  \end{tabular}}
\end{table}

\begin{table}[htbp]
  \centering
  \caption{Metric scores across subtypes of AIM and color categories on VAE of FLUX (w/o GAN).}
  \label{tab:subset_transposed_metrics_18}
  \resizebox{\textwidth}{!}{%
\begin{tabular}{c c c c c c c c c c c c}
    \toprule
    \multirow{2.5}{*}{\textbf{Subtype}} & \multirow{2.5}{*}{\textbf{Metric}} & \multicolumn{10}{c}{\textbf{Colors}} \\ \cmidrule{3-12}
     &  & \texttt{black} & \texttt{gray} & \texttt{white} & \texttt{red} & \texttt{yellow} & \texttt{green} & \texttt{cyan} & \texttt{blue} & \texttt{magenta} & \texttt{overall} \\
    \midrule
    \multirow{5}{*}{\texttt{portrait}} & \multicolumn{1}{|l}{PSNR} & 40.3127 & 43.1050 & 40.1008 & 40.6687 & 40.3176 & 40.4744 & 40.0873 & 39.7835 & 40.1739 & 40.5582 \\
    & \multicolumn{1}{|l}{SSIM} & 0.9851 & 0.9898 & 0.9881 & 0.9891 & 0.9887 & 0.9900 & 0.9897 & 0.9876 & 0.9898 & 0.9887 \\
    & \multicolumn{1}{|l}{LAION AES} & 5.1579 & 5.1723 & 5.0656 & 5.1126 & 4.9149 & 5.0332 & 5.0337 & 4.9590 & 4.9445 & 5.0437 \\
    & \multicolumn{1}{|l}{rFID} & 4.7177 & 4.0622 & 4.2203 & 2.6370 & 3.5575 & 2.3273 & 2.9697 & 2.8760 & 2.1901 & 3.2842 \\
    & \multicolumn{1}{|l}{LPIPS} & 0.0163 & 0.0122 & 0.0146 & 0.0111 & 0.0111 & 0.0088 & 0.0100 & 0.0105 & 0.0091 & 0.0115 \\
    \cmidrule{1-12}
    \multirow{5}{*}{\texttt{transparent}} & \multicolumn{1}{|l}{PSNR} & 31.6978 & 35.9299 & 30.0637 & 31.4137 & 30.6422 & 31.3051 & 30.4507 & 30.4539 & 30.4794 & 31.3818 \\
\bottomrule
  \end{tabular}}
\end{table}

\begin{table}[htbp]
  \centering
  \caption{Metric scores across subtypes of AIM and color categories on VAE of FLUX (w/o LPIPS).}
  \label{tab:subset_transposed_metrics_19}
  \resizebox{\textwidth}{!}{%
\begin{tabular}{c c c c c c c c c c c c}
    \toprule
    \multirow{2.5}{*}{\textbf{Subtype}} & \multirow{2.5}{*}{\textbf{Metric}} & \multicolumn{10}{c}{\textbf{Colors}} \\ \cmidrule{3-12}
     &  & \texttt{black} & \texttt{gray} & \texttt{white} & \texttt{red} & \texttt{yellow} & \texttt{green} & \texttt{cyan} & \texttt{blue} & \texttt{magenta} & \texttt{overall} \\
    \midrule
    \multirow{5}{*}{\texttt{portrait}} & \multicolumn{1}{|l}{PSNR} & 36.5699 & 40.8603 & 36.3526 & 37.0664 & 36.7739 & 36.7335 & 36.2499 & 36.0746 & 36.5464 & 37.0253 \\
    & \multicolumn{1}{|l}{SSIM} & 0.9786 & 0.9865 & 0.9831 & 0.9852 & 0.9843 & 0.9868 & 0.9861 & 0.9825 & 0.9864 & 0.9844 \\
    & \multicolumn{1}{|l}{LAION AES} & 5.1506 & 5.1543 & 5.0381 & 5.1037 & 4.9200 & 5.0348 & 5.0274 & 4.9431 & 4.9473 & 5.0355 \\
    & \multicolumn{1}{|l}{rFID} & 8.3673 & 6.8302 & 7.2476 & 5.4567 & 7.2509 & 5.0560 & 6.1597 & 5.7517 & 4.6405 & 6.3067 \\
    & \multicolumn{1}{|l}{LPIPS} & 0.0226 & 0.0169 & 0.0198 & 0.0173 & 0.0168 & 0.0145 & 0.0159 & 0.0164 & 0.0149 & 0.0172 \\
    \cmidrule{1-12}
    \multirow{5}{*}{\texttt{transparent}} & \multicolumn{1}{|l}{PSNR} & 24.6968 & 28.3907 & 22.5216 & 23.2511 & 22.6062 & 22.9936 & 22.2289 & 22.5321 & 22.2666 & 23.4986 \\
\bottomrule
  \end{tabular}}
\end{table}

\begin{table}[htbp]
  \centering
  \caption{Metric scores across subtypes of AIM and color categories on VAE of FLUX (w/o ABMSE).}
  \label{tab:subset_transposed_metrics_20}
  \resizebox{\textwidth}{!}{%
\begin{tabular}{c c c c c c c c c c c c}
    \toprule
    \multirow{2.5}{*}{\textbf{Subtype}} & \multirow{2.5}{*}{\textbf{Metric}} & \multicolumn{10}{c}{\textbf{Colors}} \\ \cmidrule{3-12}
     &  & \texttt{black} & \texttt{gray} & \texttt{white} & \texttt{red} & \texttt{yellow} & \texttt{green} & \texttt{cyan} & \texttt{blue} & \texttt{magenta} & \texttt{overall} \\
    \midrule
    \multirow{5}{*}{\texttt{portrait}} & \multicolumn{1}{|l}{PSNR} & 39.8594 & 42.4175 & 39.6565 & 40.2818 & 39.7875 & 40.0540 & 39.7738 & 39.4529 & 39.8838 & 40.1297 \\
    & \multicolumn{1}{|l}{SSIM} & 0.9821 & 0.9875 & 0.9858 & 0.9869 & 0.9866 & 0.9881 & 0.9877 & 0.9851 & 0.9878 & 0.9864 \\
    & \multicolumn{1}{|l}{LAION AES} & 5.1740 & 5.1882 & 5.0794 & 5.1274 & 4.9326 & 5.0511 & 5.0424 & 4.9615 & 4.9475 & 5.0560 \\
    & \multicolumn{1}{|l}{rFID} & 5.6228 & 6.8072 & 5.5474 & 2.6233 & 4.8984 & 2.9176 & 3.7935 & 3.1263 & 2.3099 & 4.1829 \\
    & \multicolumn{1}{|l}{LPIPS} & 0.0198 & 0.0160 & 0.0162 & 0.0122 & 0.0118 & 0.0090 & 0.0103 & 0.0112 & 0.0094 & 0.0129 \\
    \cmidrule{1-12}
    \multirow{5}{*}{\texttt{transparent}} & \multicolumn{1}{|l}{PSNR} & 32.3013 & 35.3760 & 30.1507 & 31.6711 & 30.6555 & 31.5462 & 30.7704 & 30.9122 & 30.7468 & 31.5700 \\
\bottomrule
  \end{tabular}}
\end{table}

\begin{table}[htbp]
  \centering
  \caption{Metric scores across subtypes of AIM and color categories on VAE of SDXL (ours).}
  \label{tab:subset_transposed_metrics_21}
  \resizebox{\textwidth}{!}{%
\begin{tabular}{c c c c c c c c c c c c}
    \toprule
    \multirow{2.5}{*}{\textbf{Subtype}} & \multirow{2.5}{*}{\textbf{Metric}} & \multicolumn{10}{c}{\textbf{Colors}} \\ \cmidrule{3-12}
     &  & \texttt{black} & \texttt{gray} & \texttt{white} & \texttt{red} & \texttt{yellow} & \texttt{green} & \texttt{cyan} & \texttt{blue} & \texttt{magenta} & \texttt{overall} \\
    \midrule
    \multirow{5}{*}{\texttt{portrait}} & \multicolumn{1}{|l}{PSNR} & 37.6675 & 38.7091 & 37.0896 & 37.6714 & 37.2382 & 37.5797 & 37.3591 & 37.3685 & 37.4252 & 37.5676 \\
    & \multicolumn{1}{|l}{SSIM} & 0.9671 & 0.9708 & 0.9694 & 0.9704 & 0.9699 & 0.9710 & 0.9707 & 0.9693 & 0.9709 & 0.9700 \\
    & \multicolumn{1}{|l}{LAION AES} & 5.1871 & 5.1979 & 5.1041 & 5.1119 & 4.9399 & 5.0166 & 5.0471 & 4.9506 & 4.9370 & 5.0547 \\
    & \multicolumn{1}{|l}{rFID} & 8.6374 & 10.1861 & 9.1757 & 6.0759 & 7.9020 & 5.9156 & 6.5183 & 5.7959 & 5.1613 & 7.2631 \\
    & \multicolumn{1}{|l}{LPIPS} & 0.0430 & 0.0423 & 0.0421 & 0.0358 & 0.0361 & 0.0324 & 0.0338 & 0.0334 & 0.0326 & 0.0368 \\
    \cmidrule{1-12}
    \multirow{5}{*}{\texttt{transparent}} & \multicolumn{1}{|l}{PSNR} & 34.0029 & 37.1520 & 33.0656 & 33.9251 & 33.3083 & 34.1245 & 33.7798 & 33.5730 & 33.5541 & 34.0539 \\
\bottomrule
  \end{tabular}}
\end{table}

\begin{table}[htbp]
  \centering
  \caption{Metric scores across subtypes of AIM and color categories on VAE of SDXL (w/o Norm KL).}
  \label{tab:subset_transposed_metrics_22}
  \resizebox{\textwidth}{!}{%
\begin{tabular}{c c c c c c c c c c c c}
    \toprule
    \multirow{2.5}{*}{\textbf{Subtype}} & \multirow{2.5}{*}{\textbf{Metric}} & \multicolumn{10}{c}{\textbf{Colors}} \\ \cmidrule{3-12}
     &  & \texttt{black} & \texttt{gray} & \texttt{white} & \texttt{red} & \texttt{yellow} & \texttt{green} & \texttt{cyan} & \texttt{blue} & \texttt{magenta} & \texttt{overall} \\
    \midrule
    \multirow{5}{*}{\texttt{portrait}} & \multicolumn{1}{|l}{PSNR} & 36.4746 & 38.3820 & 35.9757 & 36.7160 & 36.3575 & 36.5646 & 36.1745 & 36.1212 & 36.3180 & 36.5649 \\
    & \multicolumn{1}{|l}{SSIM} & 0.9641 & 0.9698 & 0.9675 & 0.9689 & 0.9683 & 0.9701 & 0.9696 & 0.9671 & 0.9698 & 0.9684 \\
    & \multicolumn{1}{|l}{LAION AES} & 5.1593 & 5.1762 & 5.0724 & 5.0840 & 4.8959 & 5.0014 & 5.0296 & 4.9389 & 4.9164 & 5.0305 \\
    & \multicolumn{1}{|l}{rFID} & 11.4563 & 11.0646 & 10.9848 & 7.5095 & 9.6601 & 7.1380 & 7.7696 & 7.6093 & 6.9458 & 8.9042 \\
    & \multicolumn{1}{|l}{LPIPS} & 0.0462 & 0.0436 & 0.0453 & 0.0381 & 0.0385 & 0.0338 & 0.0361 & 0.0356 & 0.0341 & 0.0390 \\
    \cmidrule{1-12}
    \multirow{5}{*}{\texttt{transparent}} & \multicolumn{1}{|l}{PSNR} & 29.2929 & 33.3247 & 27.8862 & 29.1718 & 28.7140 & 29.2333 & 28.3210 & 28.3100 & 28.2088 & 29.1625 \\
\bottomrule
  \end{tabular}}
\end{table}

\begin{table}[htbp]
  \centering
  \caption{Metric scores across subtypes of AIM and color categories on VAE of SDXL (w/o Ref KL).}
  \label{tab:subset_transposed_metrics_23}
  \resizebox{\textwidth}{!}{%
\begin{tabular}{c c c c c c c c c c c c}
    \toprule
    \multirow{2.5}{*}{\textbf{Subtype}} & \multirow{2.5}{*}{\textbf{Metric}} & \multicolumn{10}{c}{\textbf{Colors}} \\ \cmidrule{3-12}
     &  & \texttt{black} & \texttt{gray} & \texttt{white} & \texttt{red} & \texttt{yellow} & \texttt{green} & \texttt{cyan} & \texttt{blue} & \texttt{magenta} & \texttt{overall} \\
    \midrule
    \multirow{5}{*}{\texttt{portrait}} & \multicolumn{1}{|l}{PSNR} & 36.5102 & 38.3645 & 36.0600 & 36.7652 & 36.4221 & 36.6211 & 36.2449 & 36.1715 & 36.3789 & 36.6154 \\
    & \multicolumn{1}{|l}{SSIM} & 0.9641 & 0.9697 & 0.9677 & 0.9689 & 0.9685 & 0.9700 & 0.9696 & 0.9670 & 0.9698 & 0.9684 \\
    & \multicolumn{1}{|l}{LAION AES} & 5.1657 & 5.1846 & 5.0962 & 5.0851 & 4.9031 & 5.0043 & 5.0288 & 4.9429 & 4.9228 & 5.0371 \\
    & \multicolumn{1}{|l}{rFID} & 11.6975 & 10.3056 & 9.7826 & 7.1901 & 9.0144 & 6.5155 & 7.2468 & 7.4109 & 6.7683 & 8.4369 \\
    & \multicolumn{1}{|l}{LPIPS} & 0.0455 & 0.0426 & 0.0439 & 0.0372 & 0.0373 & 0.0329 & 0.0351 & 0.0349 & 0.0334 & 0.0381 \\
    \cmidrule{1-12}
    \multirow{5}{*}{\texttt{transparent}} & \multicolumn{1}{|l}{PSNR} & 29.9848 & 33.8929 & 28.4385 & 29.8346 & 29.2755 & 29.8737 & 28.9514 & 28.9870 & 28.8585 & 29.7885 \\
\bottomrule
  \end{tabular}}
\end{table}

\begin{table}[htbp]
  \centering
  \caption{Metric scores across subtypes of AIM and color categories on VAE of SDXL (w/o GAN).}
  \label{tab:subset_transposed_metrics_24}
  \resizebox{\textwidth}{!}{%
\begin{tabular}{c c c c c c c c c c c c}
    \toprule
    \multirow{2.5}{*}{\textbf{Subtype}} & \multirow{2.5}{*}{\textbf{Metric}} & \multicolumn{10}{c}{\textbf{Colors}} \\ \cmidrule{3-12}
     &  & \texttt{black} & \texttt{gray} & \texttt{white} & \texttt{red} & \texttt{yellow} & \texttt{green} & \texttt{cyan} & \texttt{blue} & \texttt{magenta} & \texttt{overall} \\
    \midrule
    \multirow{5}{*}{\texttt{portrait}} & \multicolumn{1}{|l}{PSNR} & 36.7320 & 38.5246 & 36.2216 & 36.9466 & 36.5775 & 36.8067 & 36.4386 & 36.3937 & 36.5608 & 36.8002 \\
    & \multicolumn{1}{|l}{SSIM} & 0.9647 & 0.9704 & 0.9683 & 0.9696 & 0.9691 & 0.9707 & 0.9702 & 0.9678 & 0.9704 & 0.9690 \\
    & \multicolumn{1}{|l}{LAION AES} & 5.1517 & 5.1743 & 5.0571 & 5.0904 & 4.8949 & 5.0084 & 5.0292 & 4.9452 & 4.9174 & 5.0298 \\
    & \multicolumn{1}{|l}{rFID} & 11.3116 & 11.4391 & 10.1436 & 7.1283 & 9.4828 & 6.6206 & 7.6032 & 7.1238 & 6.7191 & 8.6191 \\
    & \multicolumn{1}{|l}{LPIPS} & 0.0457 & 0.0429 & 0.0449 & 0.0375 & 0.0379 & 0.0332 & 0.0354 & 0.0350 & 0.0336 & 0.0385 \\
    \cmidrule{1-12}
    \multirow{5}{*}{\texttt{transparent}} & \multicolumn{1}{|l}{PSNR} & 29.5129 & 33.7601 & 27.9406 & 29.3151 & 28.7980 & 29.4456 & 28.4825 & 28.5114 & 28.3432 & 29.3455 \\
\bottomrule
  \end{tabular}}
\end{table}

\begin{table}[htbp]
  \centering
  \caption{Metric scores across subtypes of AIM and color categories on VAE of SDXL (w/o LPIPS).}
  \label{tab:subset_transposed_metrics_25}
  \resizebox{\textwidth}{!}{%
\begin{tabular}{c c c c c c c c c c c c}
    \toprule
    \multirow{2.5}{*}{\textbf{Subtype}} & \multirow{2.5}{*}{\textbf{Metric}} & \multicolumn{10}{c}{\textbf{Colors}} \\ \cmidrule{3-12}
     &  & \texttt{black} & \texttt{gray} & \texttt{white} & \texttt{red} & \texttt{yellow} & \texttt{green} & \texttt{cyan} & \texttt{blue} & \texttt{magenta} & \texttt{overall} \\
    \midrule
    \multirow{5}{*}{\texttt{portrait}} & \multicolumn{1}{|l}{PSNR} & 36.3674 & 38.3427 & 35.8862 & 36.7687 & 36.2056 & 36.2692 & 35.9600 & 36.0784 & 36.4217 & 36.4778 \\
    & \multicolumn{1}{|l}{SSIM} & 0.9641 & 0.9697 & 0.9674 & 0.9689 & 0.9682 & 0.9700 & 0.9695 & 0.9671 & 0.9698 & 0.9683 \\
    & \multicolumn{1}{|l}{LAION AES} & 5.1359 & 5.1682 & 5.0700 & 5.0833 & 4.9013 & 4.9964 & 5.0246 & 4.9396 & 4.9223 & 5.0268 \\
    & \multicolumn{1}{|l}{rFID} & 10.8159 & 10.3261 & 10.1067 & 7.1798 & 9.1609 & 6.8984 & 8.2479 & 7.2155 & 6.2245 & 8.4640 \\
    & \multicolumn{1}{|l}{LPIPS} & 0.0457 & 0.0433 & 0.0446 & 0.0370 & 0.0382 & 0.0339 & 0.0360 & 0.0351 & 0.0332 & 0.0386 \\
    \cmidrule{1-12}
    \multirow{5}{*}{\texttt{transparent}} & \multicolumn{1}{|l}{PSNR} & 29.8215 & 33.6546 & 28.3403 & 29.9983 & 28.8181 & 29.1258 & 28.6430 & 29.0760 & 29.2703 & 29.6387 \\
\bottomrule
  \end{tabular}}
\end{table}

\begin{table}[htbp]
  \centering
  \caption{Metric scores across subtypes of AIM and color categories on VAE of SDXL (w/o ABMSE).}
  \label{tab:subset_transposed_metrics_26}
  \resizebox{\textwidth}{!}{%
\begin{tabular}{c c c c c c c c c c c c}
    \toprule
    \multirow{2.5}{*}{\textbf{Subtype}} & \multirow{2.5}{*}{\textbf{Metric}} & \multicolumn{10}{c}{\textbf{Colors}} \\ \cmidrule{3-12}
     &  & \texttt{black} & \texttt{gray} & \texttt{white} & \texttt{red} & \texttt{yellow} & \texttt{green} & \texttt{cyan} & \texttt{blue} & \texttt{magenta} & \texttt{overall} \\
    \midrule
    \multirow{5}{*}{\texttt{portrait}} & \multicolumn{1}{|l}{PSNR} & 36.7638 & 38.0540 & 36.1074 & 36.9684 & 36.4335 & 36.7649 & 36.4643 & 36.4923 & 36.6376 & 36.7429 \\
    & \multicolumn{1}{|l}{SSIM} & 0.9628 & 0.9666 & 0.9624 & 0.9675 & 0.9641 & 0.9674 & 0.9664 & 0.9659 & 0.9681 & 0.9657 \\
    & \multicolumn{1}{|l}{LAION AES} & 5.1830 & 5.2021 & 5.0998 & 5.1076 & 4.9582 & 5.0416 & 5.0637 & 4.9568 & 4.9539 & 5.0630 \\
    & \multicolumn{1}{|l}{rFID} & 10.4060 & 14.1848 & 12.3171 & 7.6433 & 11.2678 & 7.4332 & 8.7145 & 7.3389 & 7.1011 & 9.6007 \\
    & \multicolumn{1}{|l}{LPIPS} & 0.0463 & 0.0519 & 0.0547 & 0.0368 & 0.0400 & 0.0327 & 0.0353 & 0.0341 & 0.0329 & 0.0405 \\
    \cmidrule{1-12}
    \multirow{5}{*}{\texttt{transparent}} & \multicolumn{1}{|l}{PSNR} & 31.6651 & 35.3412 & 30.7818 & 31.7570 & 31.3339 & 32.0047 & 31.4061 & 31.1377 & 31.1593 & 31.8430 \\
\bottomrule
  \end{tabular}}
\end{table}

%%%%%%%%%%%%%%%%%%%%%%%%%%%%%%%%%%%%%%%%%%%%%%%%%%%%%%%%%%%%

\end{document}